\documentclass{llncs}

\pdfoutput=1

\usepackage{amsmath}
\usepackage{multirow}
\usepackage{booktabs}
\usepackage{array}

\usepackage{morefloats}

\usepackage{graphicx}
\DeclareGraphicsExtensions{.png,.jpg,jpeg}

\begin{document}

\title{A CUDA-Based Real Parameter Optimization Benchmark}

\author{Ke Ding \and Ying Tan}
\institute{School of Electronics Engineering and Computer Science, Peking University}

\maketitle


%
%
\begin{abstract}
Benchmarking is key for developing and comparing optimization algorithms.
In this paper, a CUDA-based real parameter optimization benchmark (cuROB) is introduced.
Test functions of diverse properties are included within cuROB and implemented efficiently with CUDA.
Speedup of one order of magnitude can be achieved in comparison with CPU-based benchmark of CEC'14.

\keywordname{ Optimization Methods, Optimization Benchmark, GPU, CUDA. }
\end{abstract}

\section{Introduction}

Proposed algorithms are usually tested on benchmark for comparing both performance and efficiency.
However, as it can be a very tedious task to select and implement test functions rigorously.
Thanks to GPUs' massive parallelism, a GPU-based optimization function suit will be beneficial to test and compare optimization algorithms.

Based on the well known CPU-based benchmarks presented in \cite{BBOB10,CEC13,CEC14},
we proposed a CUDA-based real parameter optimization test suit, called cuROB, targeting on GPUs.
We think cuROB can be helpful for assessing GPU-based optimization algorithms, and hopefully, conventional CPU-based algorithms can benefit from cuROB's fast execution.

Considering the fact that research on the single objective optimization algorithms is the basis of the research on the more complex optimization algorithms such as constrained optimization algorithms, multi-objective optimizations algorithms and so forth,
in this first release of cuROB a suit of single objective real-parameter optimization function are defined and implemented.

The test functions are selected according to the following criteria: 1) the functions should be scalable in dimension so that algorithms can be tested under various complexity;  2) the expressions of the functions should be with good parallelism, thus efficient implementation is possible on GPUs; 3) the functions should be comprehensible such that algorithm behaviours can be analysed in the topological context; 4) last but most important, the test suit should cover functions of various properties in order to get a systematic evaluation of the optimization algorithms.

The source code and a sample can be download from \url{code.google.com/p/curob/}.

\subsection{Symbol Conventions and Definitions}

Symbols and definitions used in the report are described in the following.
By default, all vectors refer to column vectors, and are depicted by lowercase letter and typeset in bold.

\begin{itemize}
  \item $[\cdot]$ indicates the nearest integer value
  \item $\lfloor\cdot\rfloor$ indicates the largest integer less than or equal to
  \item $\mathbf{x}_i$ denotes $i$-th element of vector $\mathbf{x}$
  \item $f(\cdot)$, $g(\cdot)$ and $G(\cdot)$  multi-variable functions
  \item $f_{opt}$ optimal (minimal) value of function $f$
  \item $\mathbf{x}^{opt}$  optimal solution vector, such that $f(\mathbf{x}^{opt}) = f_{opt}$
  \item $\mathbf{R}$  normalized orthogonal matrix for rotation
  \item $D$ dimension
  \item $\mathbf{1} = (1,\dots,1)^T$ all one vector
\end{itemize}

\subsection{General Setup}

The general setup of the test suit is presented as follows.

\begin{itemize}
  \item \textbf{Dimensions} The test suit is scalable in terms of dimension.
  Within the hardware limit, any dimension $D \ge 2$ works.
  However, to construct a real hybrid function, $D$ should be at least $10$.
  \item \textbf{Search Space} All functions are defined and can be evaluated over $\mathcal{R}^D$, while the actual search domain is given as $[-100,100]^D$.
  \item \textbf{$f_{opt}$} All functions, by definition, have a minimal value of 0, a bias ($f_{opt}$) can be added to each function.
  The selection can be arbitrary, $f_{opt}$ for each function in the test suit is listed in Tab.\,\ref{tab:summary}.
  \item \textbf{$\mathbf{x}^{opt}$} The optimum point of each function is located at original.
  $\mathbf{x}^{opt}$ which is randomly distributed in $[-70,70]^D$, is selected as the new optimum.
  \item \textbf{Rotation Matrix} To derive non-separable functions from separable ones, the search space is rotated by a normalized orthogonal matrix $\mathbf{R}$.
  For a given function in one dimension, a different $\mathbf{R}$ is used.
  Variables are divided into three (almost) equal-sized subcomponents randomly.
  The rotation matrix for each subcomponent is generated from standard normally distributed entries by Gram-Schmidt orthonormalization.
  Then, these matrices consist of the $\mathbf{R}$ actually used.
\end{itemize}

\subsection{CUDA Interface and Implementation}

A simple description of the interface and implementation is given in the following.
For detail, see the source code and the accompanied readme file.

\subsubsection{Interface}
Only benchmark.h need to be included to access the test functions, and the CUDA file benchmark.cu need be compiled and linked.
Before the compiling start, two macro, DIM and MAX\_CONCURRENCY should be modified accordingly.
DIM defines the dimension of the test suit to used while MAX\_CONCURRENCY controls the most function evaluations be invoked concurrently.
As memory needed to be pre-allocated, limited by the hardware, don't set MAX\_CONCURRENCY greater than actually used.

Host interface function initialize\,() accomplish all initialization tasks, so must be called before any test function can be evaluated.
Allocated resource is released by host interface function dispose\,().

Both double precision and single precision are supported through func\_evaluate\,() and func\_evaluatef\,() respectively.
Take note that device pointers should be passed to these two functions.
For the convenience of CPU code, C interfaces are provided, with h\_func\_evaluate for double precision and h\_func\_evaluatef for single precision.
(In fact, they are just wrappers of the GPU interfaces.)

\subsubsection{Efficiency Concerns}
When configuration of the suit, some should be taken care for the sake of efficiency.
It is better to evaluation a batch of vectors than many smaller.
Dimension is a fold of 32 (the warp size) can more efficient.
For example, dimension of 96 is much more efficient than 100, even though 100 is little greater than 96.

\subsection{Test Suite Summary}

The test functions fall into four categories: unimodal functions, basic multi-modal functions, hybrid functions and composition functions.
The summary of the suit is listed in Tab.\,\ref{tab:summary}.
Detailed information of each function will given in the following sections.

\newcommand{\minitab}[2][l]{\begin{tabular}{#1}#2\end{tabular}}

\begin{table}
\caption{Summary of cuROB's Test Functions}\label{tab:summary}
\centering
\resizebox{1\textwidth}{!}{

\begin{tabular}{|p{2.1cm}|c|l|c|c|}
  \hline
  \toprule
   & No. & \hspace{3cm}Functions & ID & Description\\
  \midrule
  \multirow{6}{*}{\minitab[c]{Unimodal\\ Functions}} & 0 & Rotated Sphere  & SPHERE & \multirow{2}*{\minitab[c]{Optimum easy\\ to track}}\\
  \cline{2-4}
  & 1 & Rotated Ellipsoid  & ELLIPSOID &\\
  \cline{2-5}
  & 2 & Rotated   Elliptic  & ELLIPTIC & \multirow{6}*{\minitab[c]{Optimum hard\\ to track}}\\
  \cline{2-4}
  & 3 & Rotated Discus  & DISCUS & \\
  \cline{2-4}
  & 4 & Rotated Bent Cigar  & CIGAR & \\
  \cline{2-4}
  & 5 & Rotated Different Powers  & POWERS & \\
  \cline{2-4}
  & 6 & Rotated Sharp Valley  & SHARPV & \\
  \hline\hline

  \multirow{14}{*}{\minitab[c]{Basic\\Multi-modal\\Functions}} & 7 & Rotated Step  & STEP & \multirow{6}*{\minitab[c]{With \\ adepuate\\ global\\ structure}}\\
  \cline{2-4}
  &  8 & Rotated Weierstrass  & WEIERSTRASS &  \\
  \cline{2-4}
  & 9 & Rotated Griewank  & GRIEWANK & \\
  \cline{2-4}
   & 10 &  Rastrigin  & RARSTRIGIN\_U & \\
  \cline{2-4}
  & 11 &  Rotated Rastrigin  & RARSTRIGIN & \\
  \cline{2-4}
  & 12 & Rotated Schaffer's F7  & SCHAFFERSF7 & \\
  \cline{2-4}
  & 13 & Rotated Expanded Griewank plus Rosenbrock  & GRIE\_ROSEN & \\
  \cline{2-5}

   & 14 & Rotated Rosenbrock  & ROSENBROCK & \multirow{9}*{\minitab[c]{With\\weak\\ global\\ structure}}\\
  \cline{2-4}
  & 15 &  Modified Schwefel  & SCHWEFEL\_U & \\
  \cline{2-4}
  & 16 & Rotated Modified Schwefel  & SCHWEFEL & \\
  \cline{2-4}
    & 17 & Rotated Katsuura  & KATSUURA & \\
  \cline{2-4}
  & 18 &  Rotated Lunacek bi-Rastrigin  & LUNACEK & \\
  \cline{2-4}
  & 19 & Rotated Ackley  & ACKLEY &\\
  \cline{2-4}
  & 20 & Rotated HappyCat  & HAPPYCAT &\\
  \cline{2-4}
  & 21 & Rotated HGBat  & HGBAT & \\
  \cline{2-4}
  & 22 & Rotated Expanded Schaffer's F6  & SCHAFFERSF6 & \\
  \hline\hline

  \multirow{6}{*}{\minitab[c]{Hybrid\\ Functions}} & 23 & Hybrid Function 1 & HYBRID1 &  \multirow{6}*{\minitab[c]{With different\\ properties for\\ different variables \\ subcomponents}}\\
  \cline{2-4}
  & 24 & Hybrid Function 2 & HYBRID2 & \\
  \cline{2-4}
  & 25 & Hybrid Function 3 & HYBRID3 & \\
  \cline{2-4}
  & 26 & Hybrid Function 4 & HYBRID4 & \\
  \cline{2-4}
  & 27 & Hybrid Function 5 & HYBRID5 & \\
  \cline{2-4}
  & 28 & Hybrid Function 6 & HYBRID6 & \\
  \hline\hline

    \multirow{8}{*}{\minitab[c]{Composition \\Functions}} & 29 & Composition Function 1 & COMPOSITION1 & \multirow{8}*{\minitab[c]{Properties similar\\ to particular\\ sub-function\\ when approaching \\the corresponding\\ optimum}}\\
  \cline{2-4}
  & 30 & Composition Function 2 & COMPOSITION2 & \\
  \cline{2-4}
  & 31 & Composition Function 3 & COMPOSITION3 & \\
  \cline{2-4}
  & 32 & Composition Function 4 & COMPOSITION4 & \\
  \cline{2-4}
  & 33 & Composition Function 5 & COMPOSITION5 & \\
  \cline{2-4}
  & 34 & Composition Function 6 & COMPOSITION6 & \\
  \cline{2-4}
  & 35 & Composition Function 7 & COMPOSITION7 & \\
  \cline{2-4}
  & 36 & Composition Function 8 & COMPOSITION8 & \\
  \midrule
  \multicolumn{5}{|c|}{Search Space: $[-100,100]^D$, $f_{opt}=100$}\\

  \bottomrule
\end{tabular}

}

\end{table}

\section{Speedup}

Under different hardware, various speedups can be achieved.
30 functions are the same as CEC'14 benchmark.
We test the cuROB's speedup with these 30 functions under the following settings: Windows 7 SP1 x64 running on  Intel i5-2310 CPU with NVIDIA 560 Ti, the CUDA version is 5.5.
50 evaluations were performed concurrently and repeated 1000 runs.
The evaluation data were generated randomly from uniform distribution.

The speedups with respect to different dimension are listed by Tab.\,\ref{tab:speedup:float} (single precision) and  Tab.\,\ref{tab:speedup:double} (double precision).
Notice that the corresponding dimensions of cuROB are 10, 32, 64 and 96 respectively and the numbers are as in Tab.\,\ref{tab:summary}

\begin{figure}
  \centering
  \includegraphics[width=.8\textwidth]{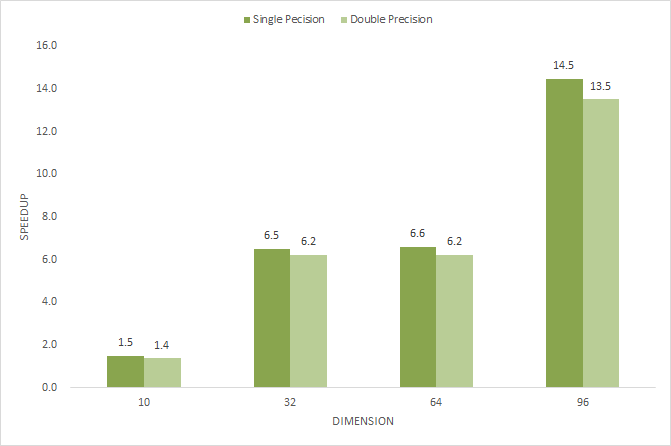}\\
  \caption{Overall Speedup}\label{fig:overall_speedup}
\end{figure}

Fig.\, \ref{fig:overall_speedup} demonstrates the overall speedup for each dimension.
On average, cuROB is never slower than its CPU-base CEC'14 benchmark, and speedup of one order of magnitude can be achieved when dimension is high.
Single precision is more efficient than double precision as far as execution time is concerned.

\begin{table}[htbp]
  \centering
  \caption{Speedup (single Precision)}
    \begin{tabular}{c|rrrrrrrrrr}
    \toprule
    D     & \multicolumn{1}{c}{NO.3} & \multicolumn{1}{c}{NO.4} & \multicolumn{1}{c}{NO.5} & \multicolumn{1}{c}{NO.8} & \multicolumn{1}{c}{NO.9} & \multicolumn{1}{c}{NO.10} & \multicolumn{1}{c}{NO.11} & \multicolumn{1}{c}{NO.13} & \multicolumn{1}{c}{NO.14} & \multicolumn{1}{c}{NO.15} \\
    \midrule
    10    & 0.59  & 0.20  & 0.18  & 12.23  & 0.49  & 0.28  & 0.31  & 0.32  & 0.14  & 0.77  \\
    32    & 3.82  & 2.42  & 2.00  & 47.19  & 3.54  & 1.67  & 3.83  & 5.09  & 2.06  & 3.54  \\
    64    & 4.67  & 2.72  & 2.29  & 50.17  & 3.56  & 0.93  & 3.06  & 2.88  & 2.20  & 3.39  \\
    94   & 13.40  & 10.10  & 8.50  & 84.31  & 11.13  & 1.82  & 9.98  & 9.66  & 8.75  & 6.73  \\

   \toprule
    D     & \multicolumn{1}{c}{NO.16} & \multicolumn{1}{c}{NO.17} & \multicolumn{1}{c}{NO.19} & \multicolumn{1}{c}{NO.20} & \multicolumn{1}{c}{NO.21} & \multicolumn{1}{c}{NO.22} & \multicolumn{1}{c}{NO.23} & \multicolumn{1}{c}{NO.24} & \multicolumn{1}{c}{NO.25} & \multicolumn{1}{c}{NO.26} \\
    \midrule
    10    & 0.80  & 3.25  & 0.36  & 0.20  & 0.26  & 0.45  & 0.63  & 0.44  & 2.80  & 0.52  \\
    32    & 5.57  & 10.04  & 3.46  & 1.22  & 1.42  & 6.44  & 3.95  & 3.43  & 11.47  & 3.36  \\
    64    & 5.45  & 13.19  & 3.27  & 2.10  & 2.27  & 3.81  & 4.62  & 3.07  & 14.17  & 3.34  \\
    96   & 14.38  & 23.68  & 11.32  & 8.26  & 8.49  & 11.60  & 13.67  & 10.64  & 30.11  & 10.71  \\
    \toprule
    D     & \multicolumn{1}{c}{NO.27} & \multicolumn{1}{c}{NO.28} & \multicolumn{1}{c}{NO.29} & \multicolumn{1}{c}{NO.30} & \multicolumn{1}{c}{NO.31} & \multicolumn{1}{c}{NO.32} & \multicolumn{1}{c}{NO.33} & \multicolumn{1}{c}{NO.34} & \multicolumn{1}{c}{NO.35} & \multicolumn{1}{c}{NO.36} \\
    \midrule
    10    & 0.65  & 0.72  & 0.70  & 0.55  & 0.71  & 3.49  & 3.50  & 0.84  & 1.28  & 0.70  \\
    32    & 2.73  & 3.09  & 3.63  & 3.10  & 4.10  & 12.39  & 12.51  & 5.25  & 5.19  & 3.33  \\
    64    & 3.86  & 4.01  & 3.21  & 2.67  & 3.38  & 12.68  & 12.63  & 3.80  & 5.27  & 3.13  \\
    96   & 12.04  & 11.32  & 8.15  & 6.27  & 8.49  & 23.67  & 23.64  & 9.50  & 11.79  & 7.93  \\
    \bottomrule
    \end{tabular}%
  \label{tab:speedup:float}%
\end{table}%

\begin{table}[htbp]
  \centering
  \caption{Speedup (Double Precision)}
    \begin{tabular}{c|rrrrrrrrrr}
    \toprule
    D     & \multicolumn{1}{c}{NO.3} & \multicolumn{1}{c}{NO.4} & \multicolumn{1}{c}{NO.5} & \multicolumn{1}{c}{NO.8} & \multicolumn{1}{c}{NO.9} & \multicolumn{1}{c}{NO.10} & \multicolumn{1}{c}{NO.11} & \multicolumn{1}{c}{NO.13} & \multicolumn{1}{c}{NO.14} & \multicolumn{1}{c}{NO.15} \\
    \midrule
    10    & 0.56 & 0.19  & 0.17  & 9.04  & 0.43  & 0.26  & 0.29  & 0.30  & 0.14  & 0.75  \\
    32    & 3.78 & 2.43  & 1.80  & 33.37  & 3.09  & 1.59  & 3.52  & 4.81  & 1.97  & 3.53  \\
    64    & 4.34 & 2.49  & 1.93  & 30.82  & 3.15  & 0.92  & 2.87  & 2.74  & 2.11  & 3.29  \\
    96   & 12.27 & 9.24  & 6.95  & 46.01  & 9.72  & 1.78  & 9.62  & 8.74  & 7.87  & 5.92  \\
    \toprule
    D     & \multicolumn{1}{c}{NO.16} & \multicolumn{1}{c}{NO.17} & \multicolumn{1}{c}{NO.19} & \multicolumn{1}{c}{NO.20} & \multicolumn{1}{c}{NO.21} & \multicolumn{1}{c}{NO.22} & \multicolumn{1}{c}{NO.23} & \multicolumn{1}{c}{NO.24} & \multicolumn{1}{c}{NO.25} & \multicolumn{1}{c}{NO.26} \\
    \midrule
       10    & 0.79  & 2.32  & 0.34  & 0.18  & 0.26  & 0.45  & 0.59  & 0.43  & 1.97  & 0.52  \\
    32    & 5.10  & 6.79  & 3.28  & 1.13  & 1.29  & 6.10  & 3.63  & 3.14  & 8.15  & 3.23  \\
    64    & 4.75  & 8.29  & 3.06  & 1.99  & 2.18  & 3.32  & 4.02  & 2.77  & 9.80  & 2.92  \\
    96   & 11.91  & 13.81  & 9.75  & 7.37  & 7.78  & 10.24  & 11.55  & 9.57  & 20.81  & 9.40  \\
    \toprule
    D     & \multicolumn{1}{c}{NO.27} & \multicolumn{1}{c}{NO.28} & \multicolumn{1}{c}{NO.29} & \multicolumn{1}{c}{NO.30} & \multicolumn{1}{c}{NO.31} & \multicolumn{1}{c}{NO.32} & \multicolumn{1}{c}{NO.33} & \multicolumn{1}{c}{NO.34} & \multicolumn{1}{c}{NO.35} & \multicolumn{1}{c}{NO.36} \\
    \midrule
    10    & 0.79  & 2.32  & 0.34  & 0.18  & 0.26  & 0.45  & 0.59  & 0.43  & 1.97  & 0.52  \\
    32    & 5.10  & 6.79  & 3.28  & 1.13  & 1.29  & 6.10  & 3.63  & 3.14  & 8.15  & 3.23  \\
    64    & 4.75  & 8.29  & 3.06  & 1.99  & 2.18  & 3.32  & 4.02  & 2.77  & 9.80  & 2.92  \\
    96   & 11.91  & 13.81  & 9.75  & 7.37  & 7.78  & 10.24  & 11.55  & 9.57  & 20.81  & 9.40  \\
    \bottomrule
    \end{tabular}%
  \label{tab:speedup:double}%
\end{table}%

\section{Unimodal Functions}

\subsection{Shifted and Rotated Sphere Function}

\begin{equation}\label{eq:sphere}
  f_1(\mathbf{x}) = \sum_{i=1}^D\mathbf{z}_i^2 + f_{opt}
\end{equation}
where $\mathbf{z}=\mathbf{R}(\mathbf{x}-\mathbf{x}^{opt})$.

\subsubsection*{Properties}

\begin{itemize}
  \item Unimodal
  \item Non-separable
  \item Highly symmetric, in particular rotationally invariant
\end{itemize}

\subsection{Shifted and Rotated Ellipsoid Function}

\begin{equation}\label{eq:ellipsoid}
  f_4(\mathbf{x}) = \sum_{i=1}^Di\cdot\mathbf{z}_i^2 + f_{opt}
\end{equation}
where $\mathbf{z}=\mathbf{R}(\mathbf{x}-\mathbf{x}^{opt})$.

\subsubsection*{Properties}

\begin{itemize}
  \item Unimodal
  \item Non-separable
\end{itemize}

\subsection{Shifted and Rotated High Conditioned Elliptic Function}

\begin{equation}\label{eq:elliptic}
  f_2(\mathbf{x}) = \sum_{i=1}^D(10^6)^{\frac{i-1}{D-1}}\mathbf{z}_i^2 + f_{opt}
\end{equation}
where $\mathbf{z}=\mathbf{R}(\mathbf{x}-\mathbf{x}^{opt})$.

\subsubsection*{Properties}

\begin{itemize}
  \item Unimodal
  \item Non-separable
  \item Quadratic ill-conditioned
  \item Smooth local irregularities
\end{itemize}

\subsection{Shifted and Rotated Discus Function}

\begin{equation}\label{eq:discus}
  f_5(\mathbf{x}) = 10^6\cdot\mathbf{z}_1^2+\sum_{i=2}^D\mathbf{z}_i^2 + f_{opt}
\end{equation}
where $\mathbf{z}=\mathbf{R}(\mathbf{x}-\mathbf{x}^{opt})$.

\subsubsection*{Properties}

\begin{itemize}
  \item Unimodal
  \item Non-separable
  \item Smooth local irregularities
  \item With One sensitive direction
\end{itemize}

\subsection{Shifted and Rotated Bent Cigar Function}

\begin{equation}\label{eq:cigar}
  f_6(\mathbf{x}) = \mathbf{z}_1^2+10^6\cdot{}\sum_{i=2}^D\mathbf{z}_i^2 + f_{opt}
\end{equation}
where $\mathbf{z}=\mathbf{R}(\mathbf{x}-\mathbf{x}^{opt})$.

\subsubsection*{Properties}

\begin{itemize}
  \item Unimodal
  \item Non-separable
  \item Optimum located in a smooth but very narrow valley
\end{itemize}

\subsection{Shifted and Rotated Different Powers Function}

\begin{equation}\label{eq:powers}
  f_4(\mathbf{x}) = \sqrt{ \sum_{i=1}^D|\mathbf{z}_i|^{2+4\frac{i-1}{D-1}} } + f_{opt}
\end{equation}
where $\mathbf{z}=\mathbf{R}(0.01(\mathbf{x}-\mathbf{x}^{opt}))$.

\subsubsection*{Properties}

\begin{itemize}
  \item Unimodal
  \item Non-separable
  \item Sensitivities of the $\mathbf{z}_i$-variables are different
\end{itemize}

\subsection{Shifted and Rotated Sharp Valley Function}

\begin{equation}\label{eq:sharp_valley}
  f_4(\mathbf{x}) = \mathbf{z}_i^2 + 100\cdot\sqrt{ \sum_{i=2}^D\mathbf{z}_i^2 } + f_{opt}
\end{equation}
where $\mathbf{z}=\mathbf{R}(\mathbf{x}-\mathbf{x}^{opt})$.

\subsubsection*{Properties}

\begin{itemize}
  \item Unimodal
  \item Non-separable
  \item Global optimum located in a sharp (non-differentiable) ridge
\end{itemize}

\section{Basic Multi-modal Functions}

\subsection{Shifted and Rotated Step Function}

\begin{equation}\label{eq:step}
  f_3(\mathbf{x}) = \sum_{i=1}^D\lfloor\mathbf{z}_i+0.5\rfloor^2 + f_{opt}
\end{equation}
where $\mathbf{z}=\mathbf{R}(\mathbf{x}-\mathbf{x}^{opt})$

\subsubsection*{Properties}

\begin{itemize}
  \item Many Plateaus of different sizes
  \item Non-separable
\end{itemize}

\subsection{Shifted and Rotated Weierstrass Function}

\begin{equation}\label{eq:weierstrass}
  f_9(\mathbf{x}) = \sum_{i=1}^{D}\left(  \sum_{k=0}^{k_{max}}a^k\cos{}(2\pi{}b^k(\mathbf{z}_i+0.5)) \right) - D\cdot\sum_{k=0}^{k_{max}}a^k\cos{}(2\pi{}b^k\cdot0.5) + f_{opt}
\end{equation}
where $a = 0.5$, $b=3$, $k_{max}=20$, $\mathbf{z}=\mathbf{R}(0.005\cdot(\mathbf{x}-\mathbf{x}^{opt}))$.

\subsubsection*{Properties}

\begin{itemize}
  \item Multi-modal
  \item Non-separable
  \item Continuous everywhere but only differentiable on a set of points
\end{itemize}

\subsection{Shifted and Rotated Griewank Function}

\begin{equation}\label{eq:griewank}
  f_{10}(\mathbf{x}) = \sum_{i=1}^{D}\frac{\mathbf{z}_i^2}{4000}-\prod_{i=1}^{D}\cos(\frac{\mathbf{z}_i}{\sqrt{i}}) + 1 + f_{opt}
\end{equation}
where $\mathbf{z}=\mathbf{R}(6\cdot(\mathbf{x}-\mathbf{x}^{opt}))$.

\subsubsection*{Properties}

\begin{itemize}
  \item Multi-modal
  \item Non-separable
  \item With many regularly distributed local optima
\end{itemize}

\subsection{Shifted Rastrigin Function}

\begin{equation}\label{eq:rastrigin}
  f_{11}(\mathbf{x}) = \sum_{i=1}^{D}\left(\mathbf{z}_i^2-10\cos(2\pi{}\mathbf{z}_i)\right) + 10\cdot{}D + f_{opt}
\end{equation}
where $\mathbf{z}=0.0512\cdot(\mathbf{x}-\mathbf{x}^{opt})$.

\subsubsection*{Properties}

\begin{itemize}
  \item Multi-modal
  \item Separable
  \item With many regularly distributed local optima
\end{itemize}

\subsection{Shifted and Rotated Rastrigin Function}

\begin{equation}\label{eq:lunacek}
  f_{12}(\mathbf{x}) = \sum_{i=1}^{D}\left(\mathbf{z}_i^2-10\cos(2\pi{}\mathbf{z}_i)+10\right) + f_{opt}
\end{equation}
where $\mathbf{z}=\mathbf{R}(0.0512\cdot(\mathbf{x}-\mathbf{x}^{opt}))$.

\subsubsection*{Properties}

\begin{itemize}
  \item Multi-modal
  \item Non-separable
  \item With many regularly distributed local optima
\end{itemize}

\subsection{Shifted Rotated Schaffer's F7 Function}

\begin{equation}\label{eq:schaffersf7}
  f_{17}(\mathbf{x}) = \left( \frac{1}{D-1}\sum_{i=1}^{D-1}\left((1+\sin^2(50\cdot{}\mathbf{w}_i^{0.2}))\cdot\sqrt{\mathbf{w}_i}\right) \right)^2 + f_{opt}
\end{equation}
where $\mathbf{w}_i = \sqrt{\mathbf{z}_i^2+\mathbf{z}_{i+1}^2}$, $\mathbf{z}=\mathbf{R}(\mathbf{x}-\mathbf{x}^{opt})$.

\subsubsection*{Properties}

\begin{itemize}
  \item Multi-modal
  \item Non-separable
\end{itemize}

\subsection{Expanded Griewank plus Rosenbrock Function}

$$
\begin{array}{rl}
\text{Rosenbrock Function:} & g_2(x,y) = 100(x^2-y)^2+(x-1)^2\\
\text{Griewank Function:} & g_3(x) = x^2/4000-\cos(x) + 1
\end{array}
$$

\begin{equation}\label{eq:expanded_griewank_rosenbrock}
  f_{18}(\mathbf{x}) = \sum_{i=1}^{D-1}g_3(g_2(\mathbf{z}_i,\mathbf{z}_{i+1})) + g_3(g_2(\mathbf{z}_D,\mathbf{z}_{1})) + f_{opt}
\end{equation}
where $\mathbf{z}=\mathbf{R}(0.05\cdot(\mathbf{x}-\mathbf{x}^{opt}))+\mathbf{1}$.

\subsubsection*{Properties}

\begin{itemize}
  \item Multi-modal
  \item Non-separable
  \item
\end{itemize}

\subsection{Shifted and Rotated Rosenbrock Function}

\begin{equation}\label{eq:rosenbrok}
  f_7(\mathbf{x}) = \sum_{i=1}^{D-1}\left(100\cdot(\mathbf{z}_i^2-\mathbf{z}_{i+1})^2+(\mathbf{z}_i-1)^2\right) + f_{opt}
\end{equation}
where $\mathbf{z}=\mathbf{R}(0.02048\cdot(\mathbf{x}-\mathbf{x}^{opt}))+\mathbf{1}$.

\subsubsection*{Properties}

\begin{itemize}
  \item Multi-modal
  \item Non-separable
  \item With a long, narrow, parabolic shaped flat valley from local optima to global optima
\end{itemize}

\subsection{Shifted Modified Schwefel Function}

\begin{equation}\label{eq:schwefel}
  f_{13}(\mathbf{x}) = 418.9829\times{}D - \sum_{i=1}^Dg_1(\mathbf{w}_i), \hspace{.1\textwidth}  \mathbf{w}_i = \mathbf{z}_i+420.9687462275036
\end{equation}

\begin{equation}\label{eq:basic_schwefel}
  g_1(\mathbf{w}_i) =
  \begin{cases}
  \mathbf{w}_i\cdot\sin(\sqrt{|\mathbf{w}_i|}) & \text{if } |\mathbf{w}_i| \le 500 \\
  (500-\bmod(\mathbf{w}_i,500))\cdot\sin\left(\sqrt{500-\bmod(\mathbf{w}_i,500)}\right) - \frac{(\mathbf{w}_i-500)^2}{10000D} & \text{if \ }  \mathbf{w}_i > 500 \\
  (\bmod(-\mathbf{w}_i,500)-500)\cdot\sin\left(\sqrt{500-\bmod(-\mathbf{w}_i,500)}\right) - \frac{(\mathbf{w}_i+500)^2}{10000D} & \text{if \ } \mathbf{w}_i < -500

  \end{cases}
\end{equation}
where $\mathbf{z}=10\cdot(\mathbf{x}-\mathbf{x}^{opt})$.

\subsubsection*{Properties}

\begin{itemize}
  \item Multi-modal
  \item Separable
  \item Having many local optima with the second better local optima far from the global optima
\end{itemize}

\subsection{Shifted Rotated Modified Schwefel Function}

\begin{equation}
  f_{14}(\mathbf{x}) = 418.9829\times{}D - \sum_{i=1}^Dg_1(\mathbf{w}_i), \hspace{.1\textwidth}  \mathbf{w}_i = \mathbf{z}_i+420.9687462275036
\end{equation}

where $\mathbf{z}=\mathbf{R}(10\cdot(\mathbf{x}-\mathbf{x}^{opt}))$ and $g_1(\cdot)$ is defined as Eq.\,\ref{eq:basic_schwefel}.

\subsubsection*{Properties}

\begin{itemize}
  \item Multi-modal
  \item Non-separable
  \item Having many local optima with the second better local optima far from the global optima
\end{itemize}

\subsection{Shifted Rotated Katsuura Function}

\begin{equation}\label{eq:katsuura}
  f_{15}(\mathbf{x}) = \frac{10}{D^2} \prod_{i=1}^{D}(1+i\sum_{j=1}^{32}\frac{|2^j\cdot\mathbf{z}_i-[2^j\cdot\mathbf{z}_i]|}{2^j})^{\frac{10}{D^{1.2}}} - \frac{10}{D^2}  + f_{opt}
\end{equation}
where $\mathbf{z}=\mathbf{R}(0.05\cdot(\mathbf{x}-\mathbf{x}^{opt}))$.

\subsubsection*{Properties}

\begin{itemize}
  \item Multi-modal
  \item Non-separable
  \item Continuous everywhere but differentiable nowhere
\end{itemize}

\subsection{Shifted and Rotated Lunacek bi-Rastrigin Function}

\begin{equation}\label{eq:rotated_rastrigin}
  f_{12}(\mathbf{x}) = \min\left( \sum_{i=1}^{D}(\mathbf{z}_i-\mu_1)^2, dD+s\sum_{i=1}^{D}(\mathbf{z}_i-\mu_2)^2)  \right) + 10\cdot(D-\sum_{i=1}^{D}\cos(2\pi{}(\mathbf{z}_i-\mu_1))) + f_{opt}
\end{equation}
where $\mathbf{z}=\mathbf{R}(0.1\cdot(\mathbf{x}-\mathbf{x}^{opt})+2.5*\mathbf{1})$, $\mu_1=2.5$, $\mu_2=-2.5$, $d=1$, $s=0.9$.

\subsubsection*{Properties}

\begin{itemize}
  \item Multi-modal
  \item Non-separable
  \item With two funnel around $\mu_1\mathbf{1}$ and $\mu_2\mathbf{1}$
\end{itemize}

\subsection{Shifted and Rotated Ackley Function}

\begin{equation}\label{eq:ackley}
  f_8(\mathbf{x}) = -20\cdot\exp\left(-0.2\sqrt{\frac{1}{D}\sum_{i=1}^D\mathbf{x}_i^2}\right) - \exp\left(  \frac{1}{D}\sum_{i=1}^D \cos(2\pi{}\mathbf{x}_i)\right) + 20 + e
   + f_{opt}
\end{equation}
where $\mathbf{z}=\mathbf{R}(\mathbf{x}-\mathbf{x}^{opt})$.

\subsubsection*{Properties}

\begin{itemize}
  \item Multi-modal
  \item Non-separable
  \item Having many local optima with the global optima located in a very small basin
\end{itemize}

\subsection{Shifted Rotated HappyCat Function}

\begin{equation}\label{eq:happycat}
  f_{16}(\mathbf{x}) = |\sum_{i=1}^D\mathbf{z}_i^2-D|^{0.25} + (\frac{1}{2}\sum_{j=1}^D\mathbf{z}_j^2+\sum_{j=1}^D\mathbf{z}_j)/D + 0.5 + f_{opt}
\end{equation}
where $\mathbf{z}=\mathbf{R}(0.05\cdot(\mathbf{x}-\mathbf{x}^{opt}))-\mathbf{1}$.

\subsubsection*{Properties}

\begin{itemize}
  \item Multi-modal
  \item Non-separable
  \item Global optima located in curved narrow valley
\end{itemize}

\subsection{Shifted Rotated HGBat Function}

\begin{equation}\label{eq:hgbat}
  f_{17}(\mathbf{x}) = |(\sum_{i=1}^D\mathbf{z}_i^2)^2-(\sum_{j=1}^D\mathbf{z}_j)^2|^{0.5} + (\frac{1}{2}\sum_{j=1}^D\mathbf{z}_j^2+\sum_{j=1}^D\mathbf{z}_j)/D + 0.5 + f_{opt}
\end{equation}
where $\mathbf{z}=\mathbf{R}(0.05\cdot(\mathbf{x}-\mathbf{x}^{opt}))-\mathbf{1}$.

\subsubsection*{Properties}

\begin{itemize}
  \item Multi-modal
  \item Non-separable
  \item Global optima located in curved narrow valley
\end{itemize}

\subsection{Expanded Schaffer's F6 Function}
$$
\text{Schaffer's F6 Function: }g_4(x,y)=\frac{\sin^2(\sqrt{x^2+y^2})-0.5}{(1+0.001\cdot(x^2+y^2))^2} + 0.5
$$
\begin{equation}\label{eq:expanded_schaffersf6}
  f_{19}(\mathbf{x}) = \sum_{i=1}^{D-1}g_4(\mathbf{z}_i,\mathbf{z}_{i+1}) + g_4(\mathbf{z}_D,\mathbf{z}_1) + f_{opt}
\end{equation}
where $\mathbf{z}=\mathbf{R}(\mathbf{x}-\mathbf{x}^{opt})$.

\subsubsection*{Properties}

\begin{itemize}
  \item Multi-modal
  \item Non-separable
\end{itemize}

\section{Hybrid Functions}

Hybrid functions are constructed according to \cite{CEC14}.
For each hybrid function, the variables are randomly divided into subcomponents and different basic functions (unimodal and multi-modal) are used for different subcomponents, as depicted by Eq.\,\ref{eq:hybrid}.
\begin{equation}\label{eq:hybrid}
  F(\mathbf{x}) = \sum_{i=1}^{N}G_i(\mathbf{R}_i\cdot\mathbf{z}^i) + f^{opt}
\end{equation}
where $F(\cdot)$ is the constructed hybrid function and $G_i(\cdot)$ is the $i$-th basic function used, $N$ is the number of basic functions.
$\mathbf{z}_i$ is constructed as follows.
$$
\begin{array}{rl}
\mathbf{y}\quad = & \mathbf{x}-\mathbf{x}^{opt}\\
\mathbf{z}^1\quad = & [\mathbf{y}_{S_1},\mathbf{y}_{S_2},\dots,\mathbf{y}_{S_{n_{1}}}]\\
\mathbf{z}^2\quad =  & [\mathbf{y}_{S_{n_{1}+1}},\mathbf{y}_{S_{n_{1}+2}} \cdots \mathbf{y}_{S_{n_{1}+n2}}]\\
 & \hspace{ 2cm } \vdots\\
\mathbf{z}^N\quad = & [\mathbf{y}_{S_{(\sum_{i=1}^{N-1}n_i) + 1}},\mathbf{y}_{S_{(\sum_{i=1}^{N-1}n_i)+2}},\dots,\mathbf{y}_{S_{n_D}}]
\end{array}
$$
where $S$ is a permutation of $(1:D)$, such that $\mathbf{z} = [\mathbf{z}^1, \mathbf{z}^2,\dots,\mathbf{z}^N]$ forms the transformed vector and $n_i, i=1,\dots,N$ are the dimensions of the basic functions, which is derived as Eq.\,\ref{eq:numbers}.
 \begin{equation}\label{eq:numbers}
   n_i = \lceil p_iD\rceil (i = 1,2,\cdots,N-1), n_N = D-\sum_{i=1}^{N-1}n_i
 \end{equation}
$p_i$ is used to control the percentage of each basic functions.

\subsection{Hybrid Function 1}
\begin{itemize}
  \item $N = 3$
  \item $p = [0.3,0.3,0.4]$
  \item $G_1$: Modified Schwefel's Function
  \item $G_2$: Rastrigin Function
  \item $G_3$: High Conditioned Elliptic Function
\end{itemize}

\subsection{Hybrid Function 2}
\begin{itemize}
  \item $N = 5$
  \item $p = [0.3,0.3,0.4]$
  \item $G_1$: Bent Cigar Function
  \item $G_2$: HGBat Function
  \item $G_3$: Rastrigin Function
\end{itemize}

\subsection{Hybrid Function 3}
\begin{itemize}
  \item $N = 4$
  \item $p = [0.2,0.2,0.3,0.3]$
  \item $G_1$: Griewank Function
  \item $G_2$: Weierstrass Function
  \item $G_3$: Rosenbrock Function
  \item $G_4$: Expanded Scaffer's F6 Function
\end{itemize}

\subsection{Hybrid Function 4}
\begin{itemize}
  \item $N = 4$
  \item $p = [0.2,0.2,0.3,0.3]$
  \item $G_1$: HGBat Function
  \item $G_2$: Discus Function
  \item $G_3$: Expanded Griewank plus Rosenbrock Function
  \item $G_4$: Rastrigin Function
\end{itemize}

\subsection{Hybrid Function 5}
\begin{itemize}
  \item $N = 5$
  \item $p = [0.1,0.2,0.2,0.2,0.3]$
  \item $G_1$: Expanded Scaffer's F6 Function
  \item $G_2$: HGBat Function
  \item $G_3$: Rosenbrock Function
  \item $G_4$: Modified Schwefel's Function
  \item $G_5$: High Conditioned Elliptic Function
\end{itemize}

\subsection{Hybrid Function 6}
\begin{itemize}
  \item $N = 5$
  \item $p = [0.1,0.2,0.2,0.2,0.3]$
  \item $G_1$: Katsuura Function
  \item $G_2$: HappyCat Function
  \item $G_3$: Expanded Griewank plus Rosenbrock Function
  \item $G_4$: Modified Schwefel's Function
  \item $G_5$: Ackley Function
\end{itemize}

\section{Composition Functions}

Composition functions are constructed in the same manner as in \cite{CEC13,CEC14}.

\begin{equation}\label{eq:composition}
  F(\mathbf{x}) = \sum_{i=1}^{N}\left[\omega_i*(\lambda\cdot{}G_i(\mathbf{x})+bias_i)\right] + f^{opt}
\end{equation}

\begin{itemize}
  \item $F(\mathbf{\cdot})$: the constructed composition function
  \item $G_i(\cdot)$: $i$-th basic function
  \item $N$: number of basic functions used
  \item $bias_i$: define which optimum is the global optimum
  \item $\sigma_i$: control $G_i(\cdot)$'s coverage range, a small $\sigma_i$ gives a narrow range for $G_i(\cdot)$
  \item $\lambda_i$: control $G_i(\cdot)$'s height
  \item $\omega_i$: weighted value for $G_i(\cdot)$, calculated as follows:
  \begin{equation}\label{eq:weight}
    w_i = \frac{1}{\sqrt{\sum_{j=1}^D(\mathbf{x}_j-\mathbf{x}_j^{opt,i})^2}}\exp(    -\frac{\sum_{j=1}^D(\mathbf{x}_j-\mathbf{x}_j^{opt,i})^2}{2D\sigma_i^2}        )
  \end{equation}
  where $\mathbf{x}^{opt,i}$ represents the optimum position for $G_i(\cdot)$.
  Then normalized $w_i$ to get $\omega_i$:
  $\omega_i = w_i/\sum_{i=1}^Nw_i$.

  When $\mathbf{x} = \mathbf{x}^{opt,i}$ , $\omega_j=\left\{\begin{aligned}1\quad& j=i\\0\quad&j \ne i\end{aligned}\right. \quad(j = 1,2,\cdots,N)$,
  such that $F(\mathbf{x})=bias_i+f^{opt,i}$.

\end{itemize}

The constructed functions are multi-modal and non-separable and merge the properties of the sub-functions better and maintains continuity around the global/local optima.
The local optimum which has the smallest bias value is the global optimum.
The optimum of the third basic function is set to the origin as a trip in order to test the algorithms' tendency to converge to the search center.

Note that, the landscape is not only changes along with the selection of basic function, but the optima and $\sigma$ and $\lambda$ can effect it greatly.

\subsection{Composition Function 1}

\begin{itemize}
  \item $N = 5$
  \item $\sigma = [10,20,30,40,50]$
  \item $\lambda = [1e-10, 1e-6,1e-26,1e-6,1e-6]$
  \item $bias=[0,100,200,300,400]$
  \item $G_1$: Rotated Rosenbrock Function
  \item $G_2$: High Conditioned Elliptic Function
  \item $G_3$: Rotated Bent Cigar Function
  \item $G_4$: Rotated Discus Function
  \item $G_5$: High Conditioned Elliptic Function
\end{itemize}

\subsection{Composition Function 2}

\begin{itemize}
  \item $N = 3$
  \item $\sigma = [15, 15, 15]$
  \item $\lambda =  [1,1,1] $
  \item $bias=[0,100,200]$
  \item $G_1$: Expanded Schwefel Function
  \item $G_2$: Rotated Rstrigin Function
  \item $G_3$: Rotated HGBat Function
\end{itemize}

\subsection{Composition Function 3}

\begin{itemize}
  \item $N = 3$
  \item $\sigma = [20,50,40]$
  \item $\lambda = [0.25, 1, 1e-7]$
  \item $bias=[0,100,200]$
  \item $G_1$: Rotated Schwefel Function
  \item $G_2$: Rotated Rastrigin Function
  \item $G_3$: Rotated High Conditioned Elliptic Function
\end{itemize}

\subsection{Composition Function 4}

\begin{itemize}
  \item $N = 5$
  \item $\sigma = [20,15,10,10,40]$
  \item $\lambda = [2.5e-2,0.1,1e-8,0.25,1]$
  \item $bias=[0,100,200,300,400]$
  \item $G_1$: Rotated Schwefel Function
  \item $G_2$: Rotated HappyCat Function
  \item $G_3$: Rotated High Conditioned Elliptic Function
  \item $G_4$: Rotated Weierstrass Function
  \item $G_5$: Rotated Griewank Function
\end{itemize}

\subsection{Composition Function 5}

\begin{itemize}
  \item $N = 5$
  \item $\sigma = [15,15,15,15,15]$
  \item $\lambda = [10,10,2.5,2.5,1e-6]$
  \item $bias=[0,100,200,300,400]$
  \item $G_1$: Rotated HGBat Function
  \item $G_2$: Rotated Rastrigin Function
  \item $G_5$: Rotated Schwefel Function
  \item $G_4$: Rotated Weierstrass Function
  \item $G_3$: Rotated High Conditioned Elliptic Function
\end{itemize}

\subsection{Composition Function 6}

\begin{itemize}
  \item $N = 5$
  \item $\sigma = [10,20,30,40,50]$
  \item $\lambda = [2.5,10,2.5,5e-4,1e-6]$
  \item $bias=[0,100,200,300,400]$
  \item $G_1$: Rotated Expanded Griewank plus Rosenbrock Function
  \item $G_2$: Rotated HappyCat Function
  \item $G_3$: Rotated Schwefel Function
  \item $G_4$: Rotated Expanded Scaffer's F6 Function
  \item $G_5$: High Conditioned Elliptic Function
\end{itemize}

\subsection{Composition Function 7}

\begin{itemize}
  \item $N = 3$
  \item $\sigma = [10,30,50]$
  \item $\lambda = [1, 1, 1]$
  \item $bias=[0,100,200]$
  \item $G_1$: Hybrid Function 1
  \item $G_2$: Hybrid Function 2
  \item $G_3$: Hybrid Function 3
\end{itemize}

\subsection{Composition Function 8}

\begin{itemize}
  \item $N = 3$
  \item $\sigma = [10,30,50]$
  \item $\lambda = [1, 1, 1]$
  \item $bias=[0,100,200]$
  \item $G_1$: Hybrid Function 4
  \item $G_2$: Hybrid Function 5
  \item $G_3$: Hybrid Function 6
\end{itemize}

\newpage

\appendix

\section*{Appendices}



\section{Figures for 2-D Functions}

\begin{figure}[h]
  \centering
  \includegraphics[width=.5\textwidth]{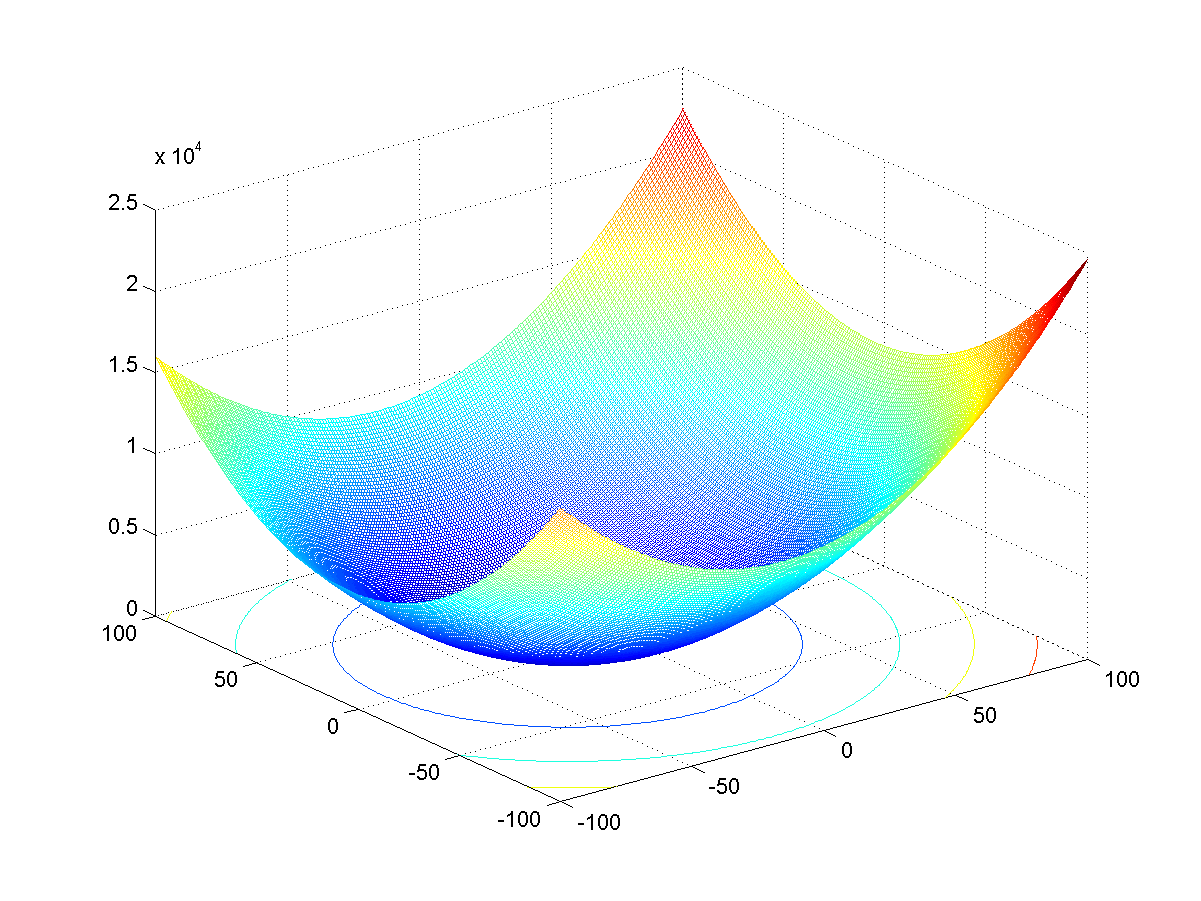}\includegraphics[width=.5\textwidth]{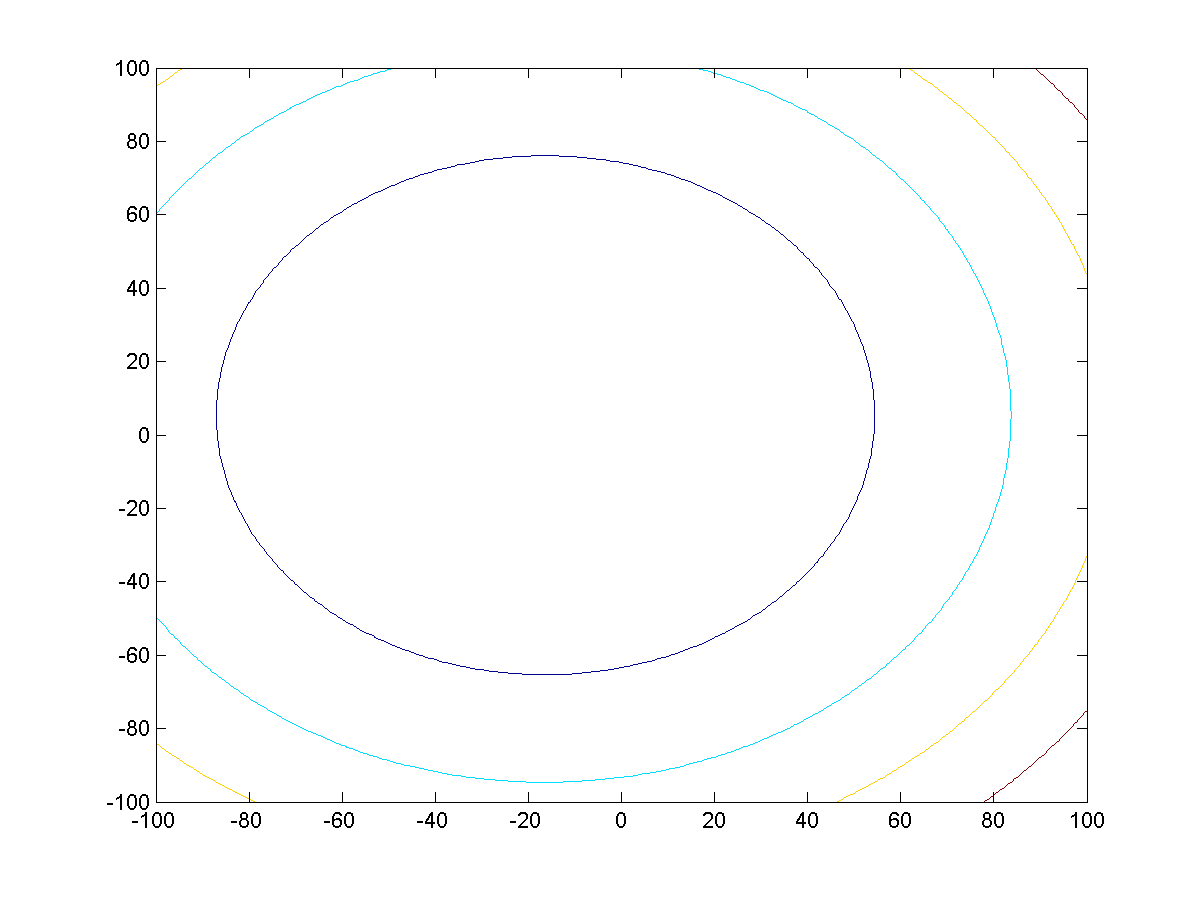}\\
  \caption{Sphere Function}\label{fig:sphere}
\end{figure}

\begin{figure}[h]
  \centering
  \includegraphics[width=.5\textwidth]{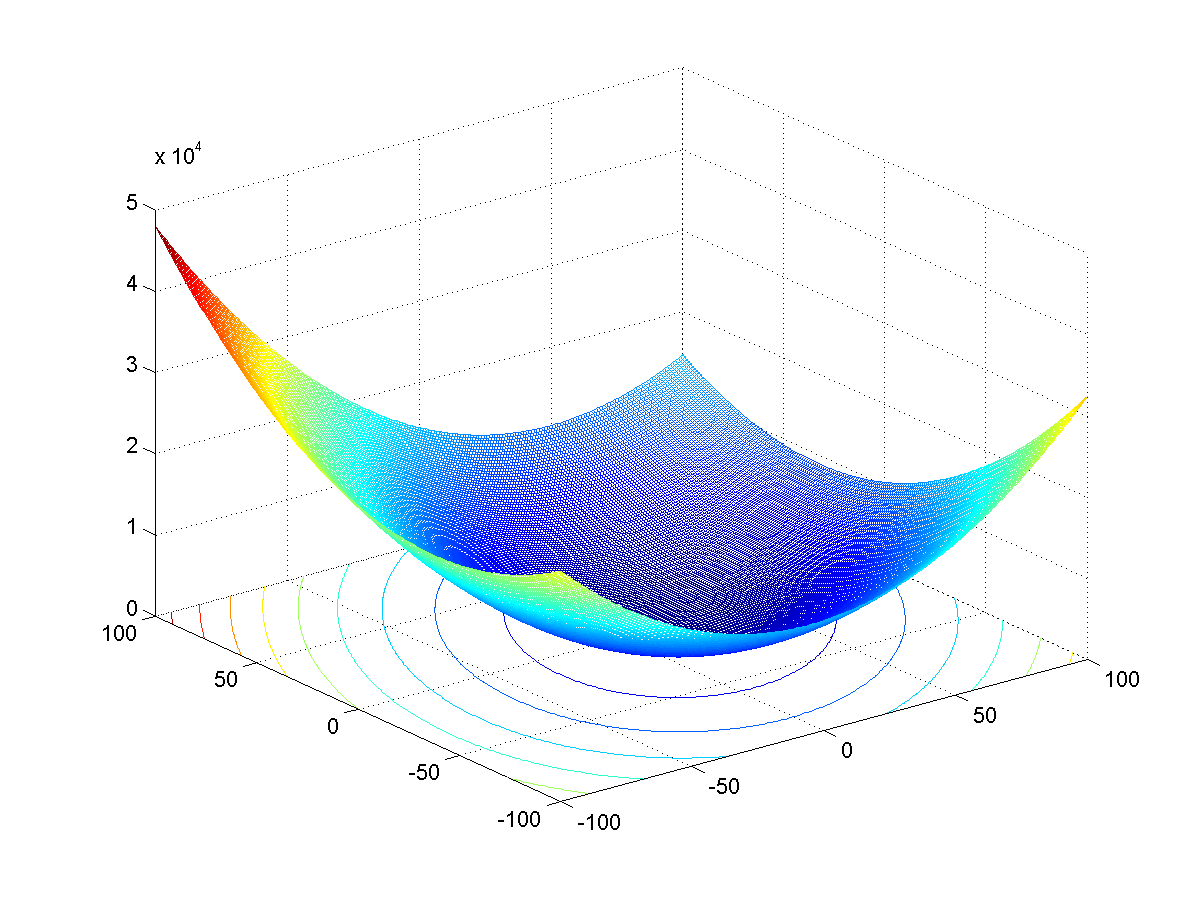}\includegraphics[width=.5\textwidth]{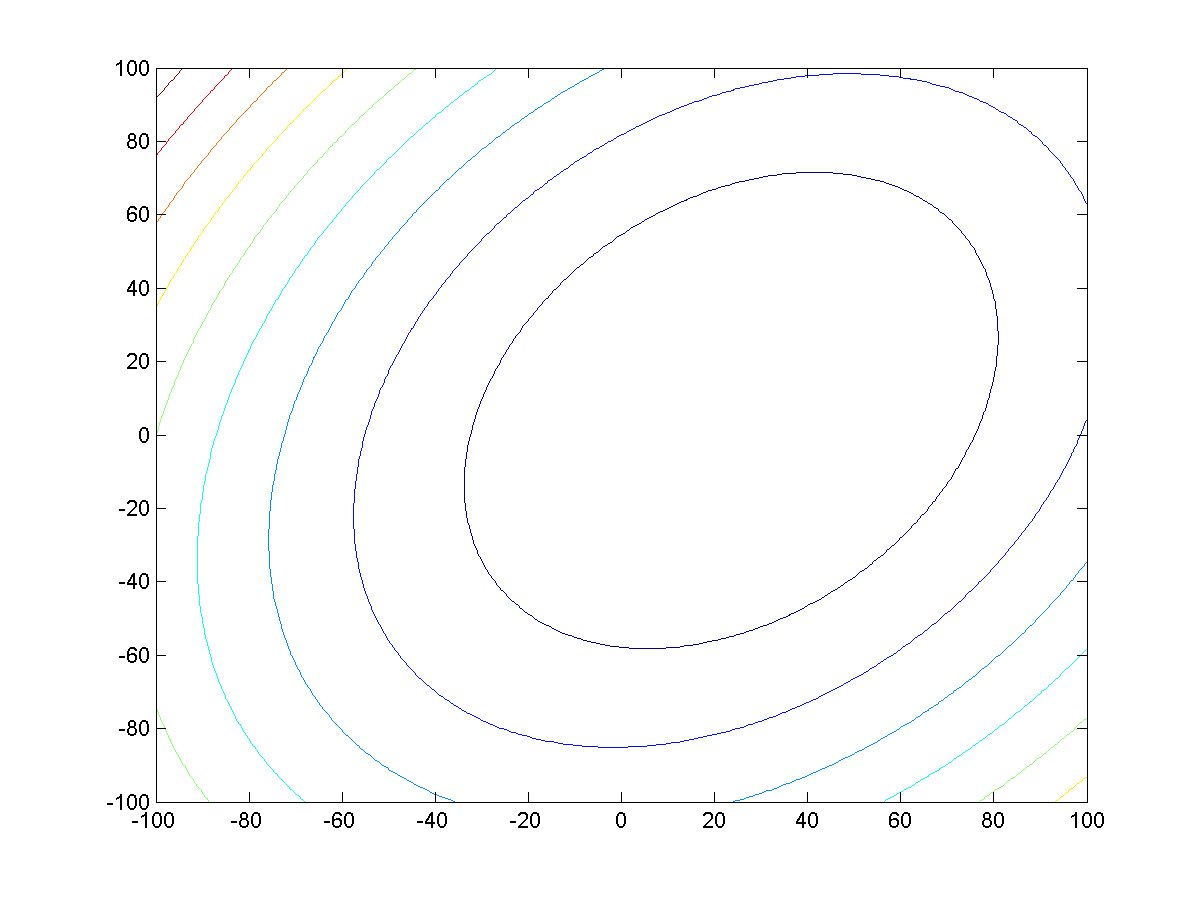}\\
  \caption{Ellipsoid Function}\label{fig:ellipsoid}
\end{figure}

\begin{figure}[tbp]
  \centering
  \includegraphics[width=.5\textwidth]{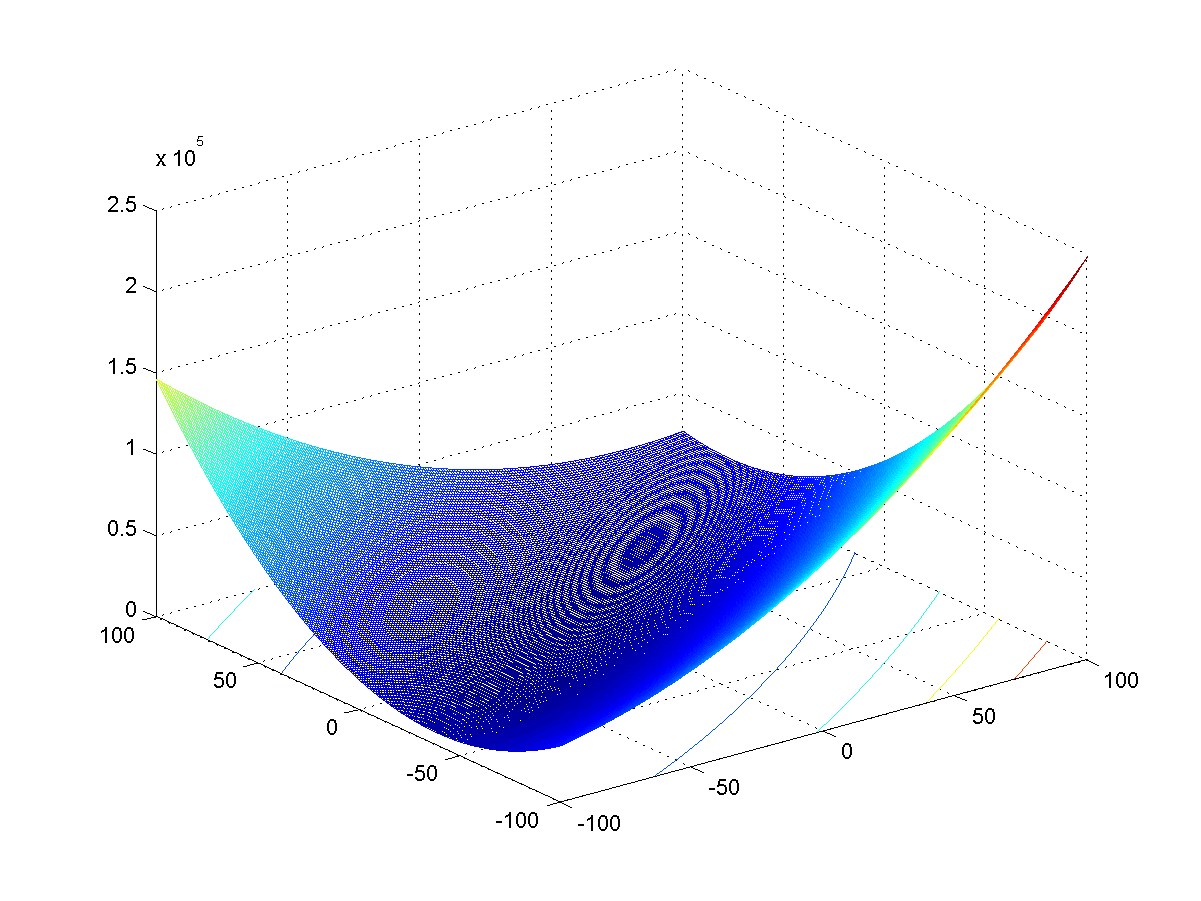}\includegraphics[width=.5\textwidth]{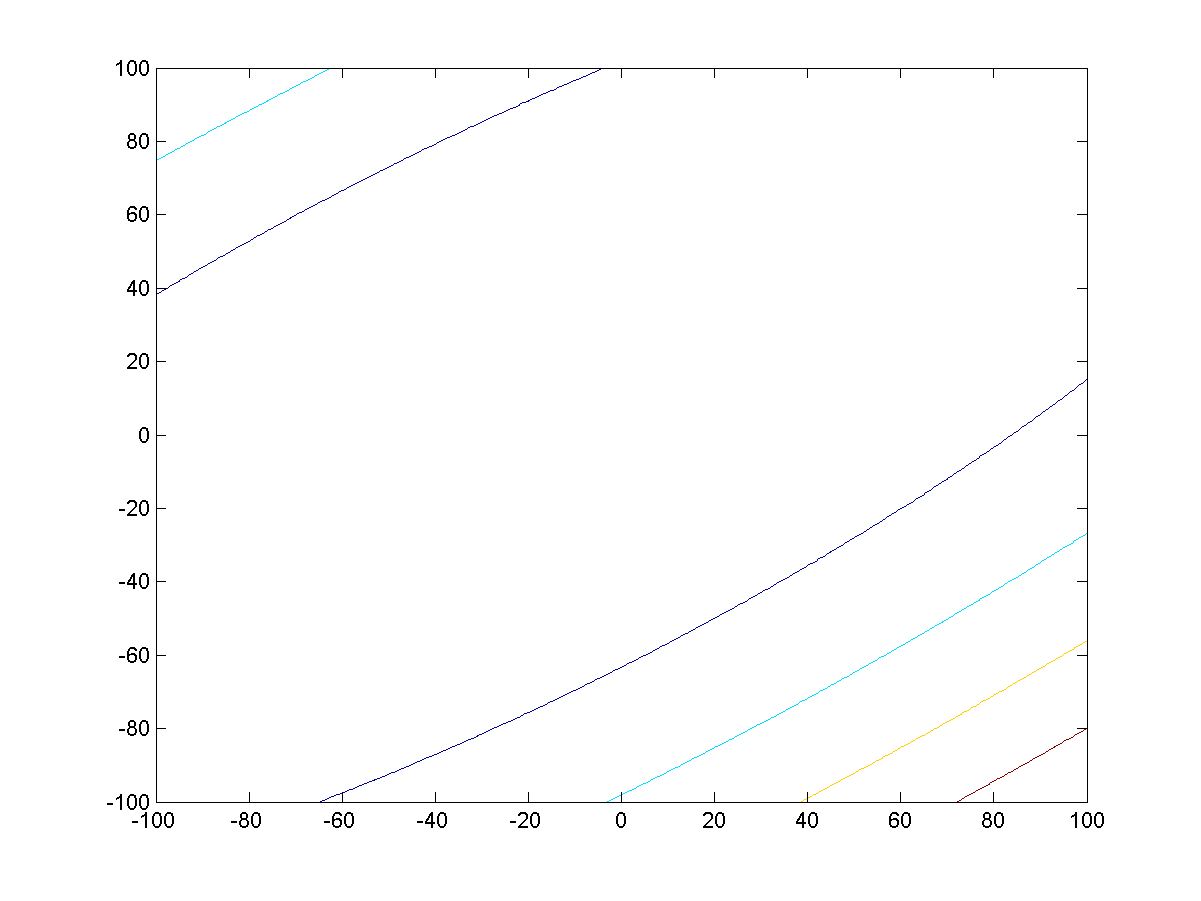}\\
  \caption{Elliptic Function}\label{fig:elliptic}
\end{figure}

\begin{figure}[tbp]
  \centering
  \includegraphics[width=.5\textwidth]{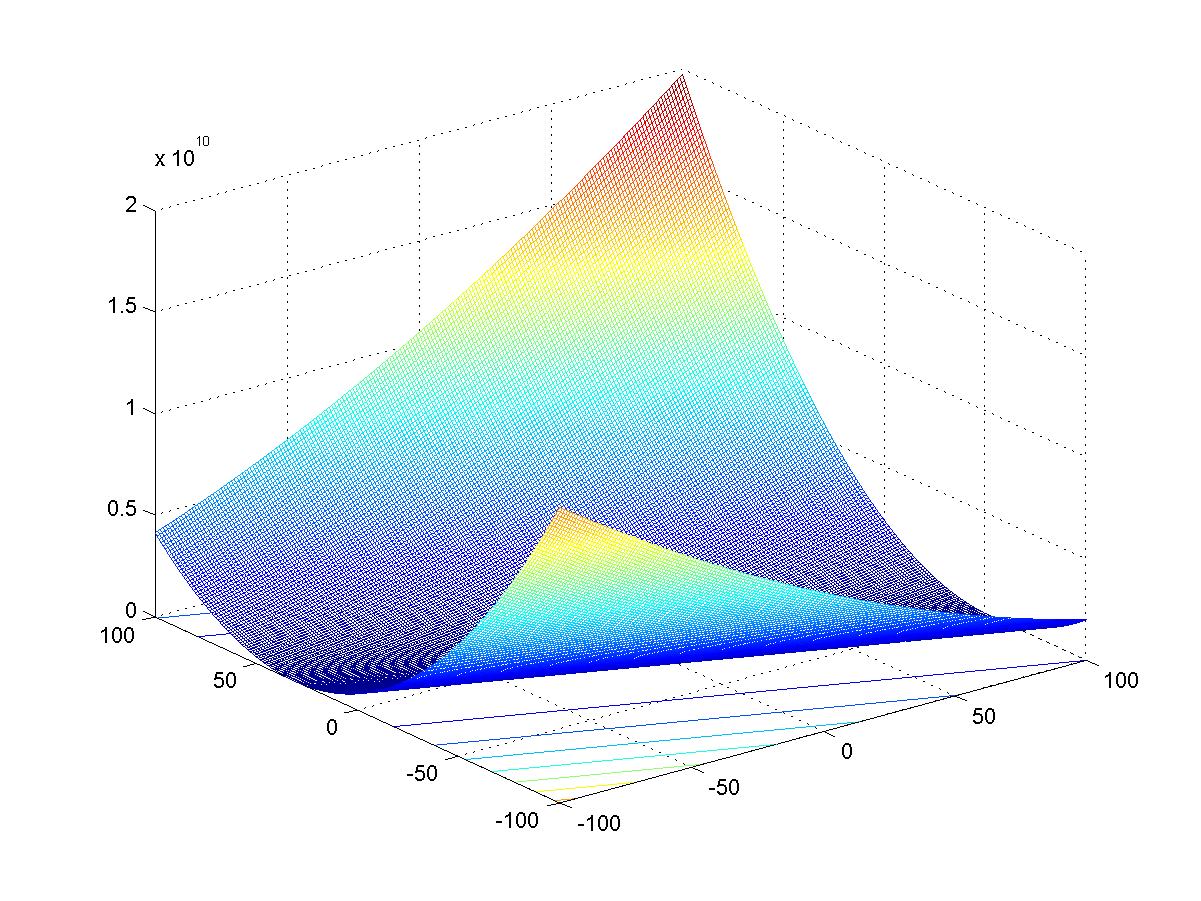}\includegraphics[width=.5\textwidth]{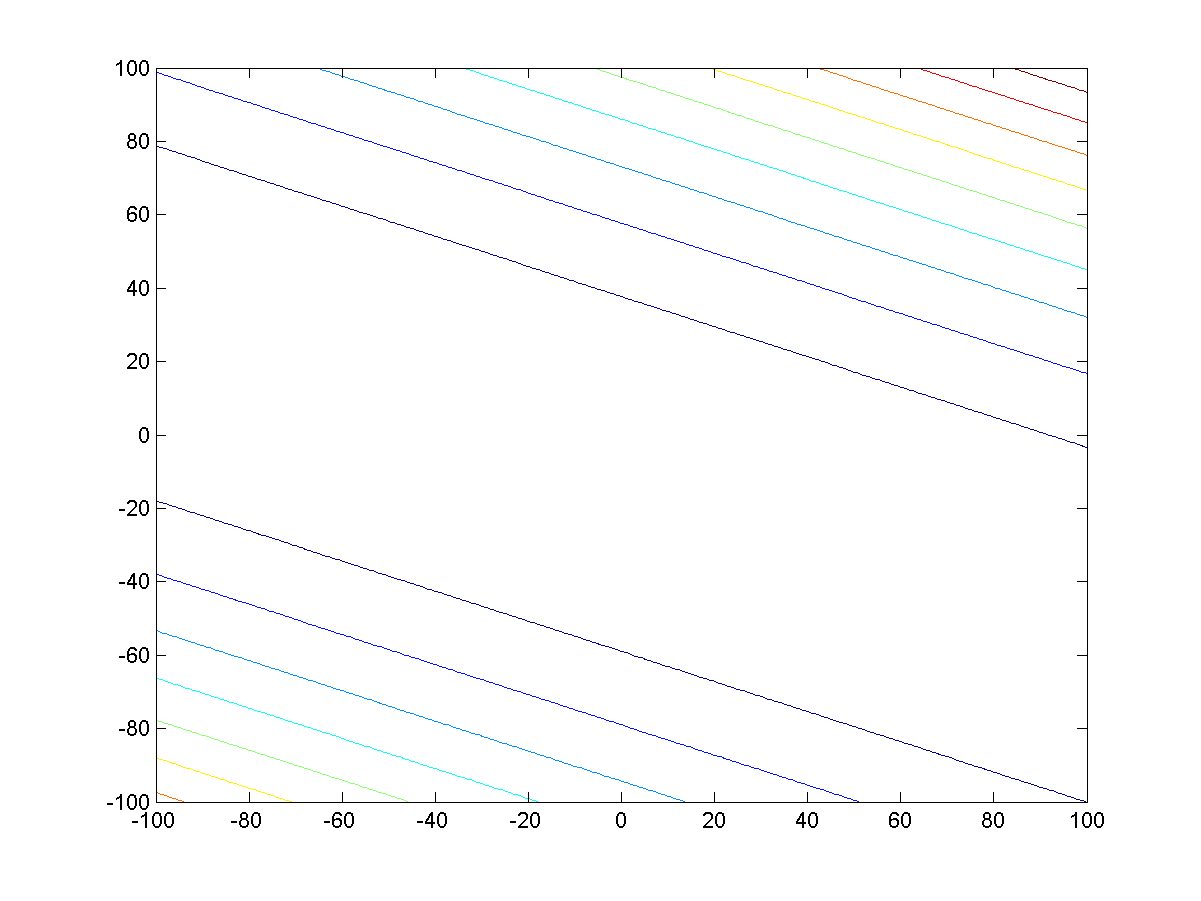}\\
  \caption{Discus Function}\label{fig:discu}
\end{figure}

\begin{figure}[tbp]
  \centering
  \includegraphics[width=.5\textwidth]{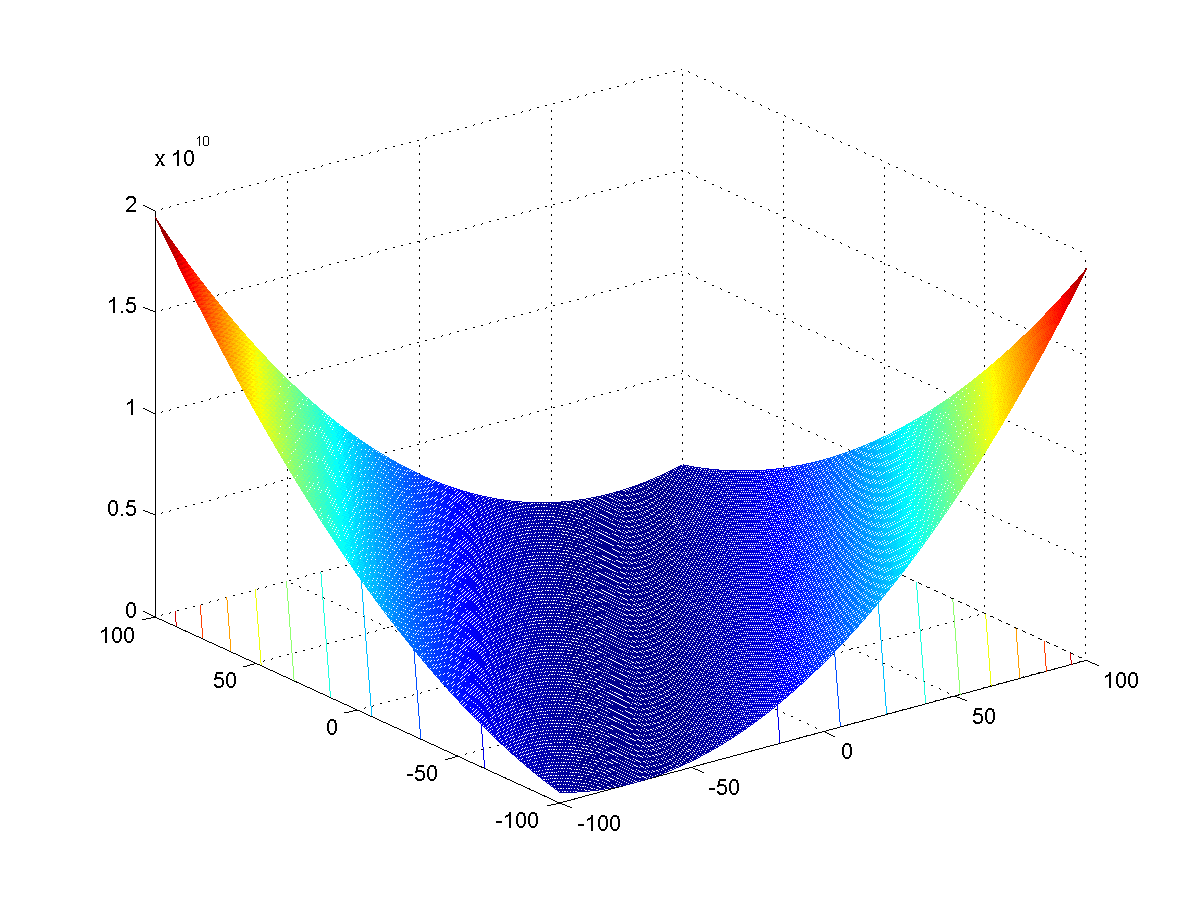}\includegraphics[width=.5\textwidth]{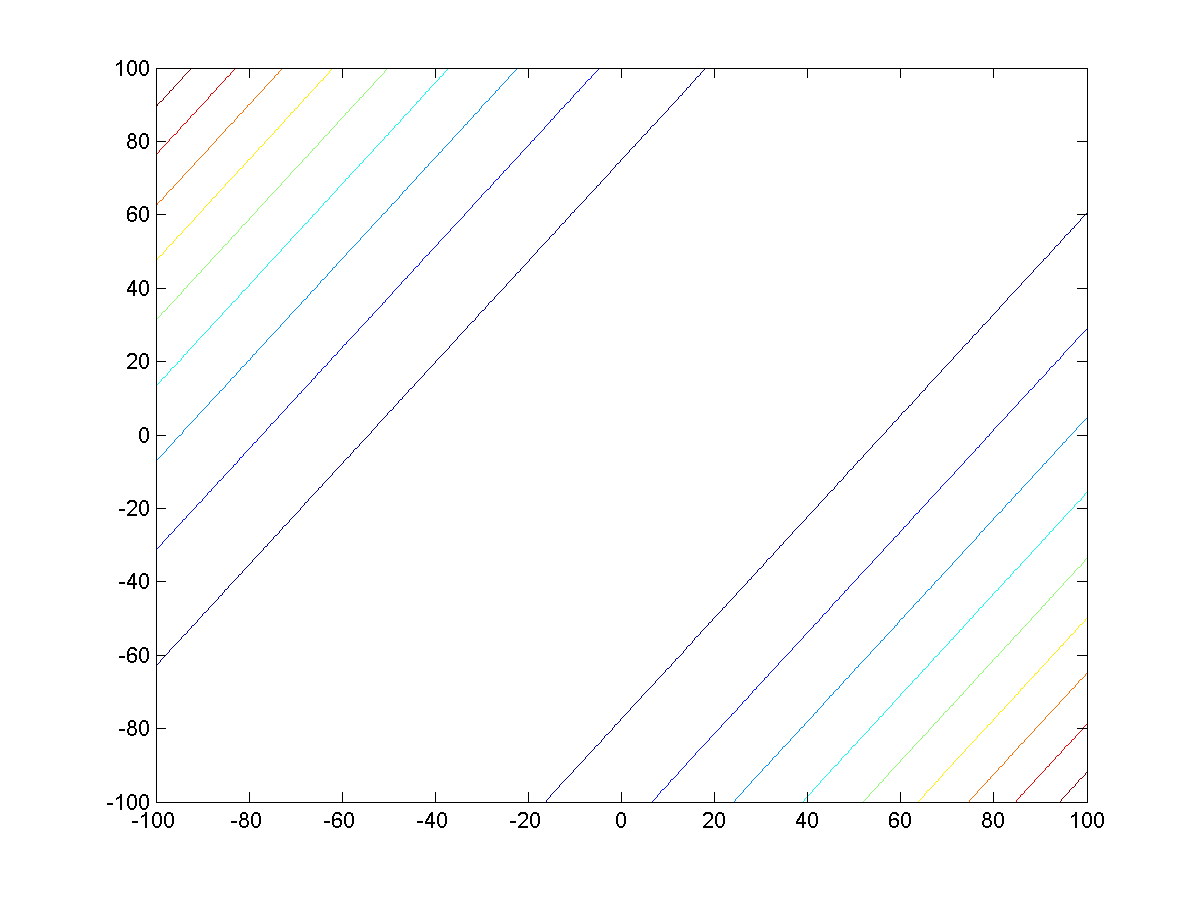}\\
  \caption{Bent Cigar Function}\label{fig:cigar}
\end{figure}

\begin{figure}[tbp]
  \centering
  \includegraphics[width=.5\textwidth]{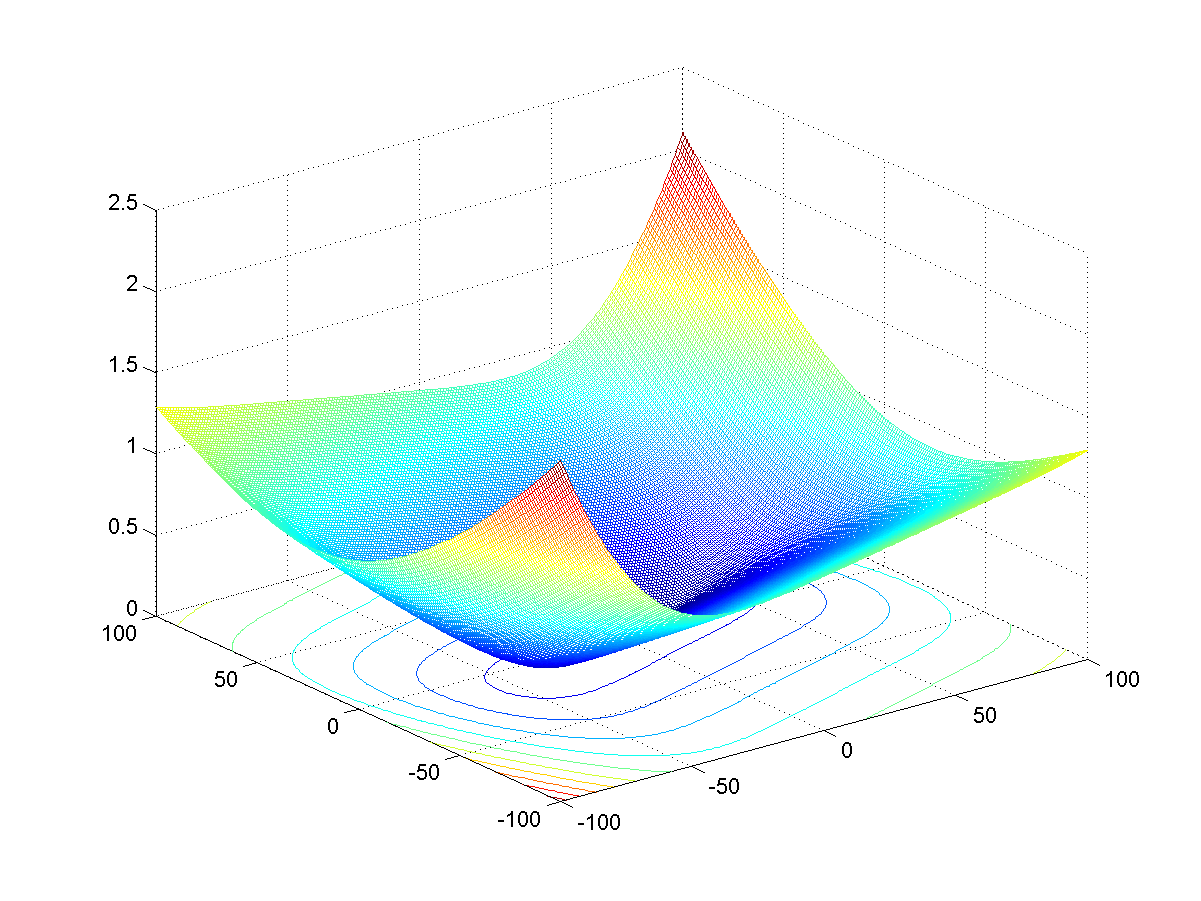}\includegraphics[width=.5\textwidth]{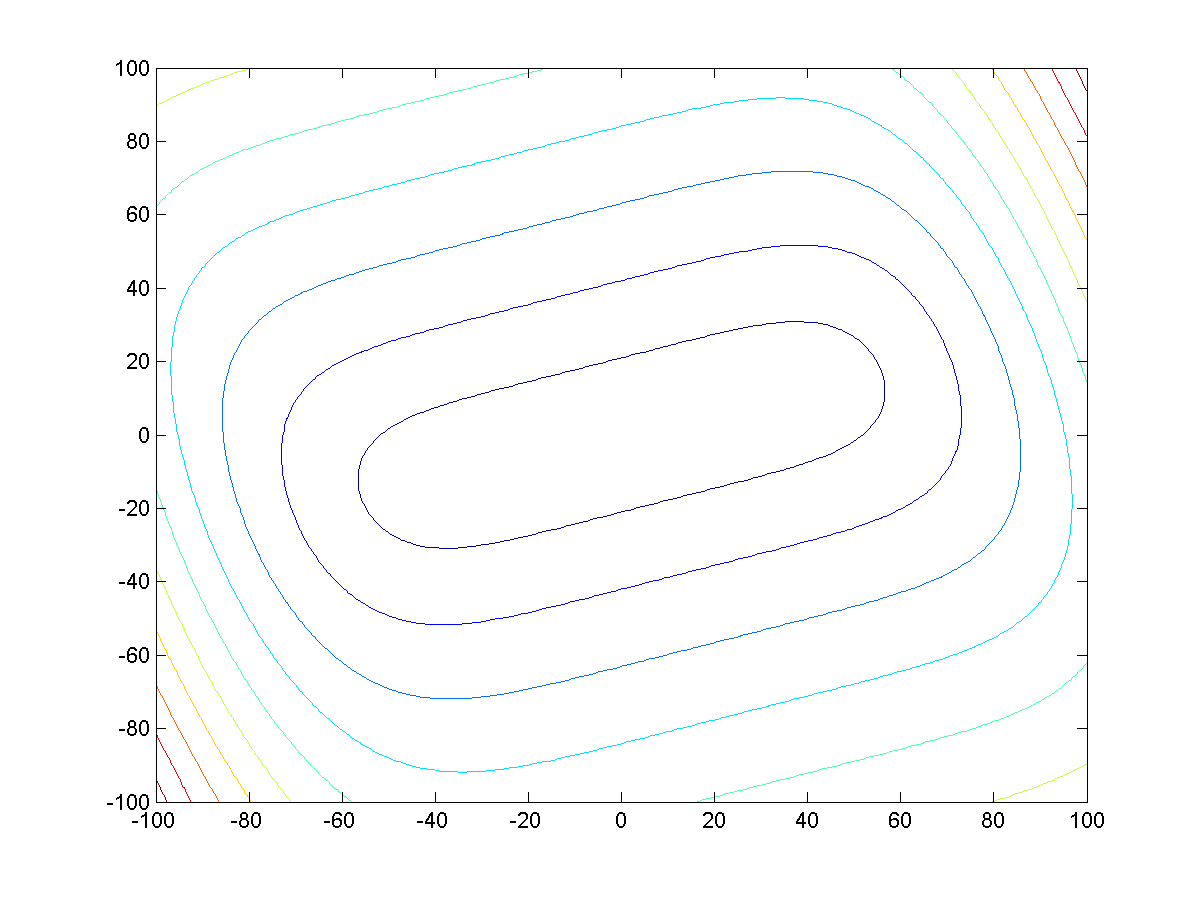}\\
  \caption{Different Powers Function}\label{fig:powers}
\end{figure}

\begin{figure}[tbp]
  \centering
  \includegraphics[width=.5\textwidth]{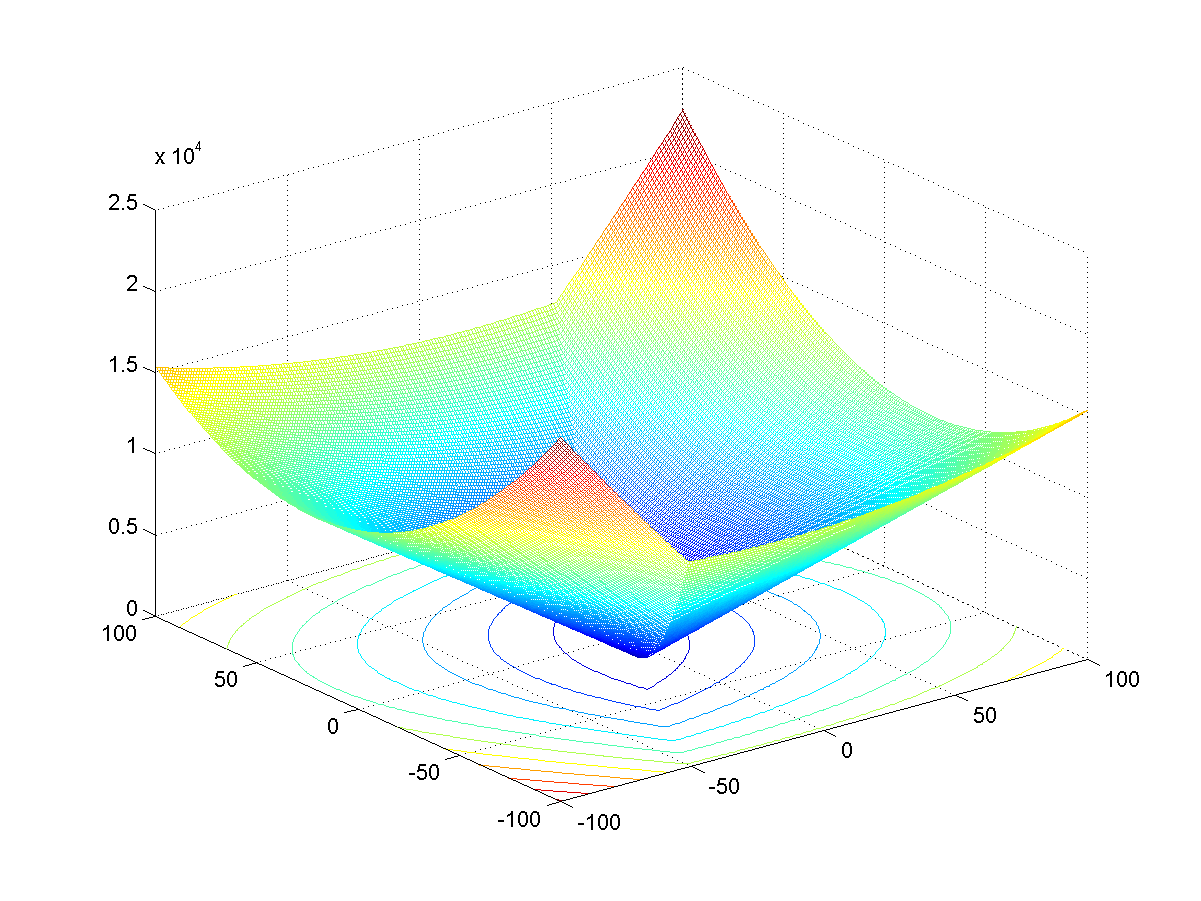}\includegraphics[width=.5\textwidth]{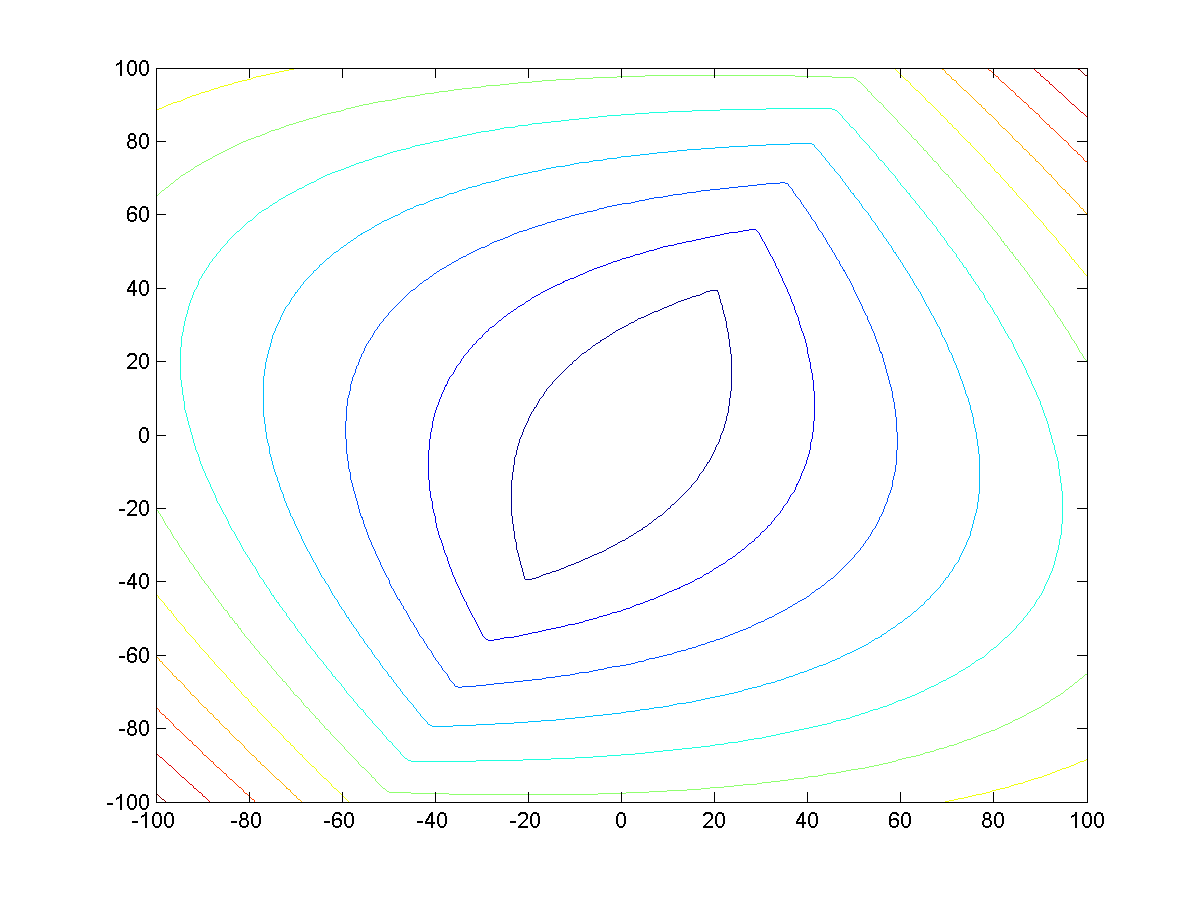}\\
  \caption{Sharp Valley Function}\label{fig:valley}
\end{figure}

\begin{figure}[tbp]
  \centering
  \includegraphics[width=.5\textwidth]{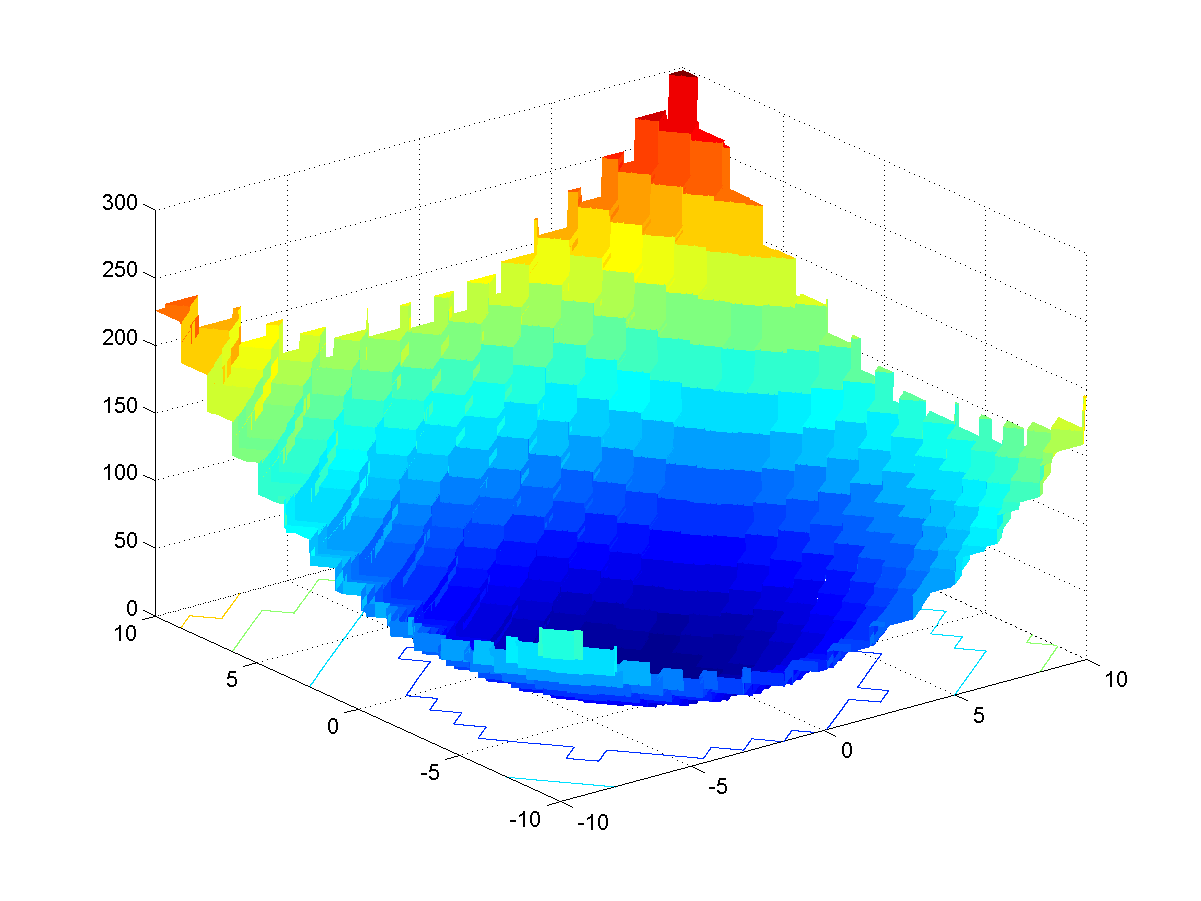}\includegraphics[width=.5\textwidth]{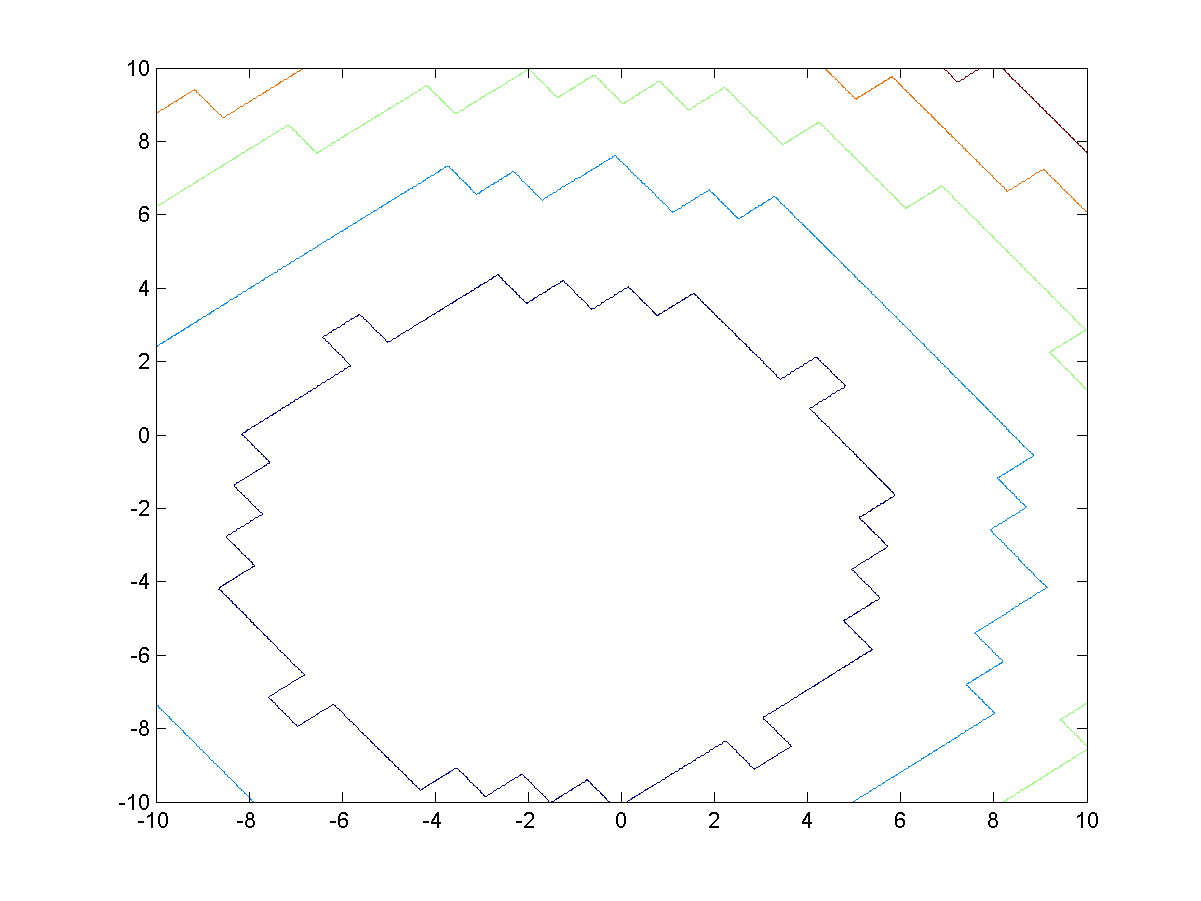}\\
  \caption{Step Function}\label{fig:step}
\end{figure}

\begin{figure}[tbp]
  \centering
  \includegraphics[width=.5\textwidth]{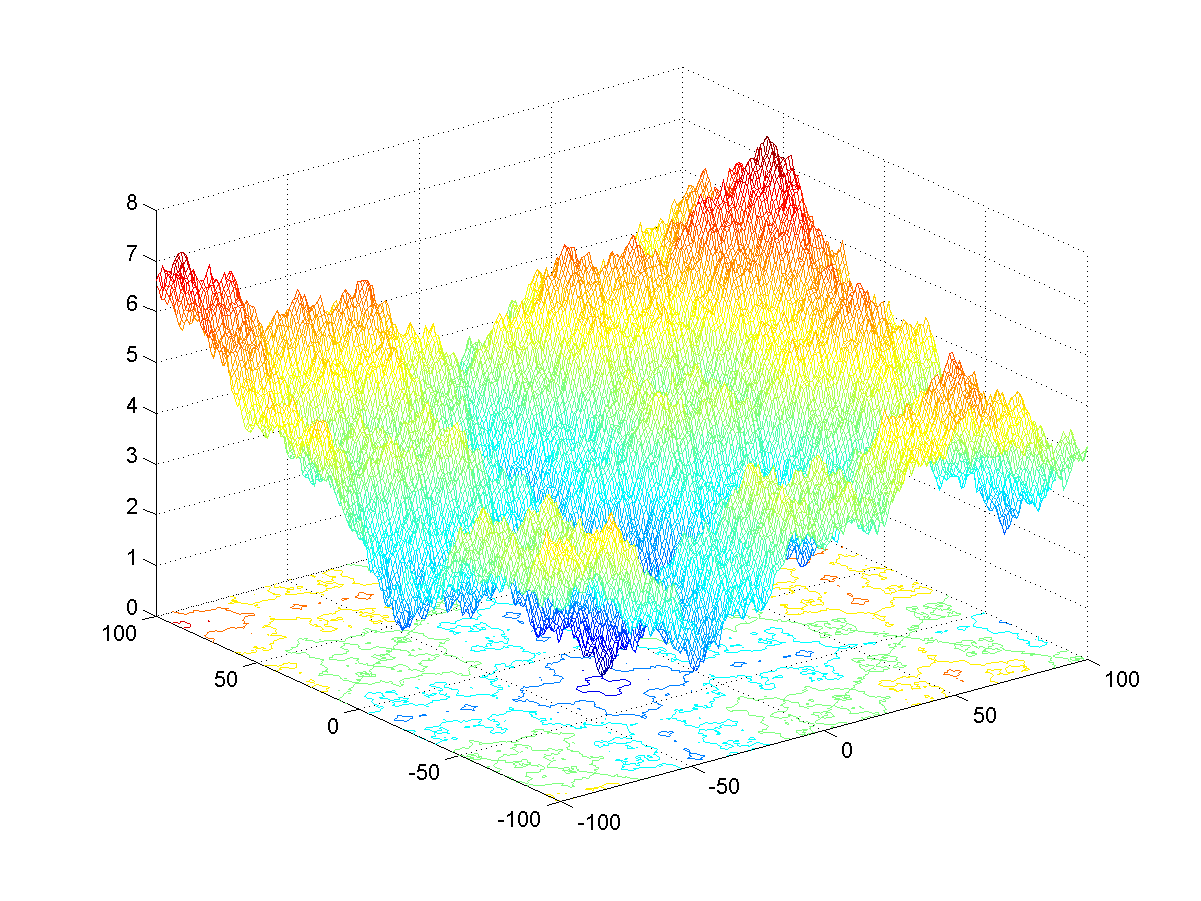}\includegraphics[width=.5\textwidth]{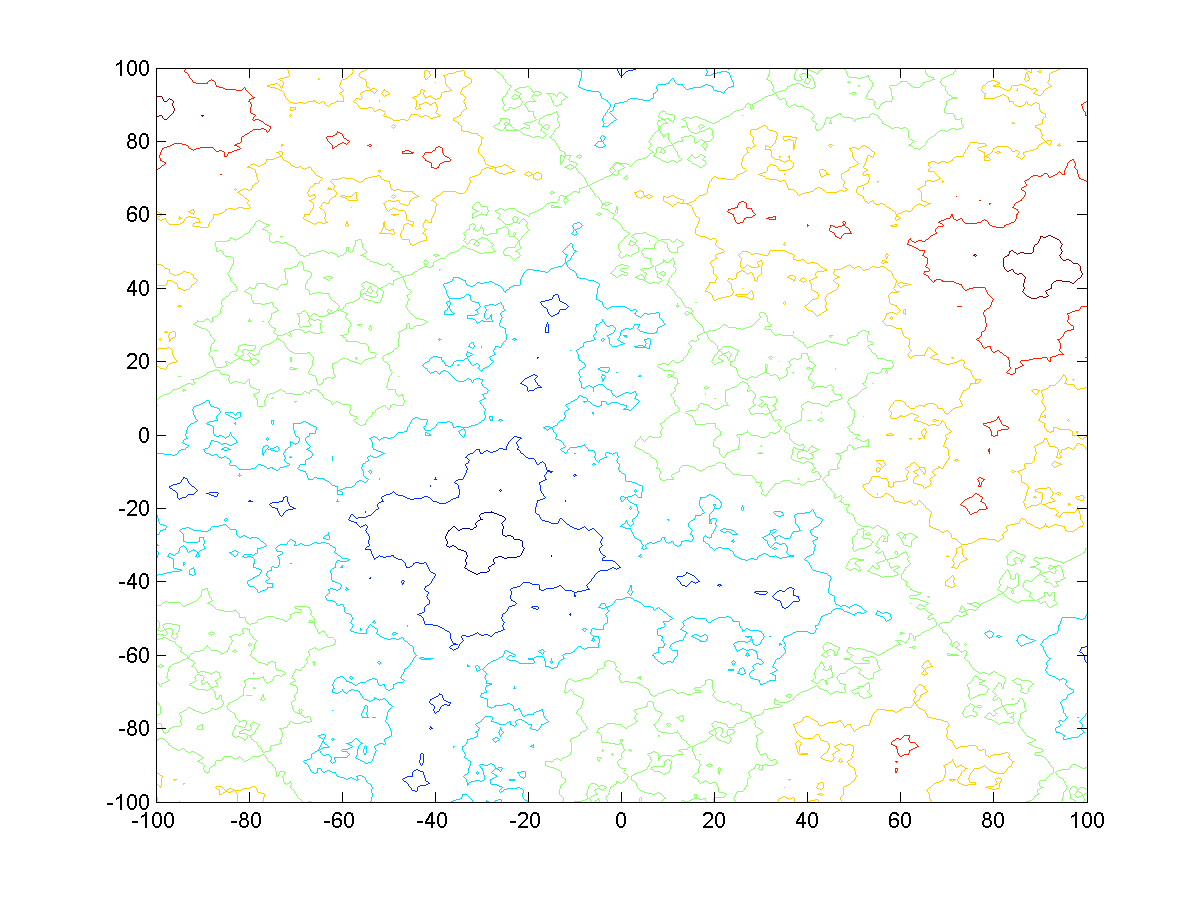}\\
  \caption{Weierstrass Function}\label{fig:weierstrass}
\end{figure}

\begin{figure}[tbp]
  \centering
  \includegraphics[width=.5\textwidth]{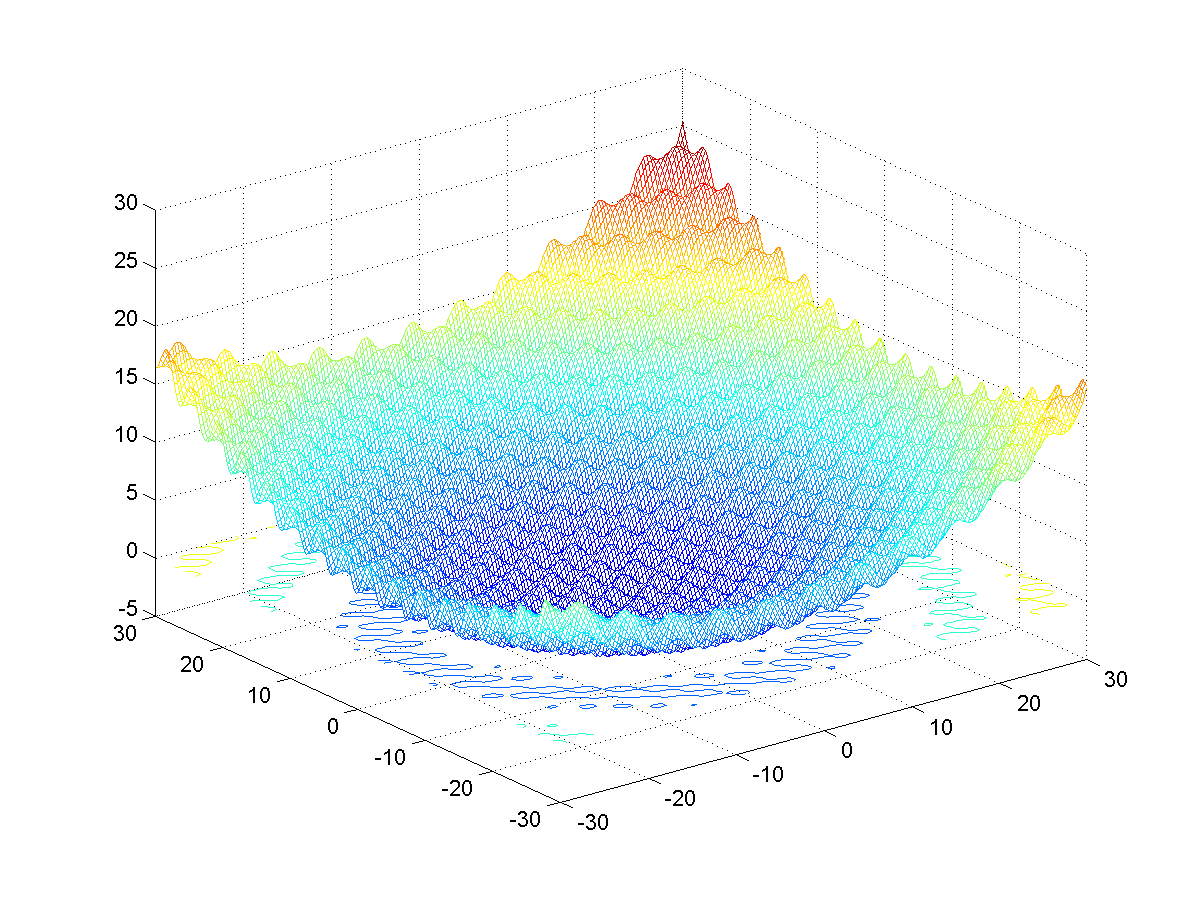}\includegraphics[width=.5\textwidth]{weierstrass-contour}\\
  \caption{Weierstrass Function}\label{fig:weierstrass}
\end{figure}

\begin{figure}[tbp]
  \centering
  \includegraphics[width=.5\textwidth]{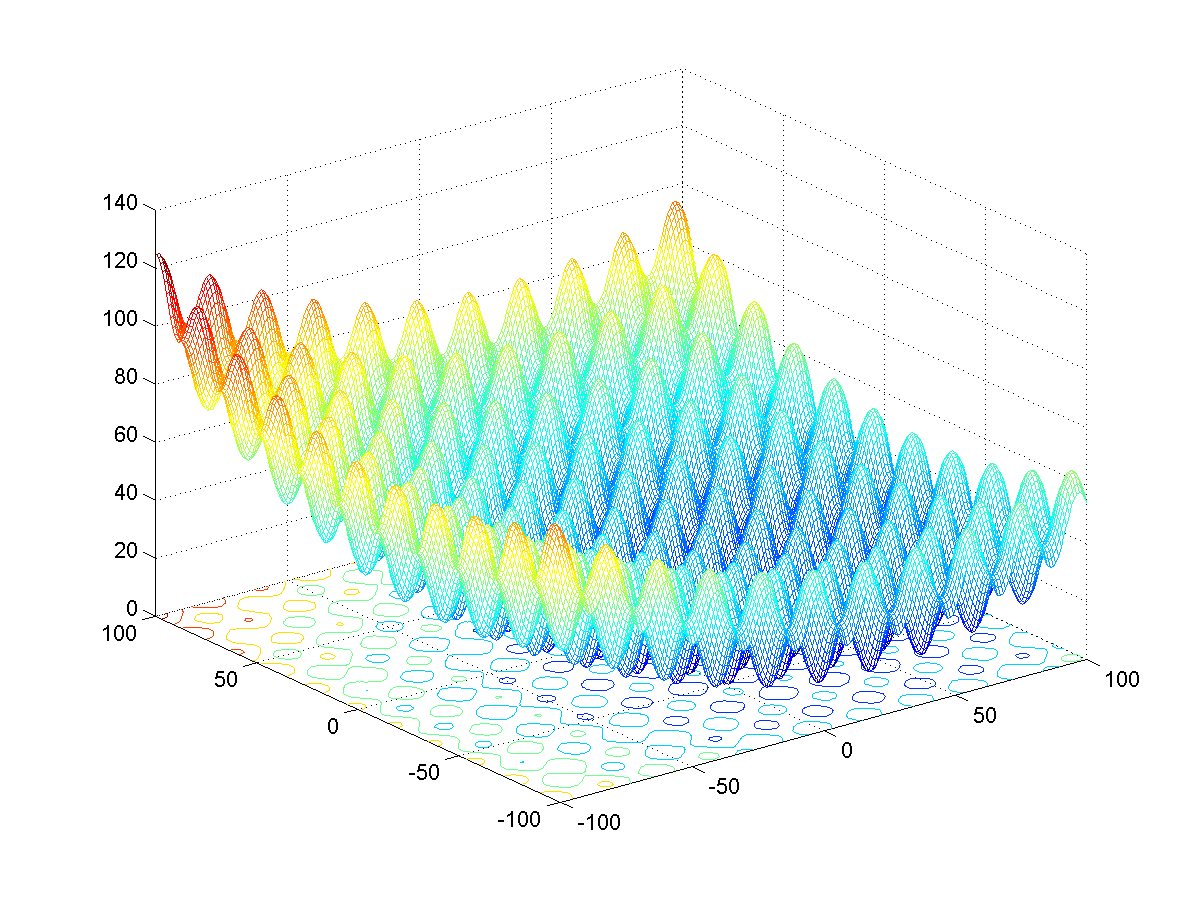}\includegraphics[width=.5\textwidth]{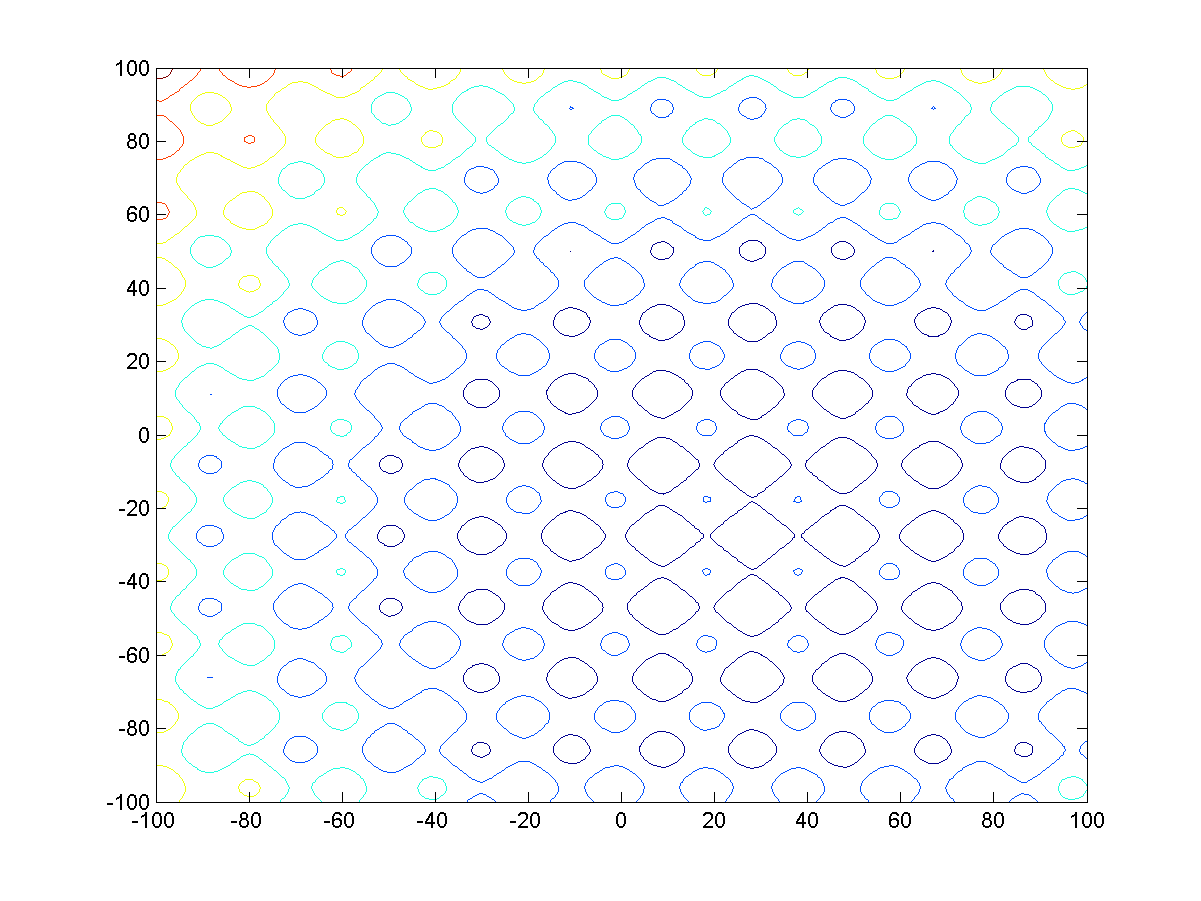}\\
  \caption{Rastrigin Function}\label{fig:rastrigin}
\end{figure}

\begin{figure}[tbp]
  \centering
  \includegraphics[width=.5\textwidth]{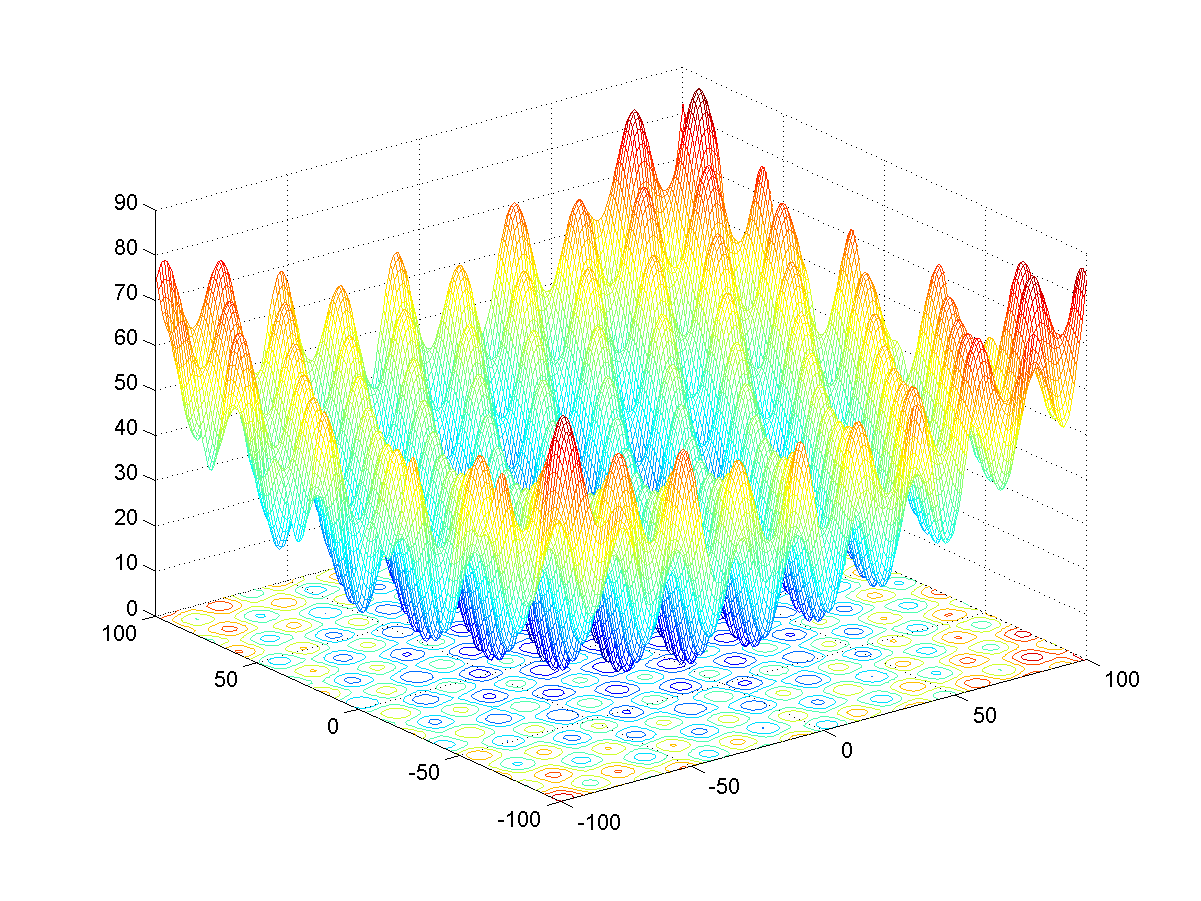}\includegraphics[width=.5\textwidth]{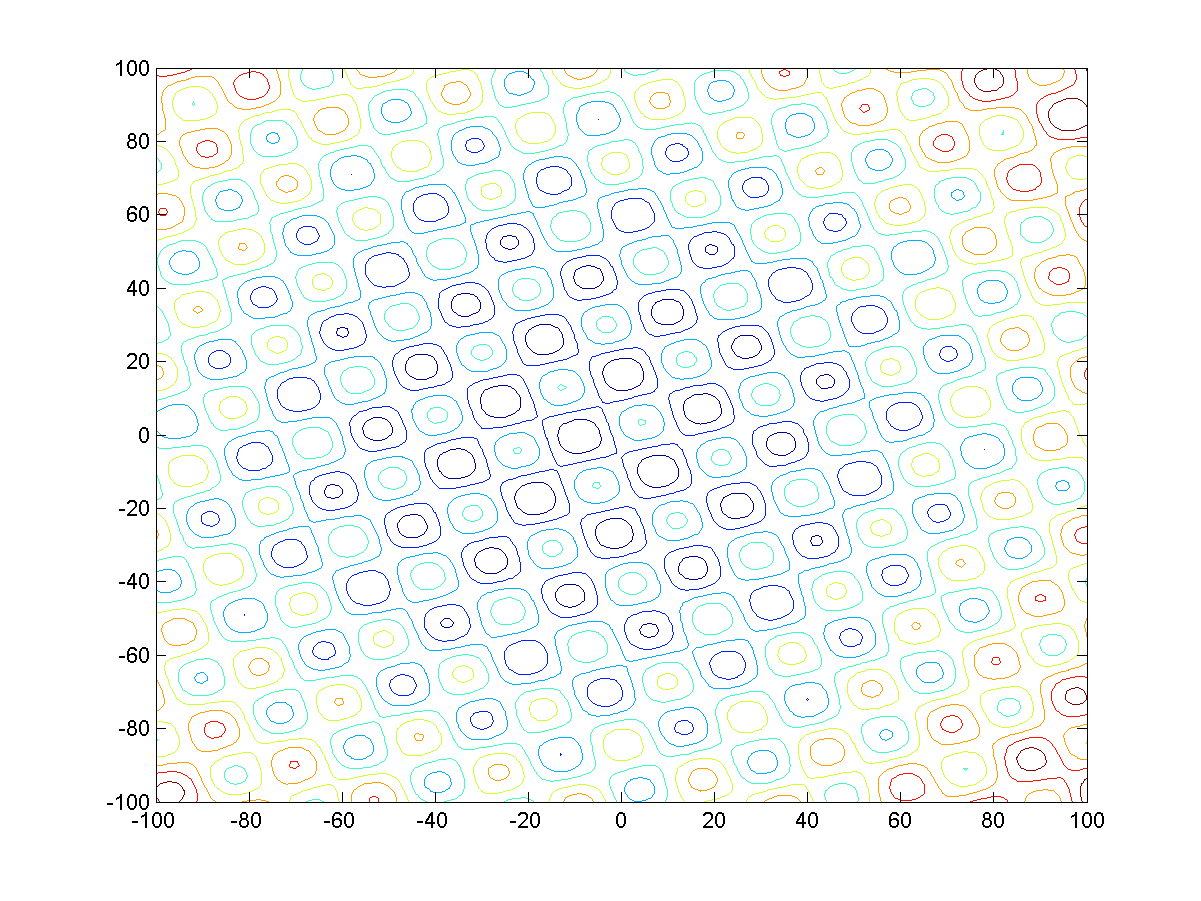}\\
  \caption{Rotated Rastrigin Function}\label{fig:rotated_rastrigin}
\end{figure}

\begin{figure}[tbp]
  \centering
  \includegraphics[width=.5\textwidth]{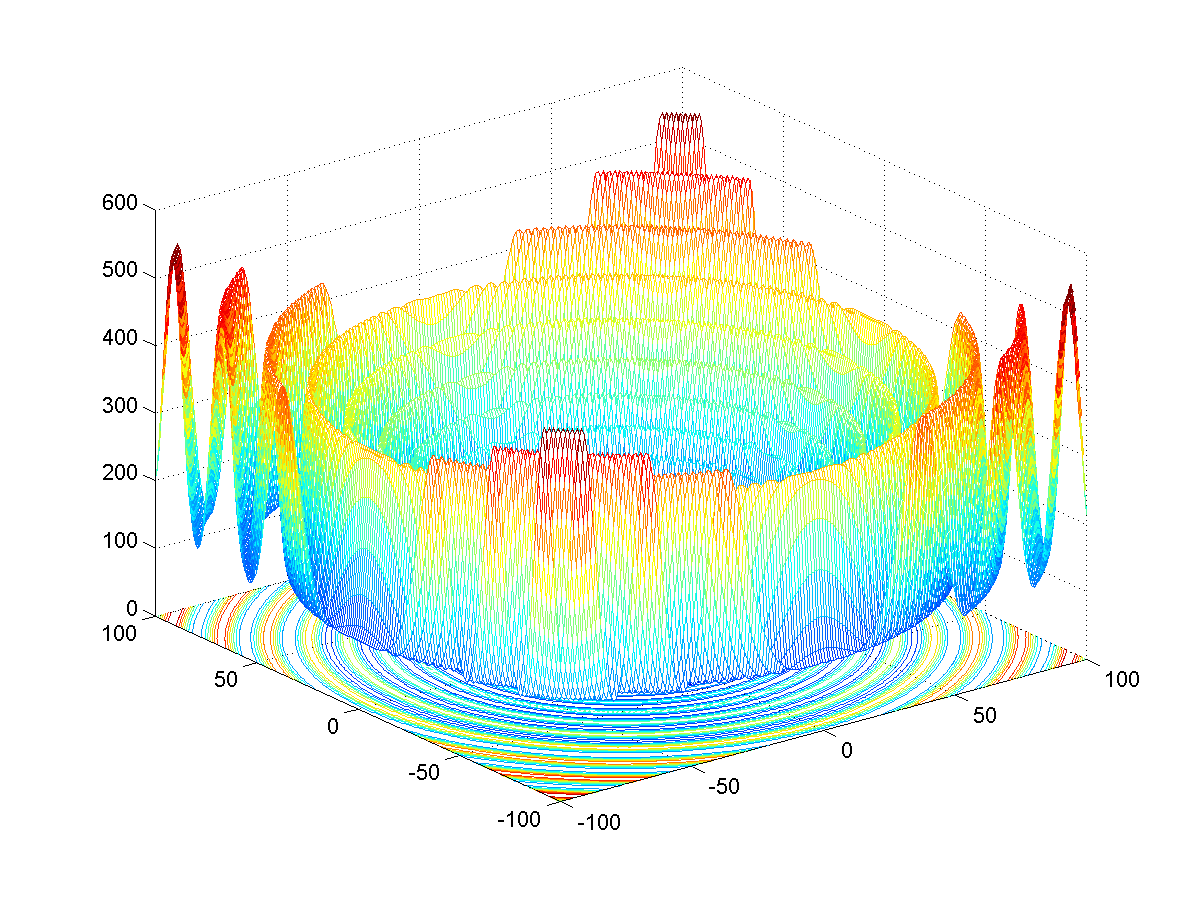}\includegraphics[width=.5\textwidth]{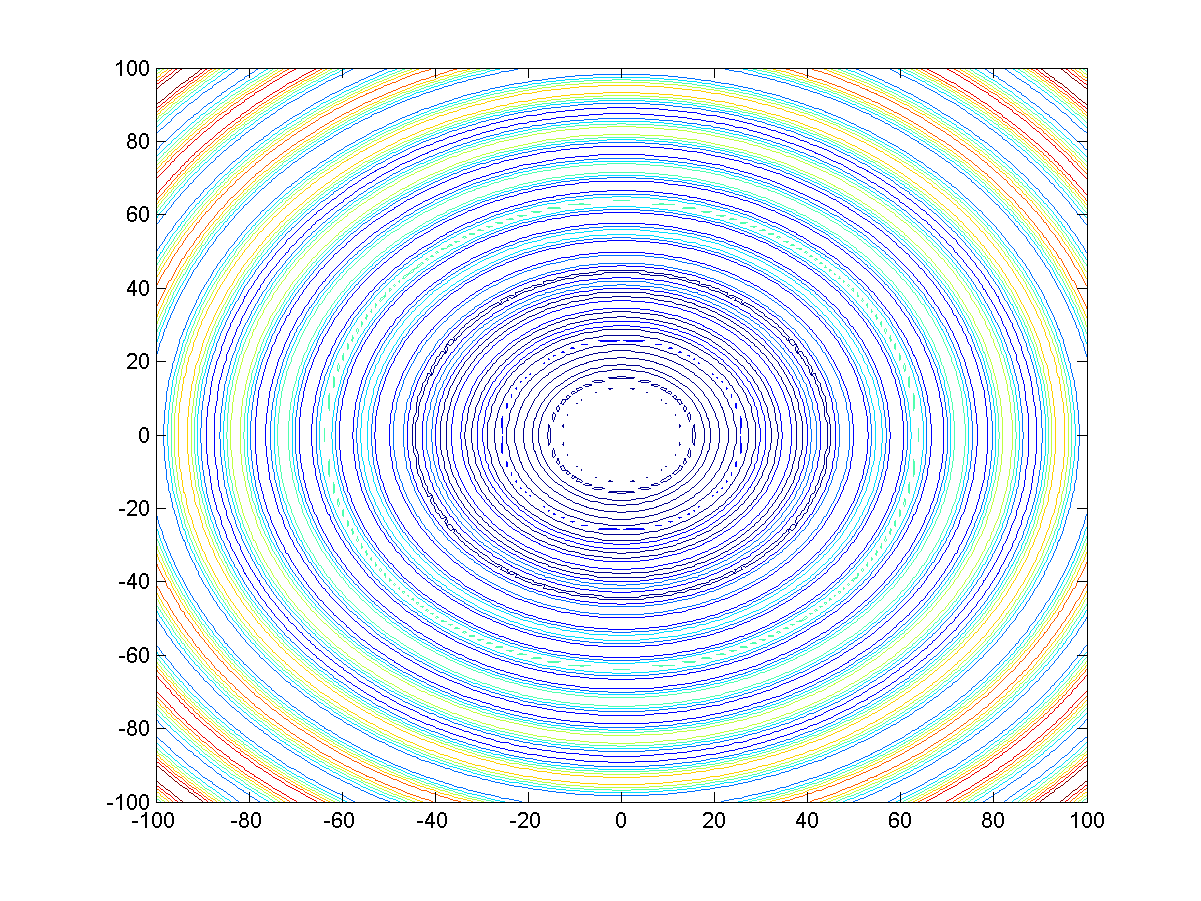}\\
  \caption{Schaffer's F7 Function}\label{fig:schaffersf7}
\end{figure}

\begin{figure}[tbp]
  \centering
  \includegraphics[width=.5\textwidth]{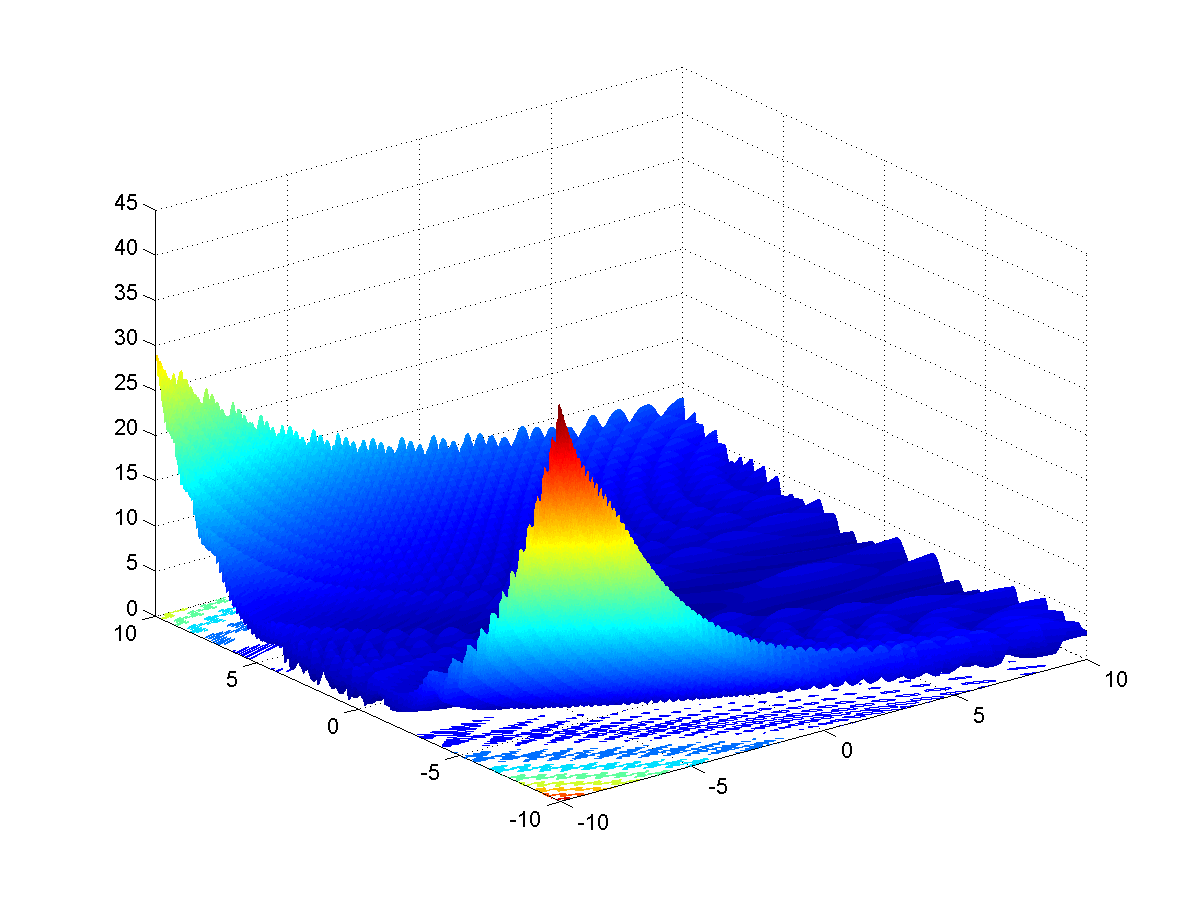}\includegraphics[width=.5\textwidth]{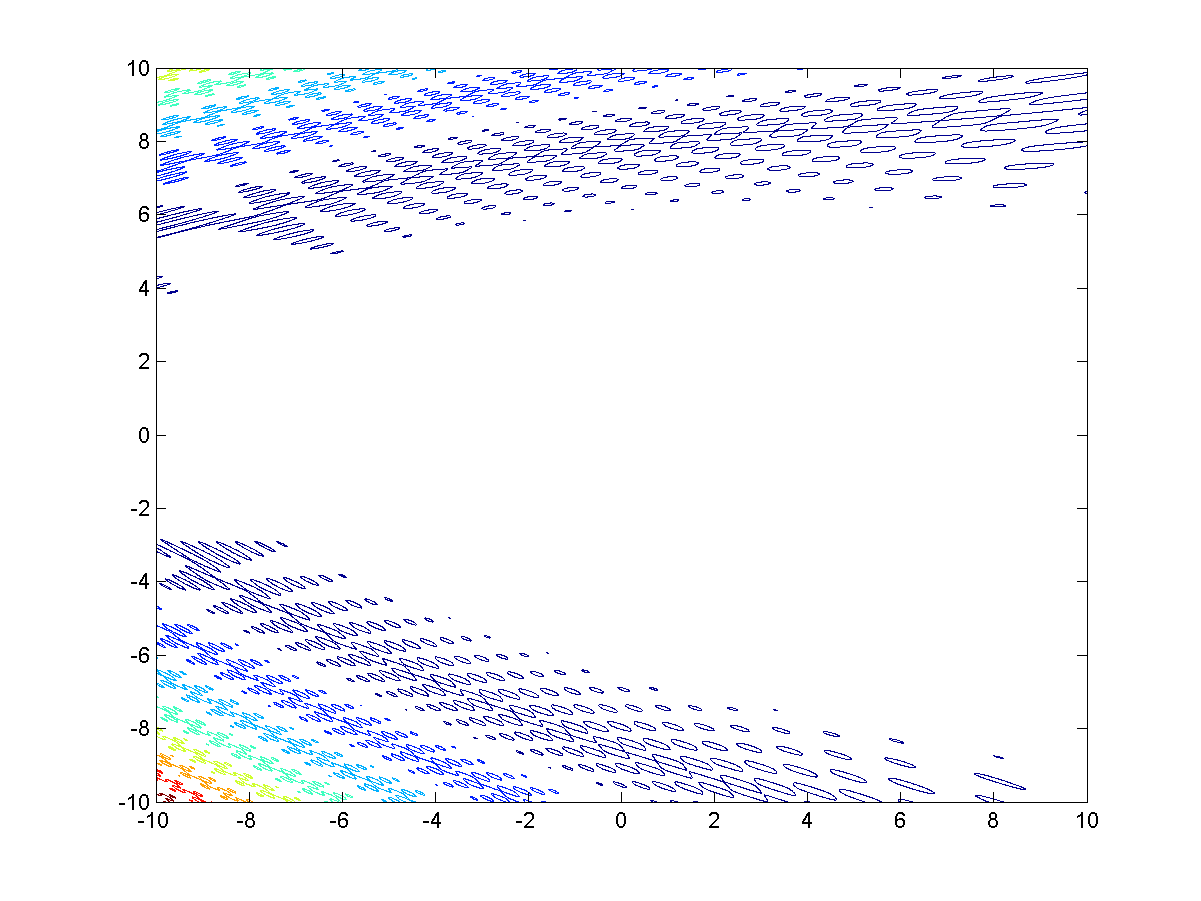}\\
  \caption{Expanded Griewank Rosenbrock Function}\label{fig:expanded_griewank_rosenbrock}
\end{figure}

\begin{figure}[tbp]
  \centering
  \includegraphics[width=.5\textwidth]{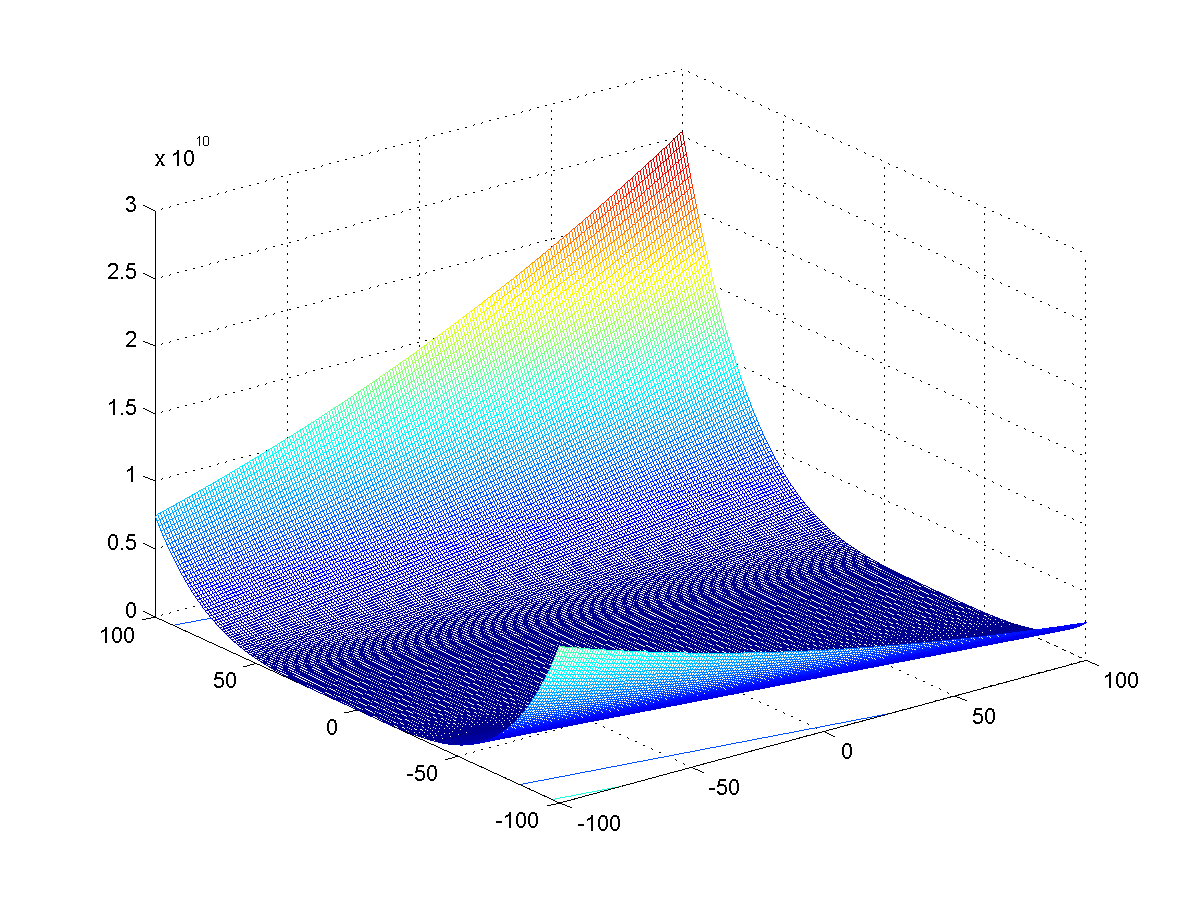}\includegraphics[width=.5\textwidth]{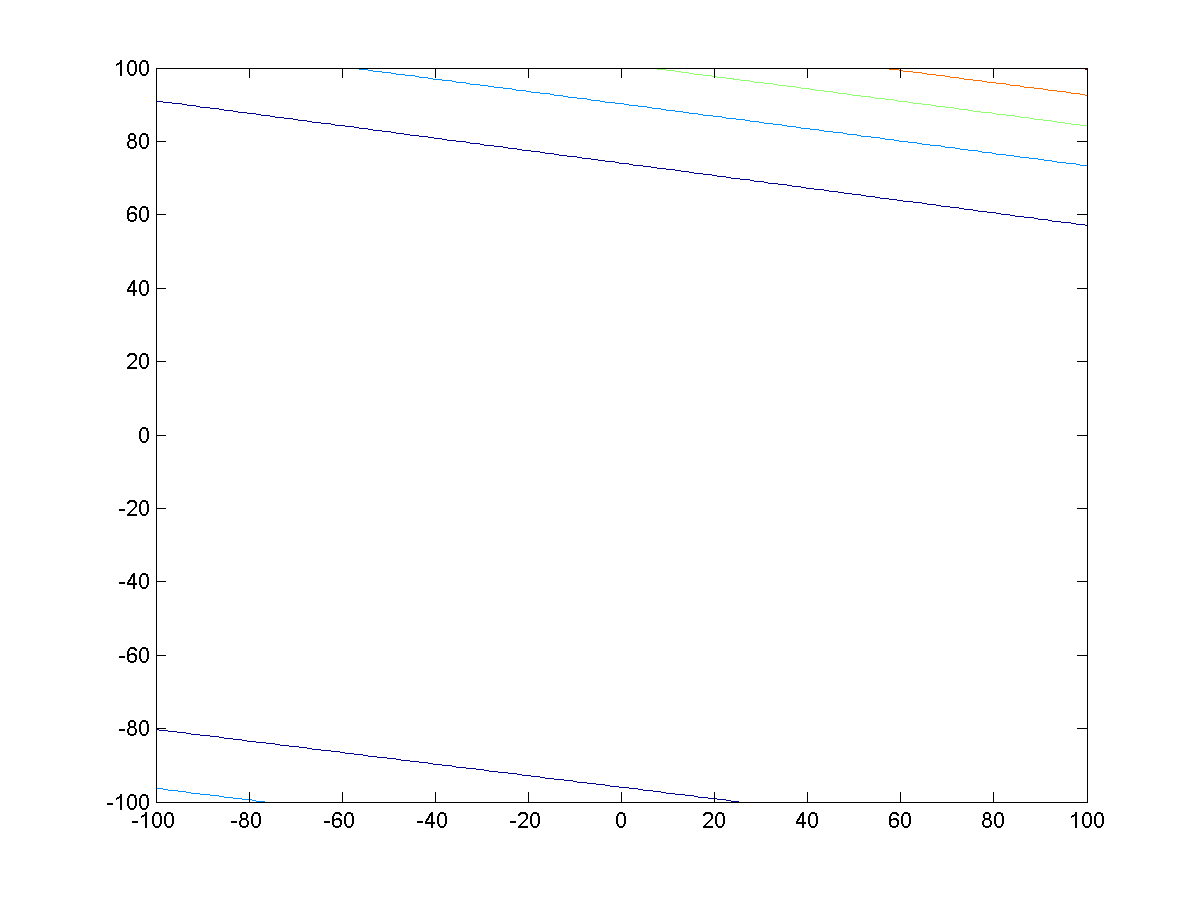}\\
  \caption{Rosenbrock Function}\label{fig:rosenbrock}
\end{figure}

\begin{figure}[tbp]
  \centering
  \includegraphics[width=.5\textwidth]{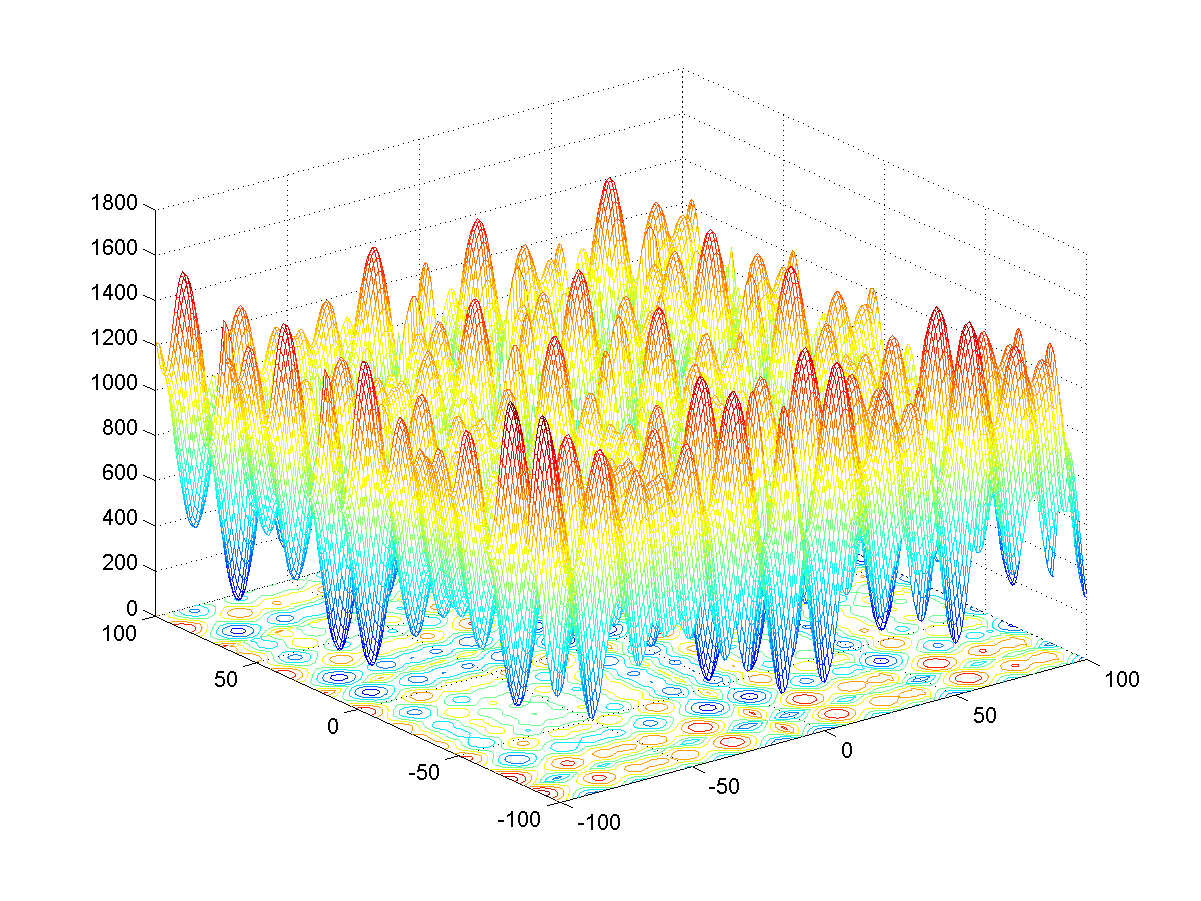}\includegraphics[width=.5\textwidth]{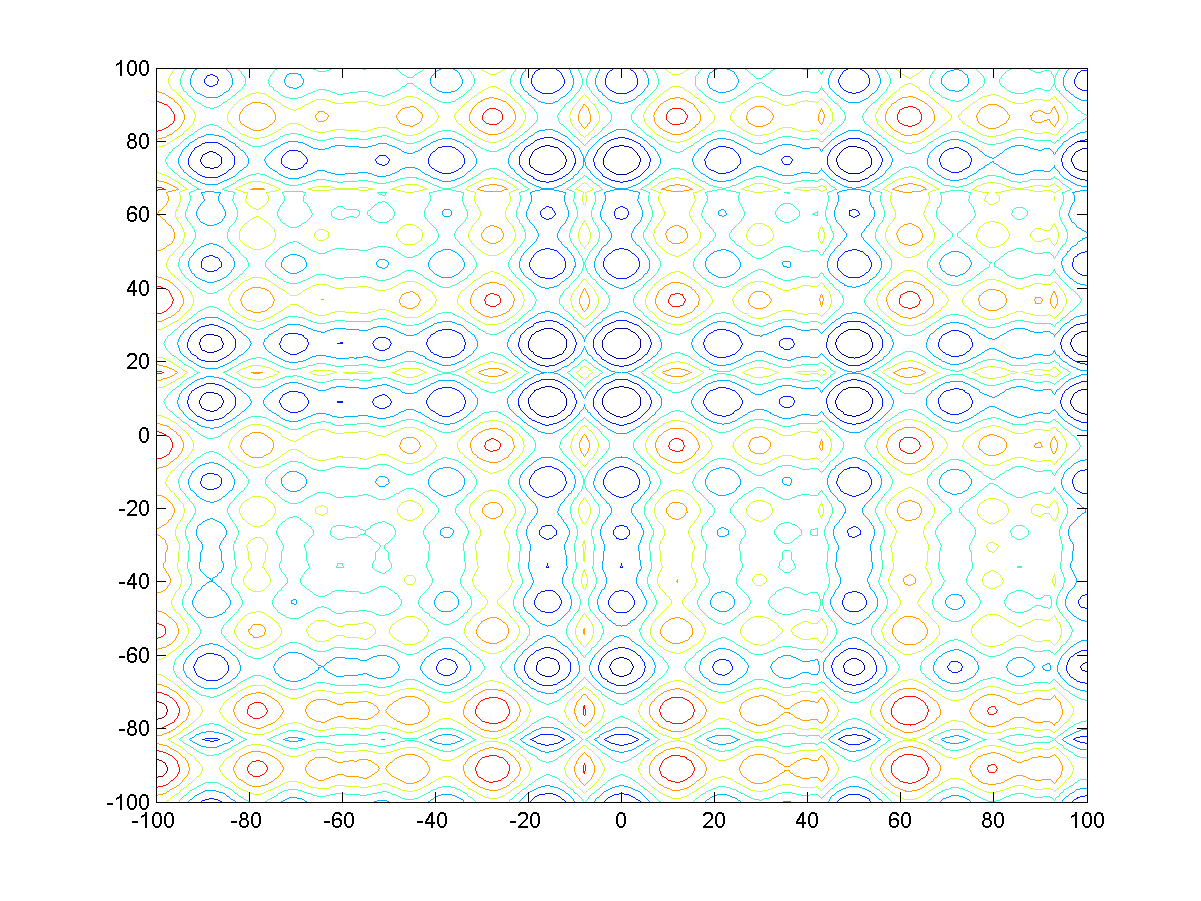}\\
  \caption{Schwefel Function}\label{fig:schwefel}
\end{figure}


\begin{figure}[tbp]
  \centering
  \includegraphics[width=.5\textwidth]{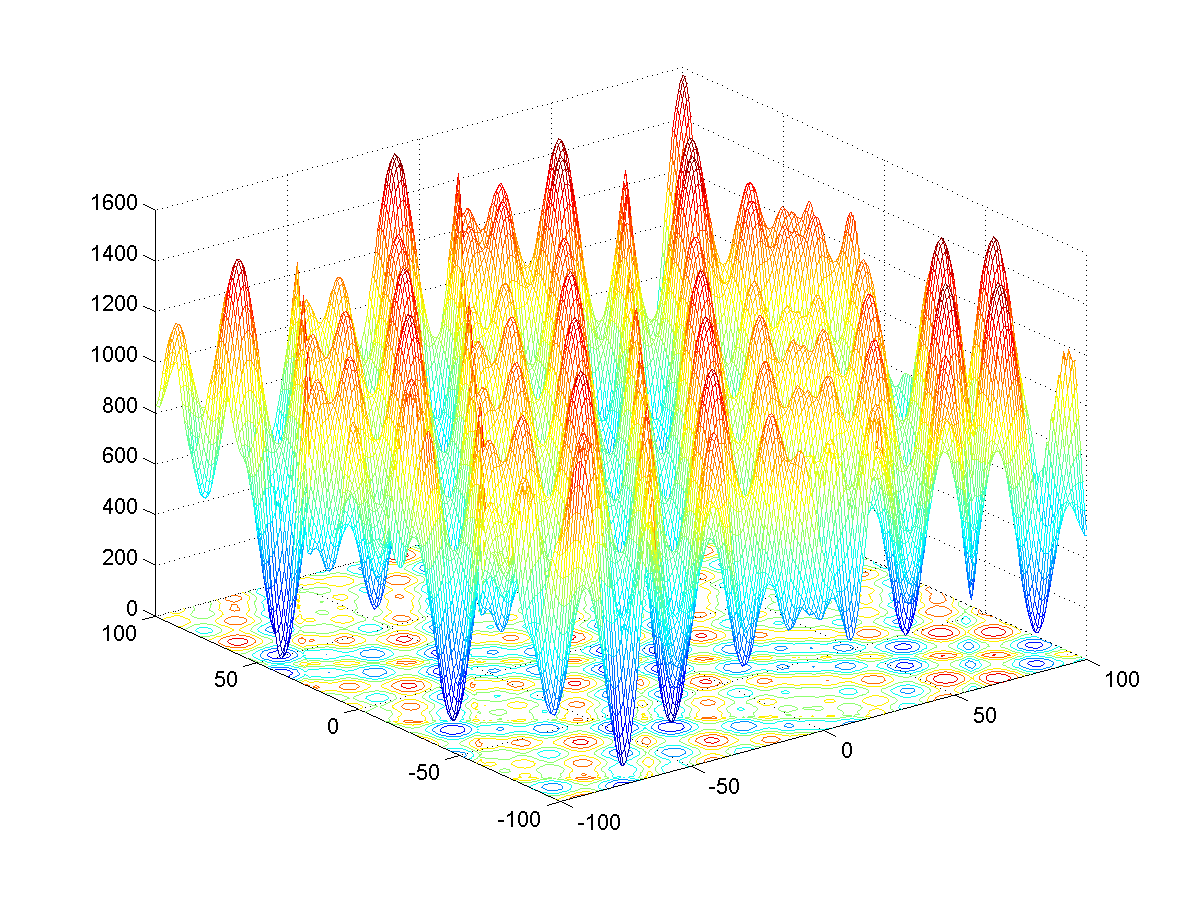}\includegraphics[width=.5\textwidth]{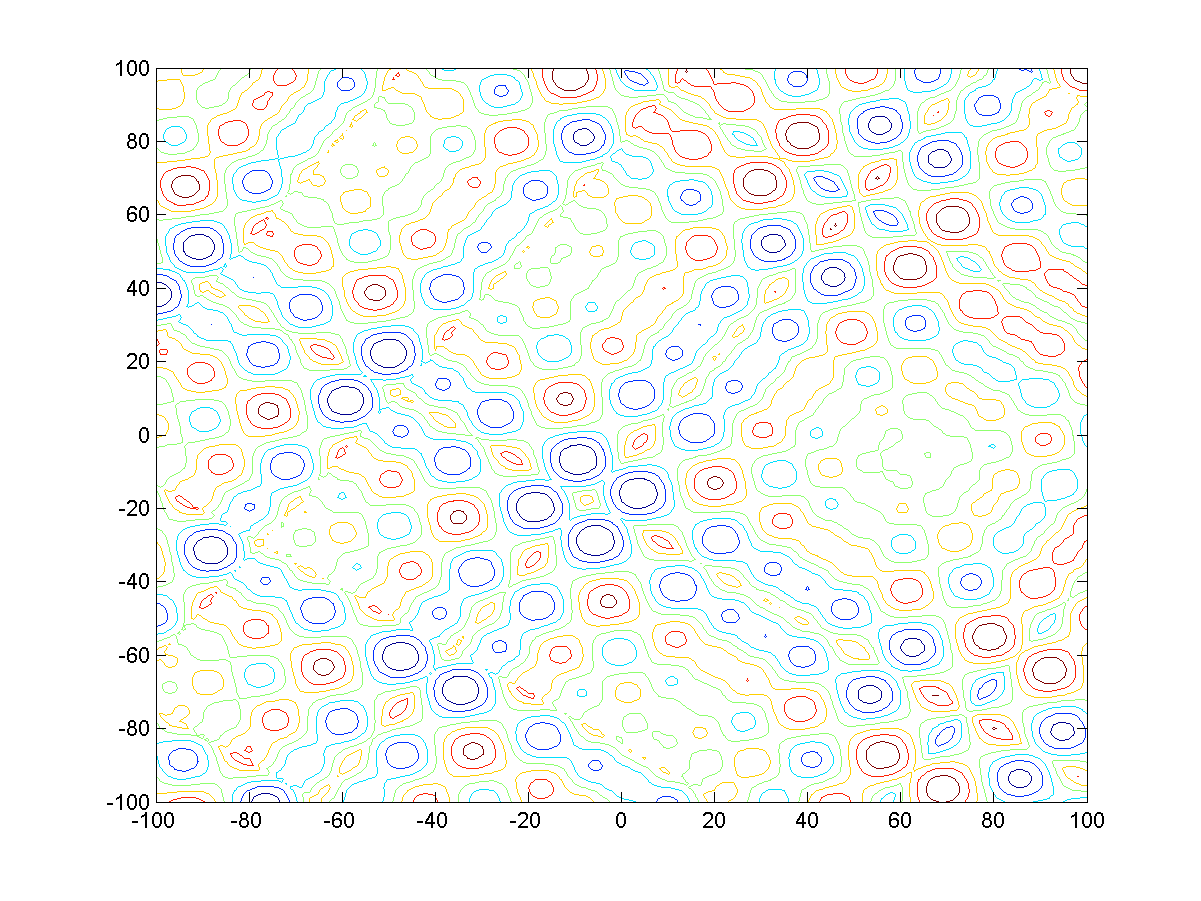}\\
  \caption{Rotated Schwefel Function}\label{fig:rotated_schwefel}
\end{figure}

\begin{figure}[tbp]
  \centering
  \includegraphics[width=.5\textwidth]{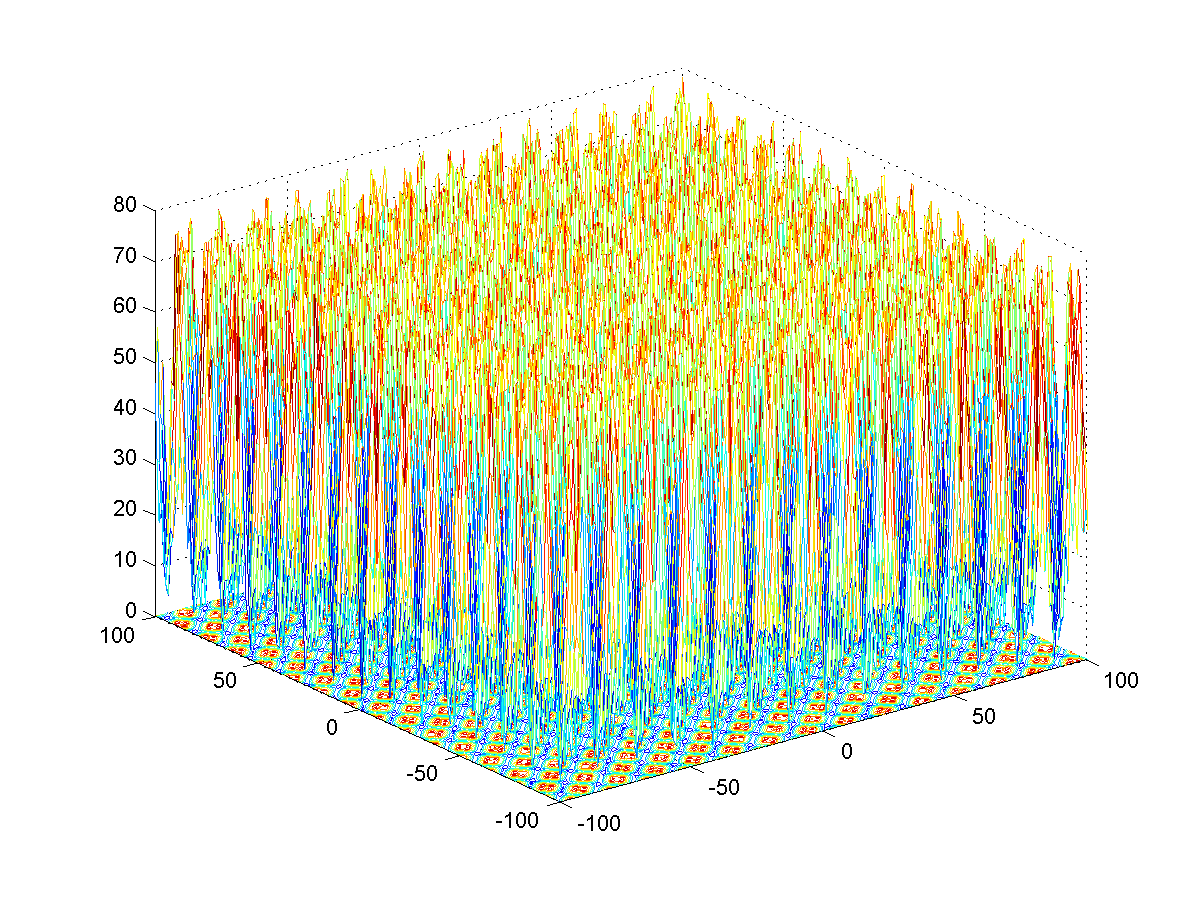}\includegraphics[width=.5\textwidth]{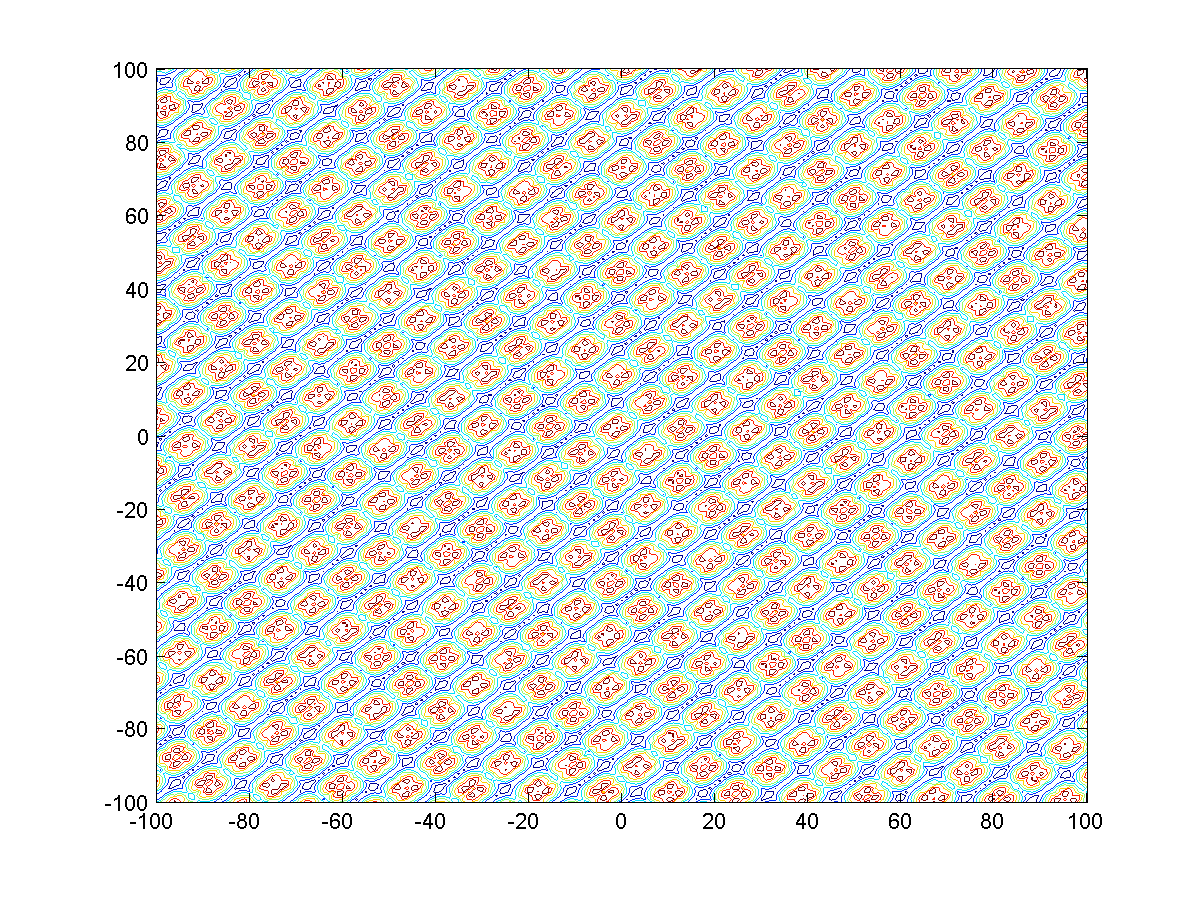}\\
  \caption{Katsuura Function}\label{fig:katsuura}
\end{figure}

\begin{figure}[tbp]
  \centering
  \includegraphics[width=.5\textwidth]{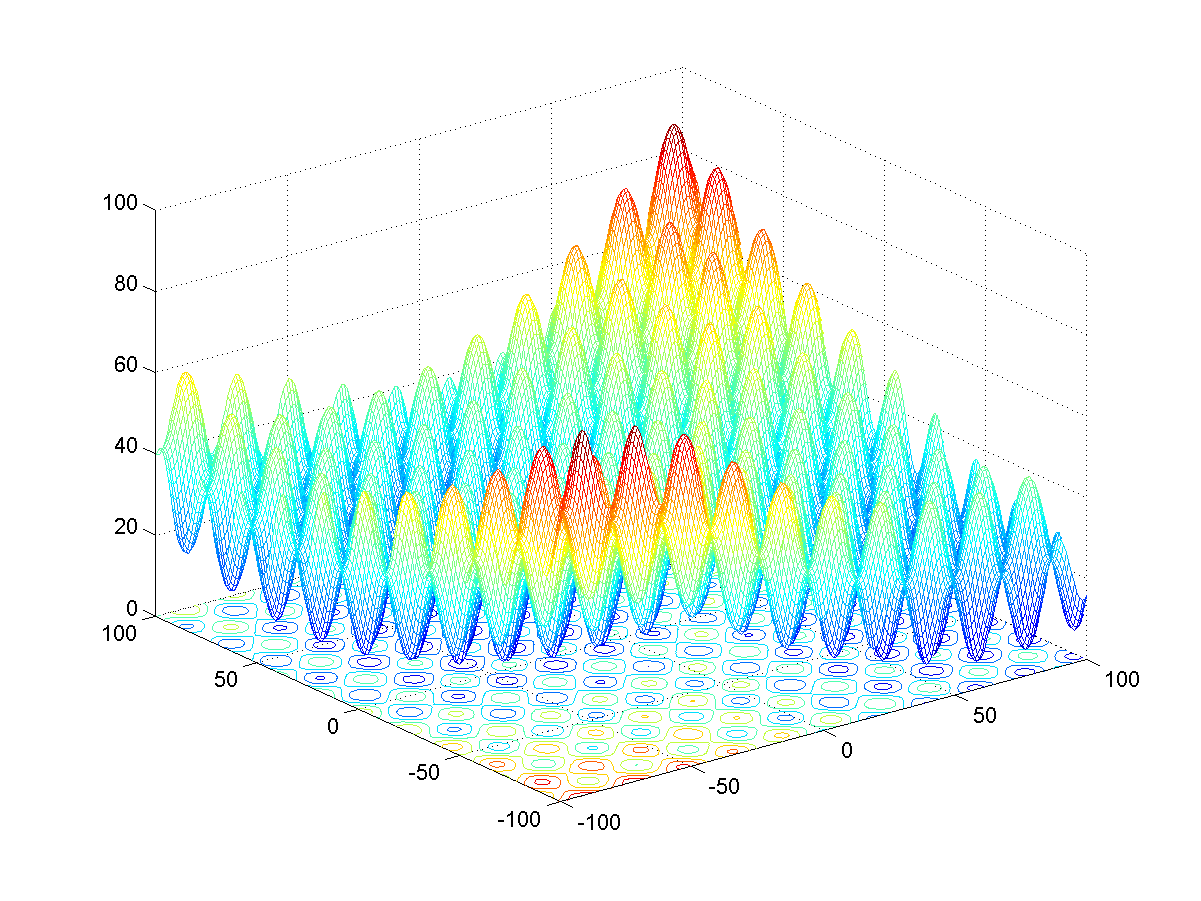}\includegraphics[width=.5\textwidth]{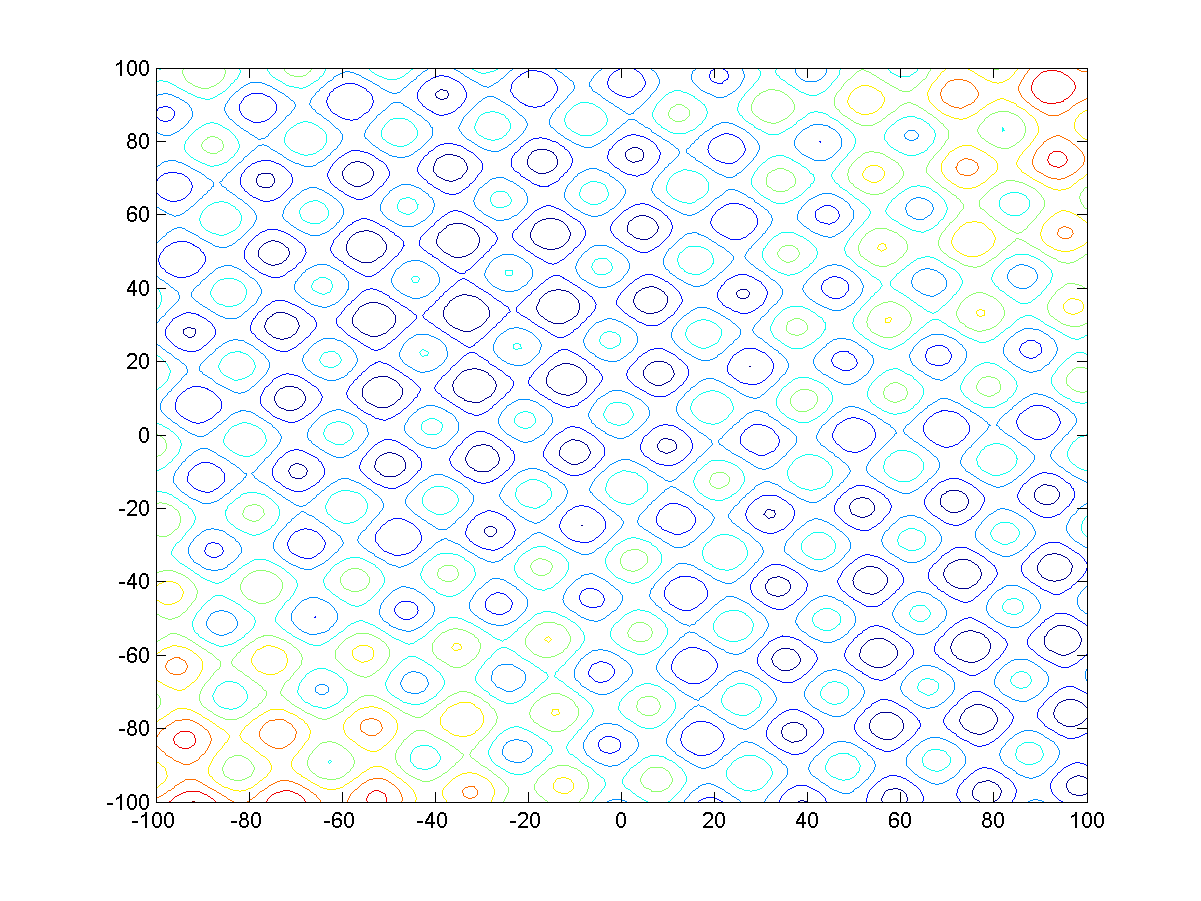}\\
  \caption{Lunacek Function}\label{fig:lunacek}
\end{figure}

\begin{figure}[tbp]
  \centering
  \includegraphics[width=.5\textwidth]{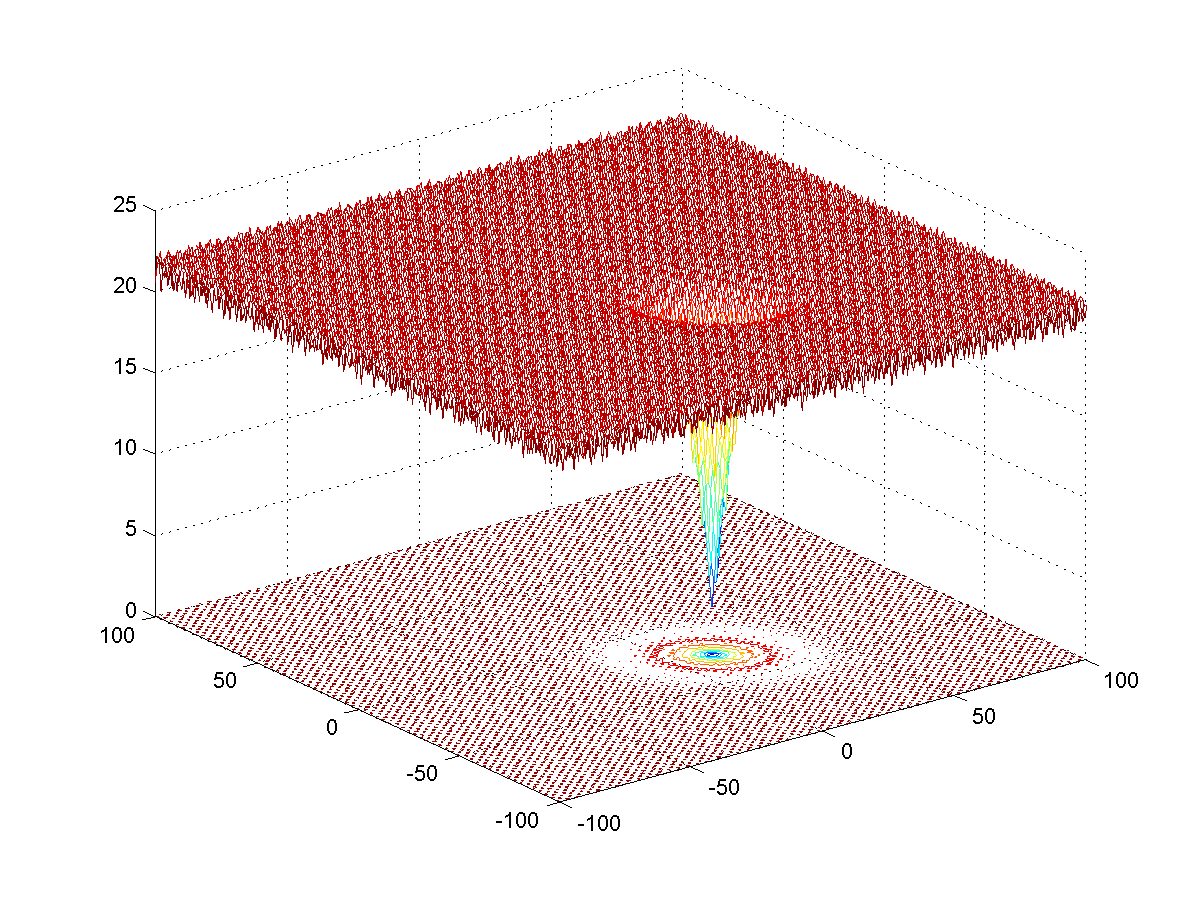}\includegraphics[width=.5\textwidth]{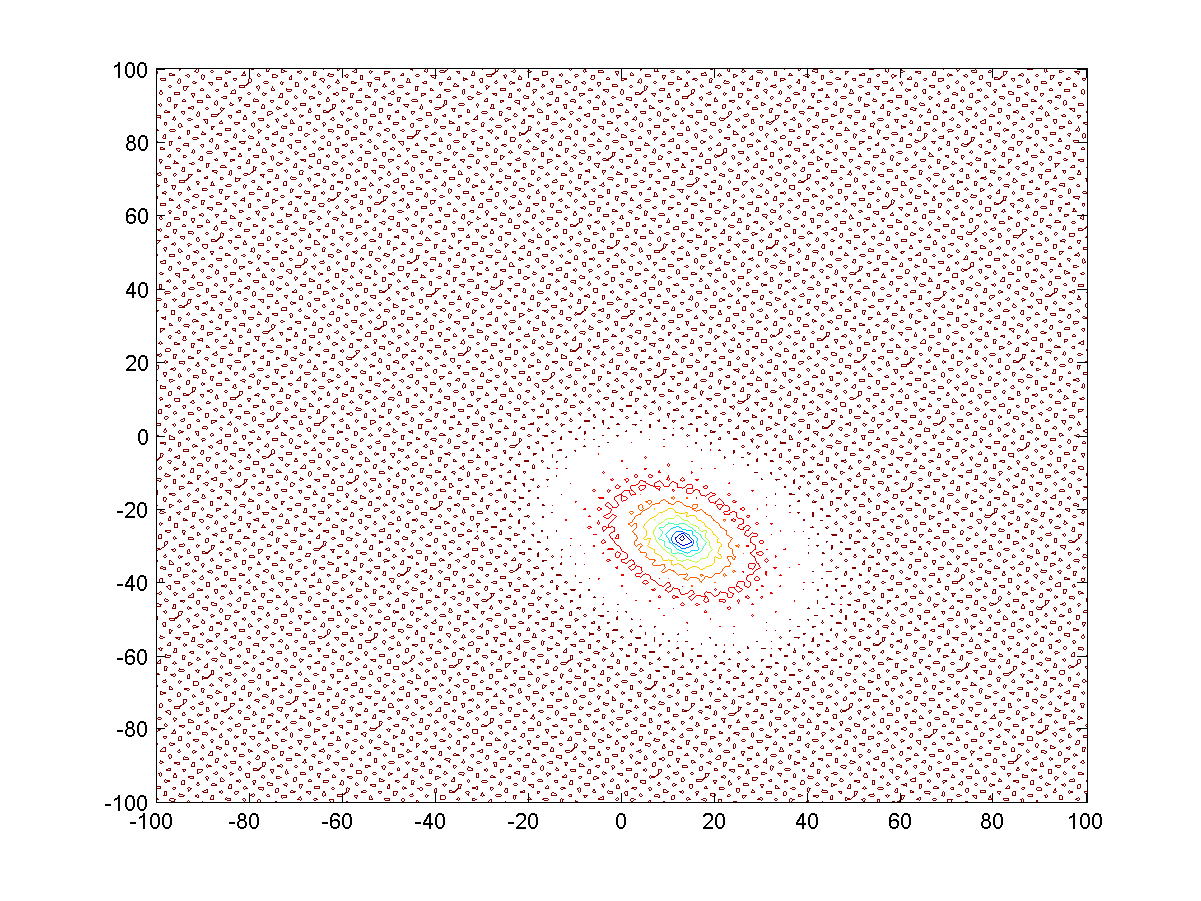}\\
  \caption{Ackley Function}\label{fig:ackley}
\end{figure}

\begin{figure}[tbp]
  \centering
  \includegraphics[width=.5\textwidth]{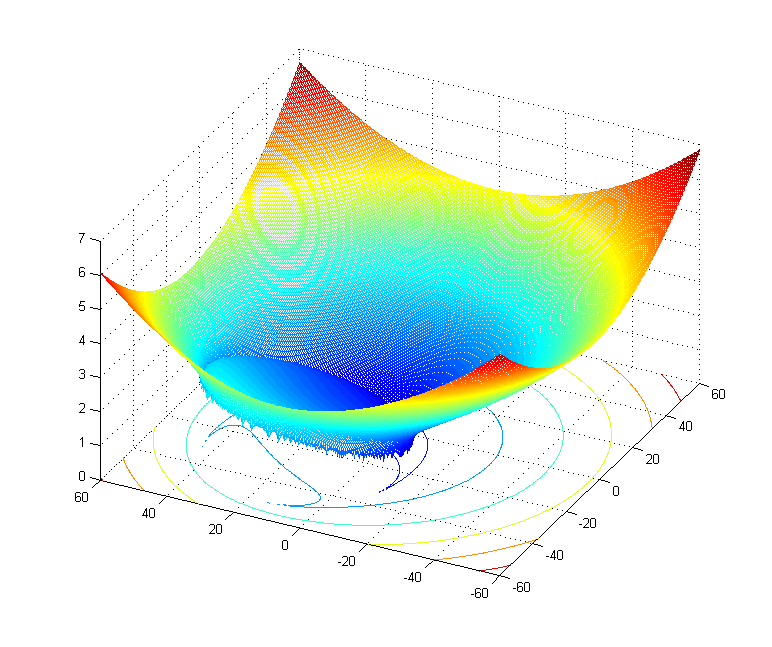}\includegraphics[width=.5\textwidth]{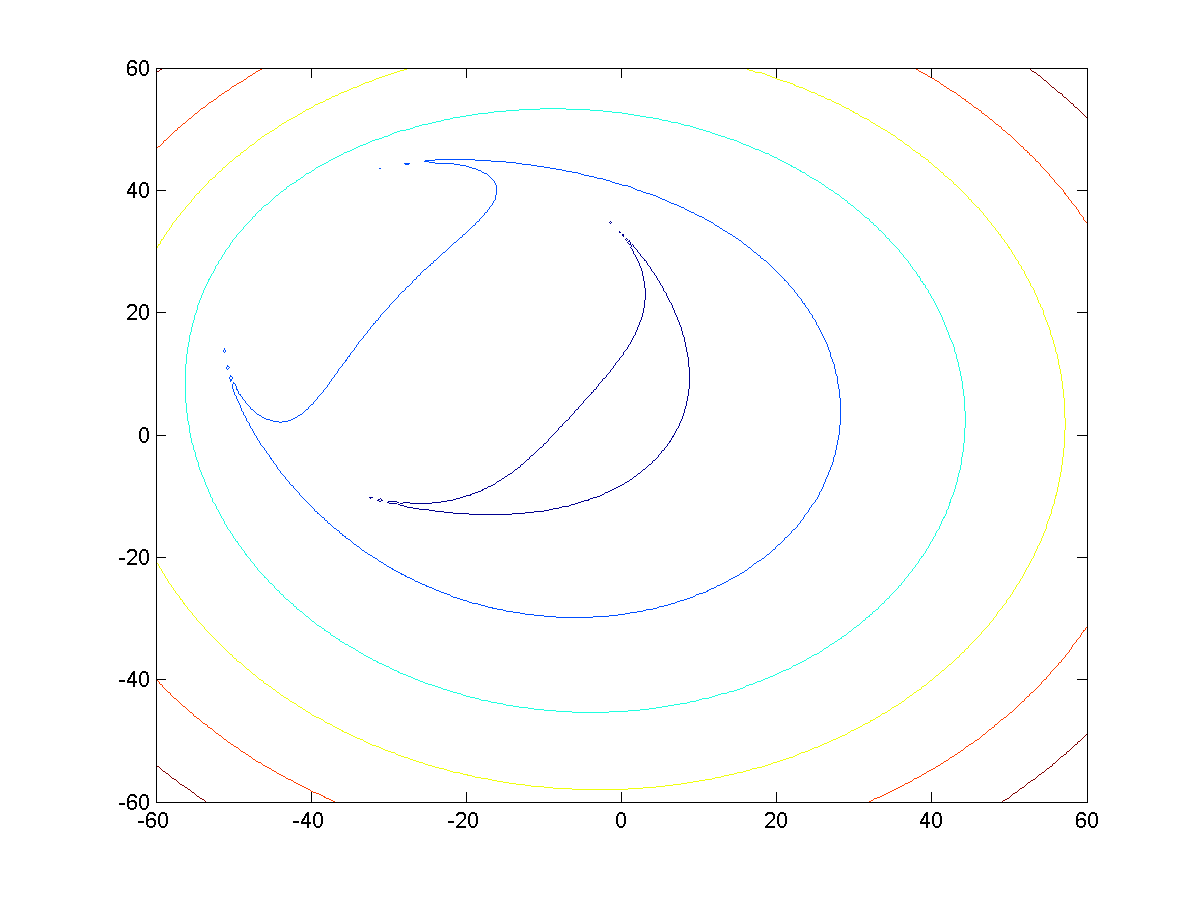}\\
  \caption{HappyCat Function}\label{fig:happycat}
\end{figure}

\begin{figure}[tbp]
  \centering
  \includegraphics[width=.5\textwidth]{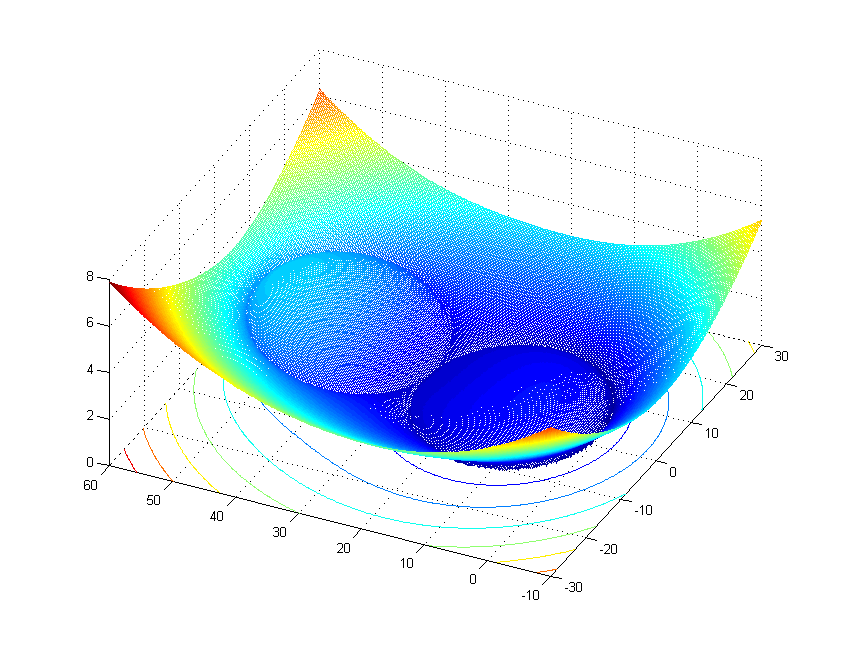}\includegraphics[width=.5\textwidth]{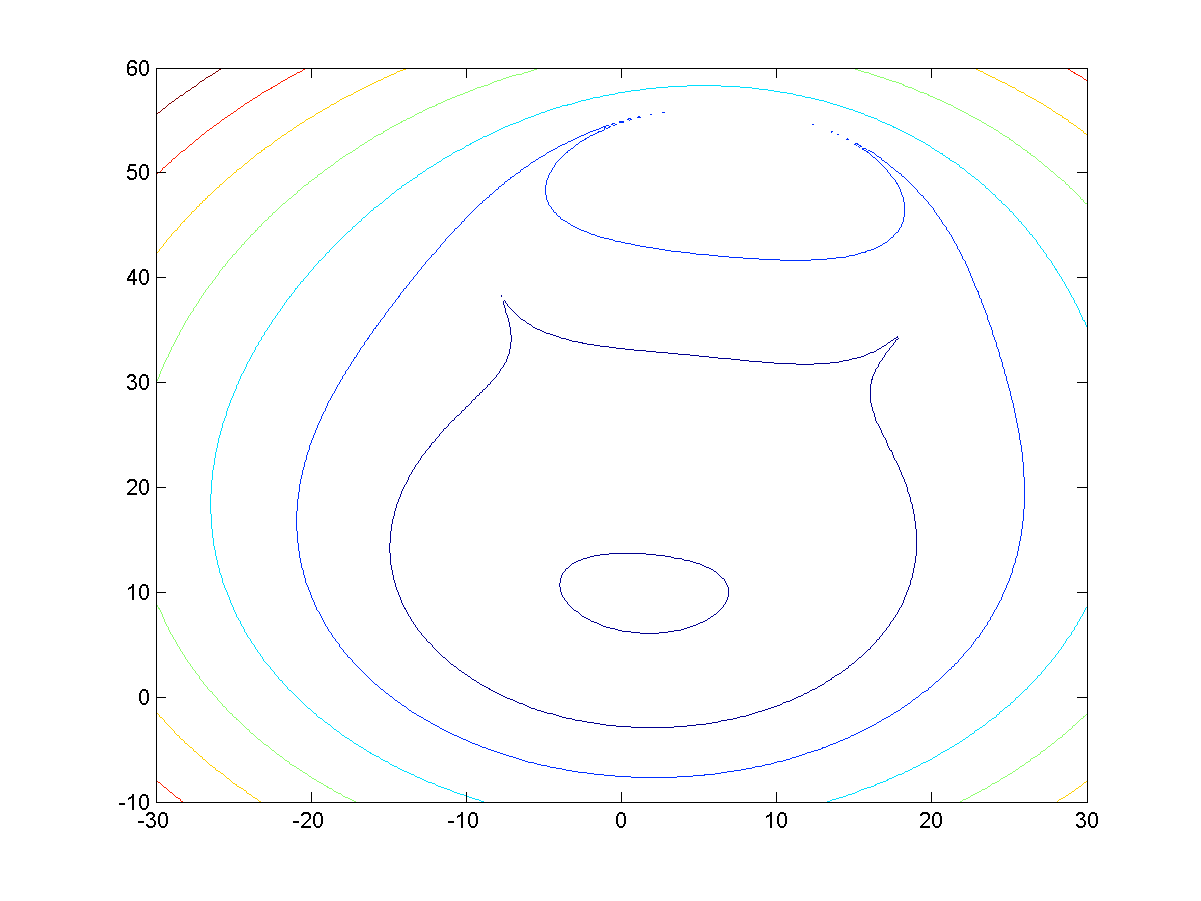}\\
  \caption{HGBat Function}\label{fig:hgbat}
\end{figure}

\begin{figure}[tbp]
  \centering
  \includegraphics[width=.5\textwidth]{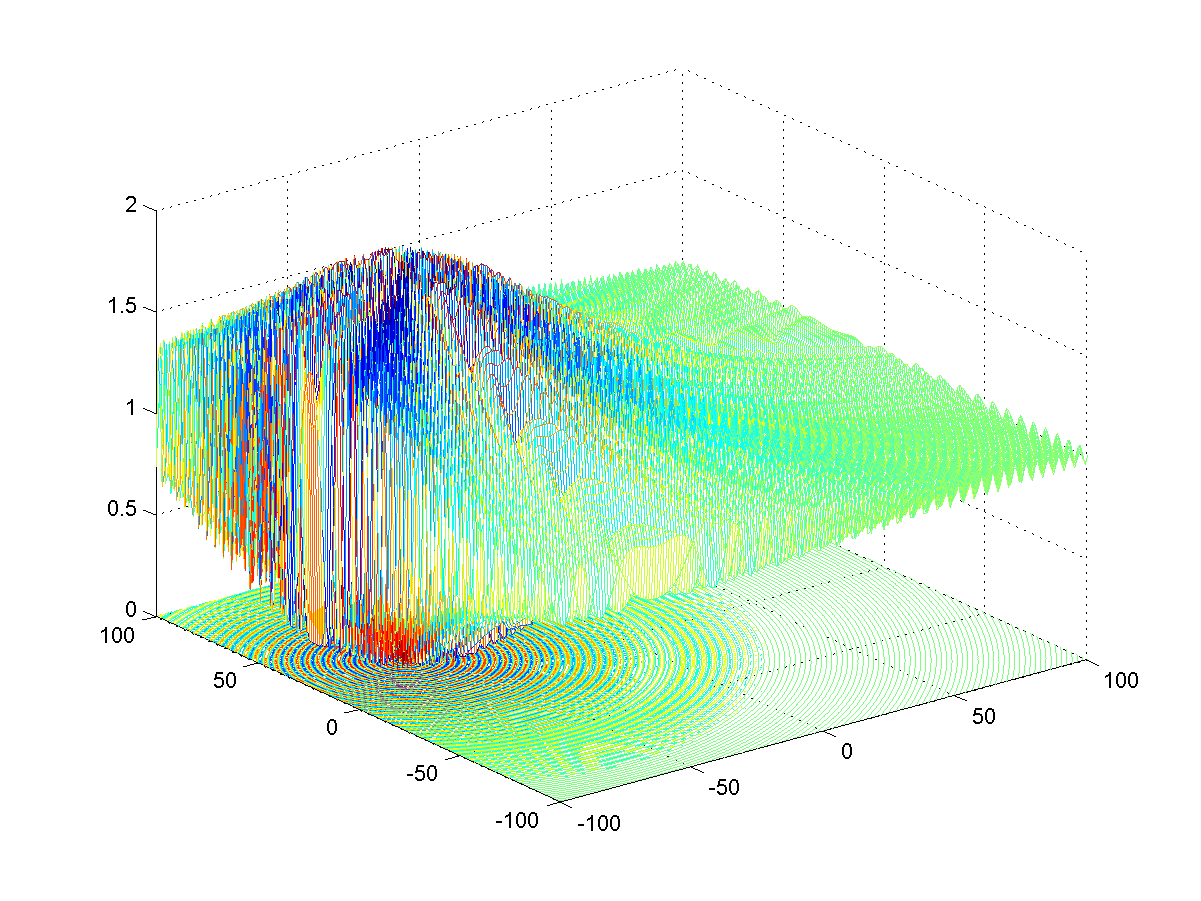}\includegraphics[width=.5\textwidth]{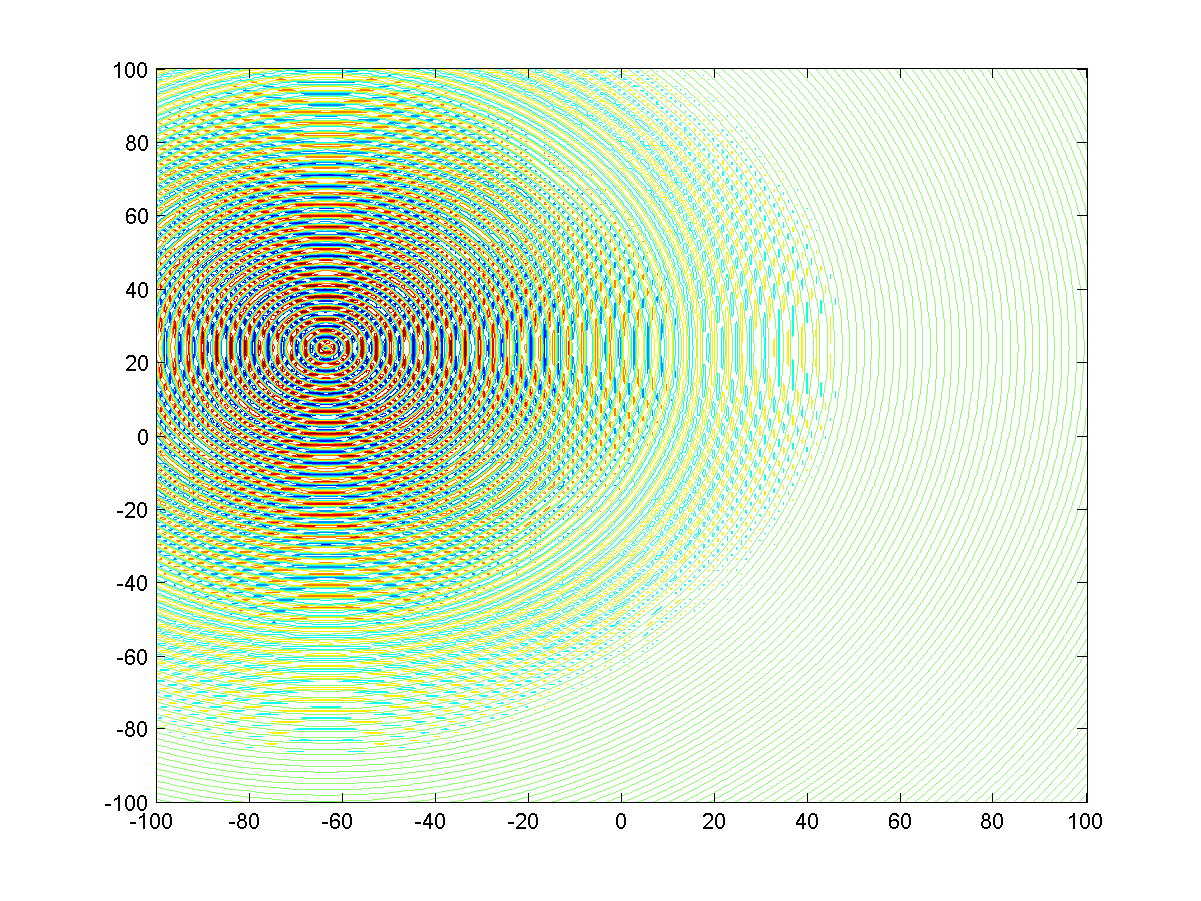}\\
  \caption{Expanded Scaffers' F6 Function}\label{fig:expanded_scaffersf6}
\end{figure}

\begin{figure}[tbp]
  \centering
  \includegraphics[width=.5\textwidth]{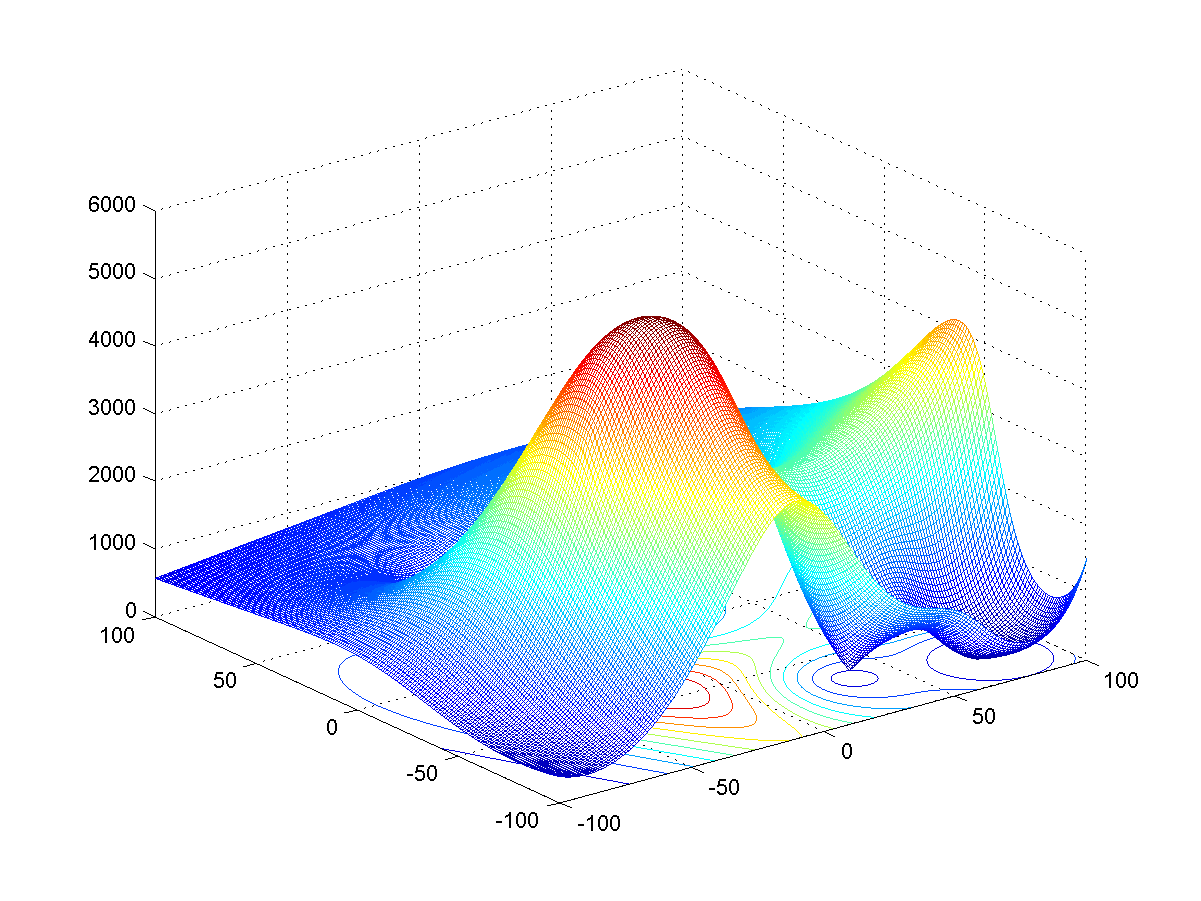}\includegraphics[width=.5\textwidth]{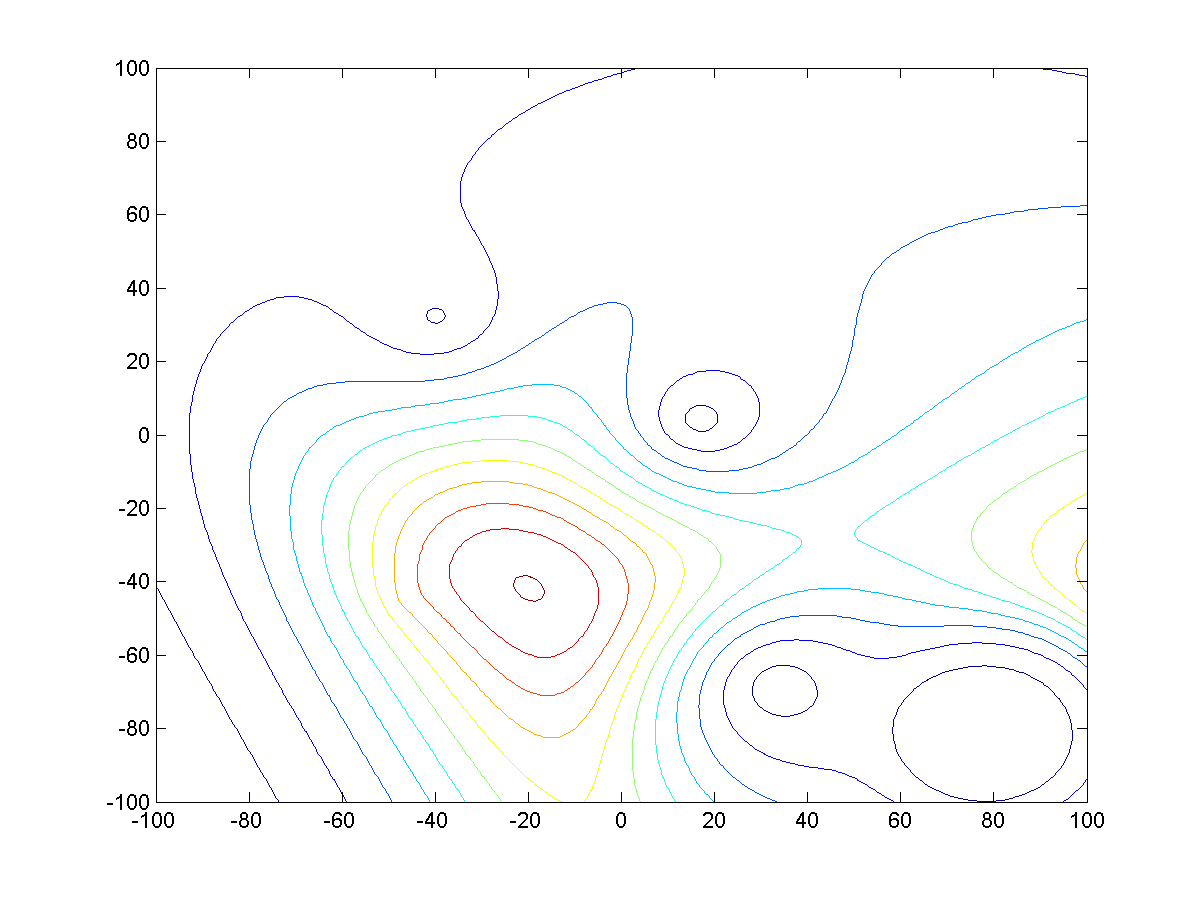}\\
  \caption{Composition Function 1}\label{fig:composition1}
\end{figure}

\begin{figure}[tbp]
  \centering
  \includegraphics[width=.5\textwidth]{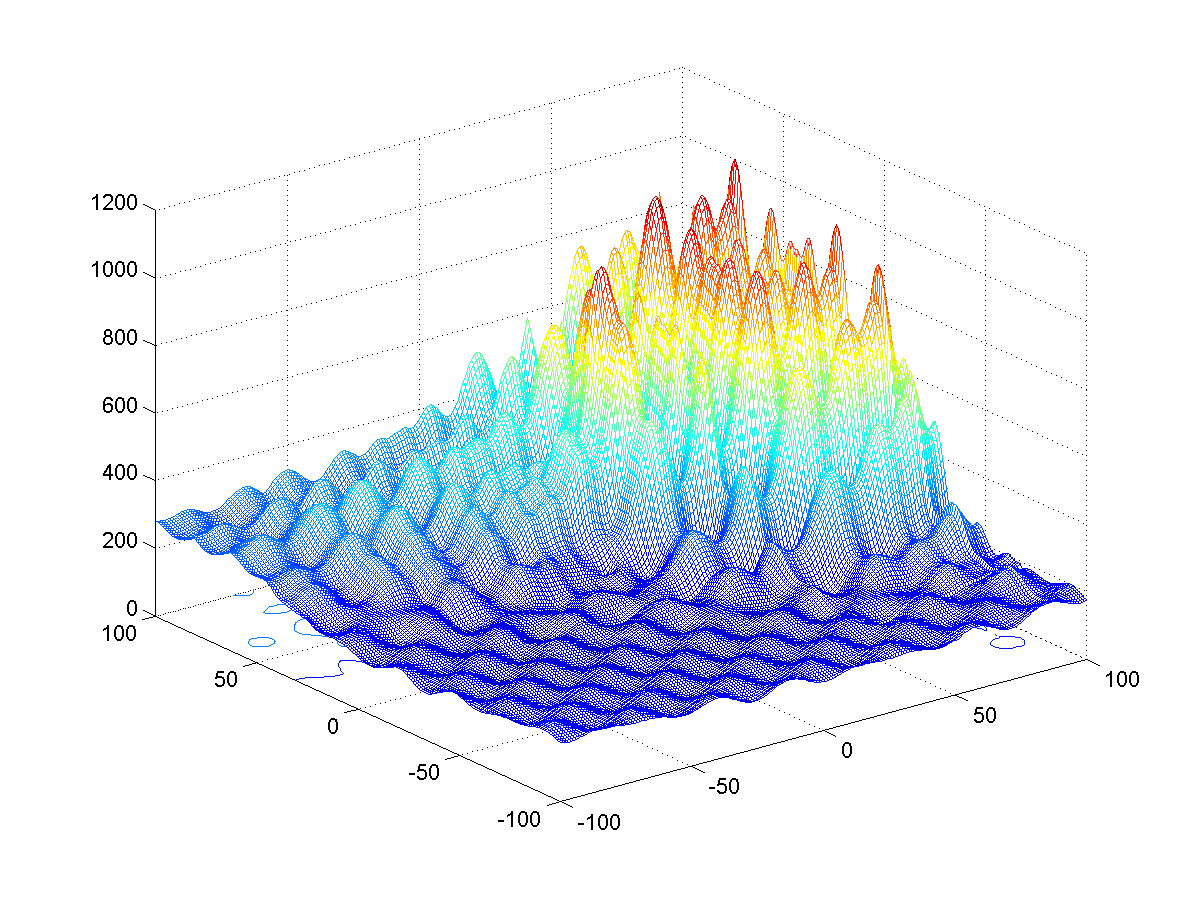}\includegraphics[width=.5\textwidth]{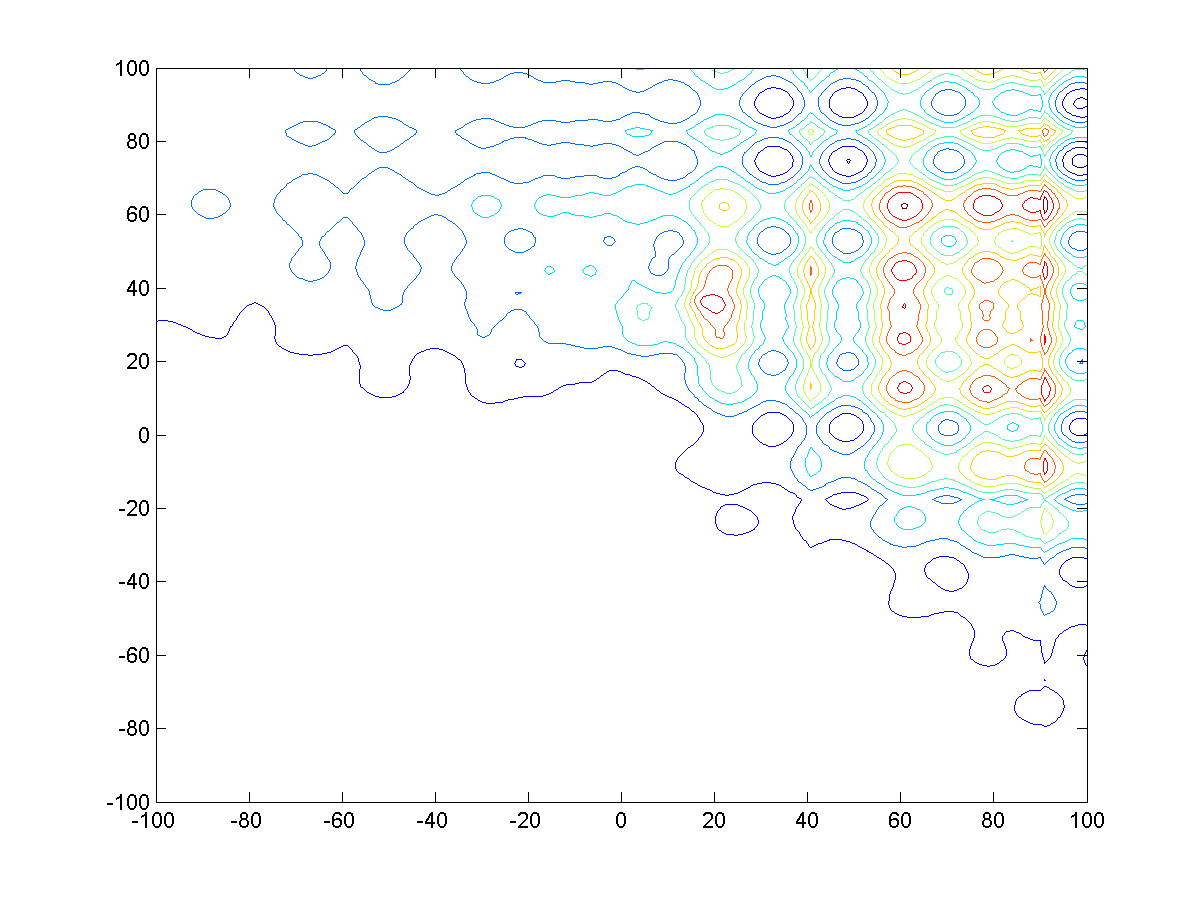}\\
  \caption{Composition Function 2}\label{fig:composition2}
\end{figure}

\begin{figure}[tbp]
  \centering
  \includegraphics[width=.5\textwidth]{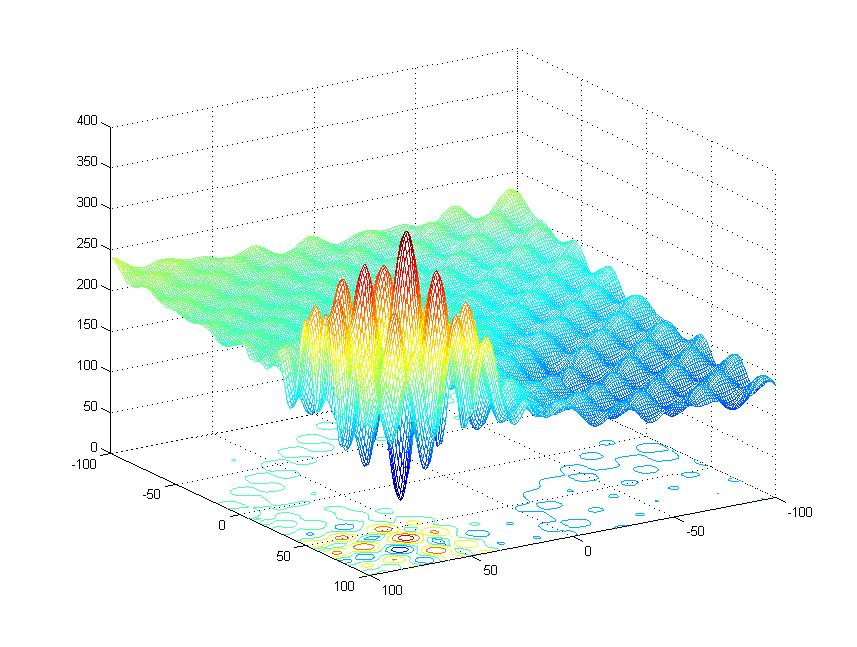}\includegraphics[width=.5\textwidth]{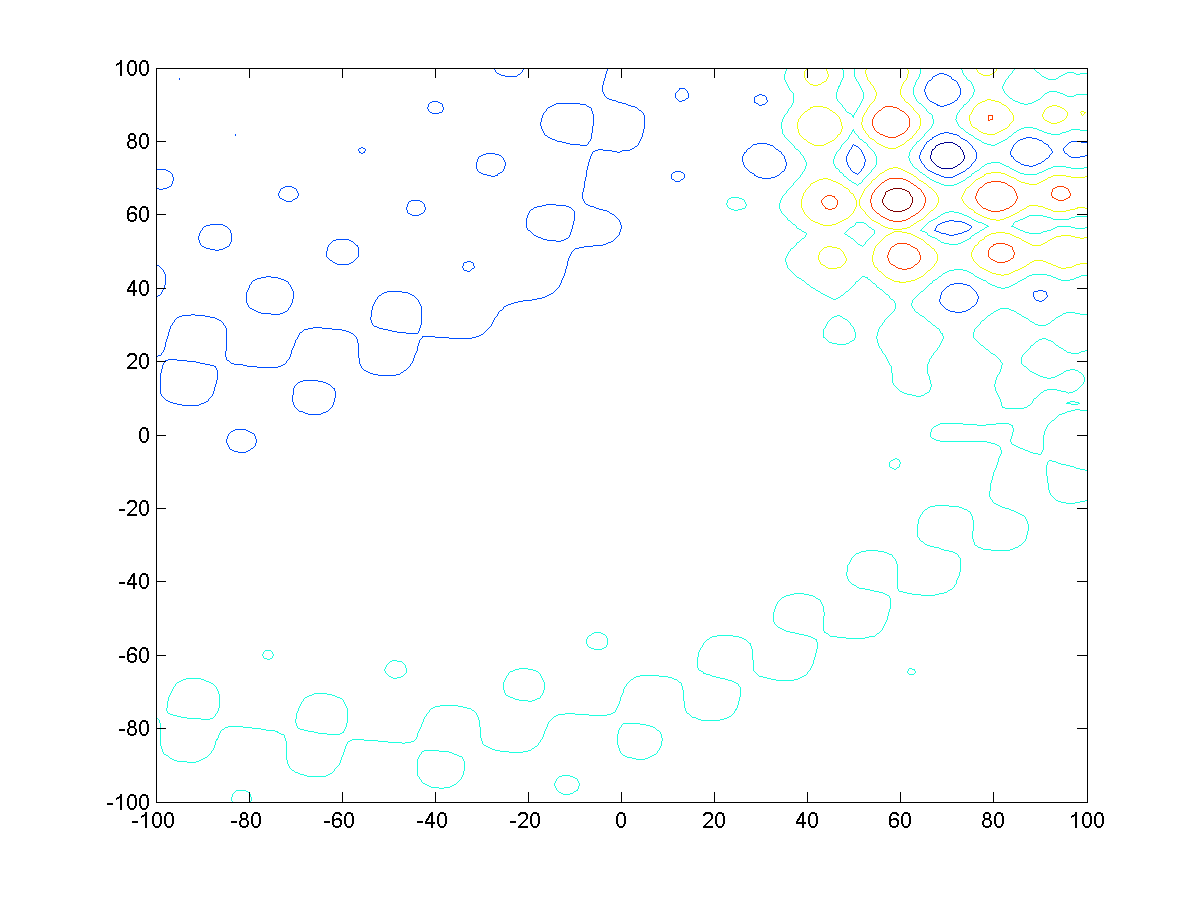}\\
  \caption{Composition Function 3}\label{fig:composition3}
\end{figure}

\begin{figure}[tbp]
  \centering
  \includegraphics[width=.5\textwidth]{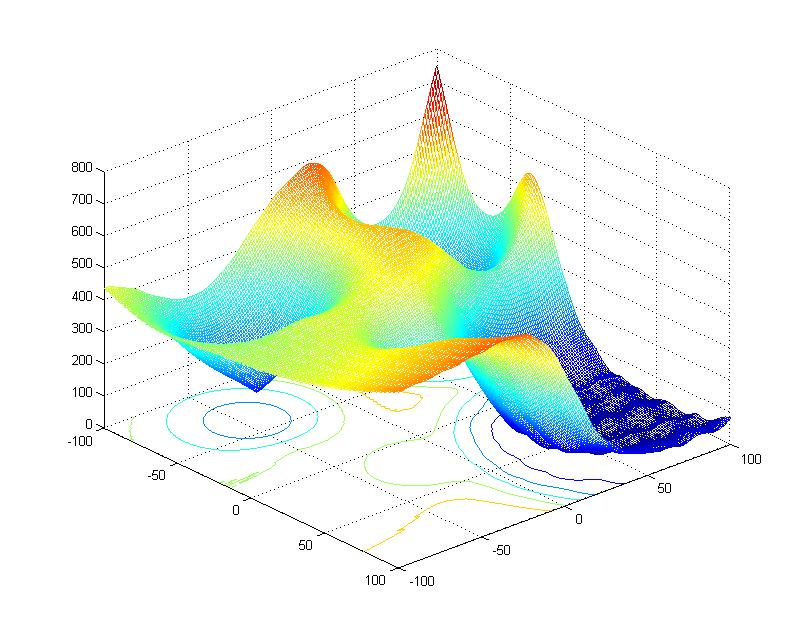}\includegraphics[width=.5\textwidth]{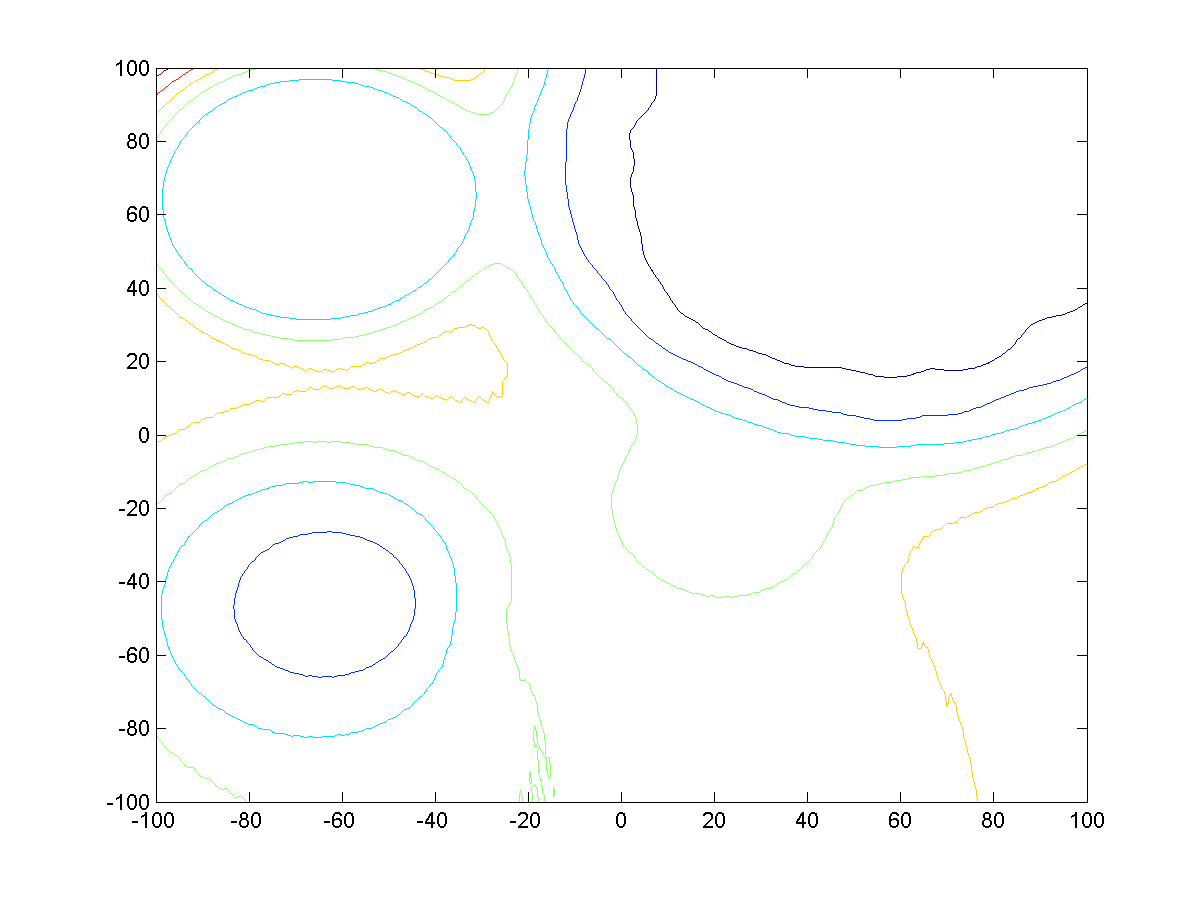}\\
  \caption{Composition Function 4}\label{fig:composition4}
\end{figure}

\begin{figure}[tbp]
  \centering
  \includegraphics[width=.5\textwidth]{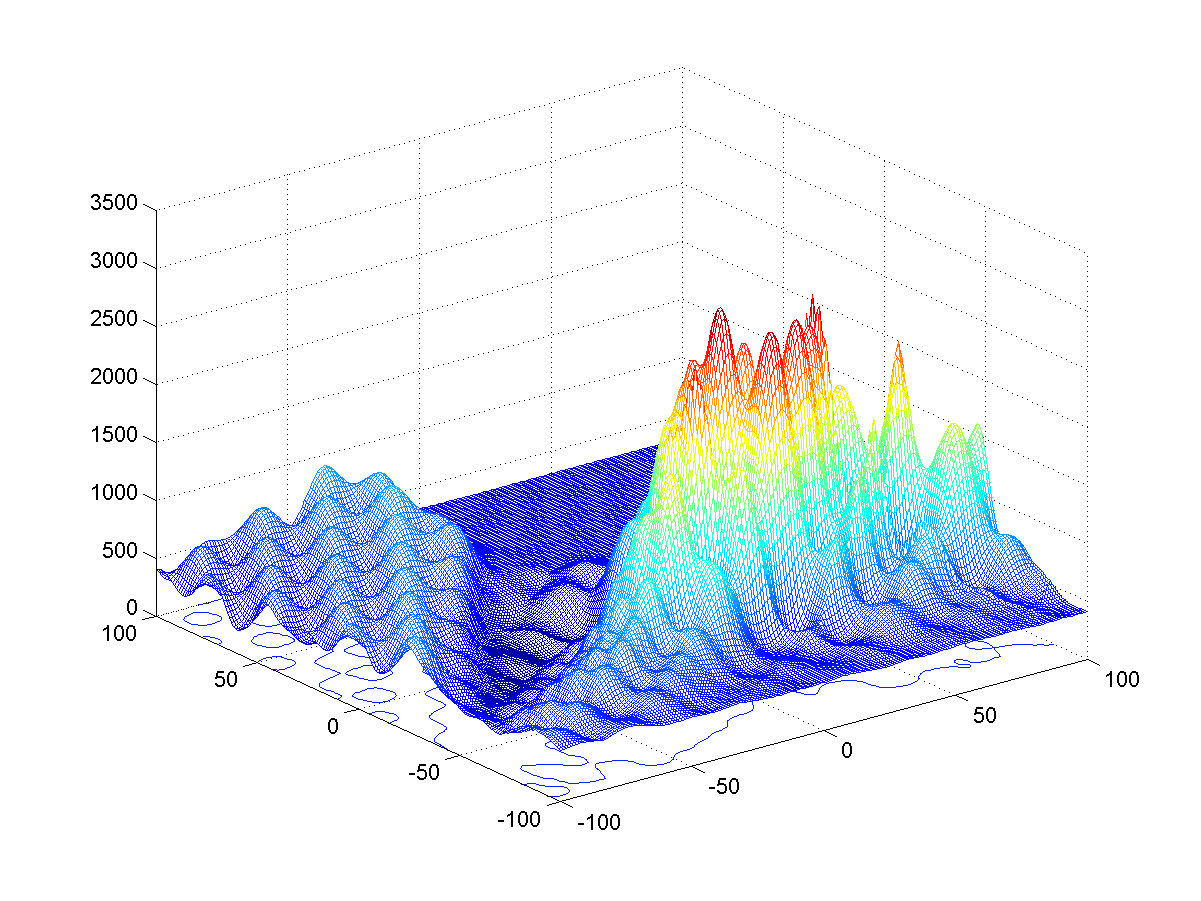}\includegraphics[width=.5\textwidth]{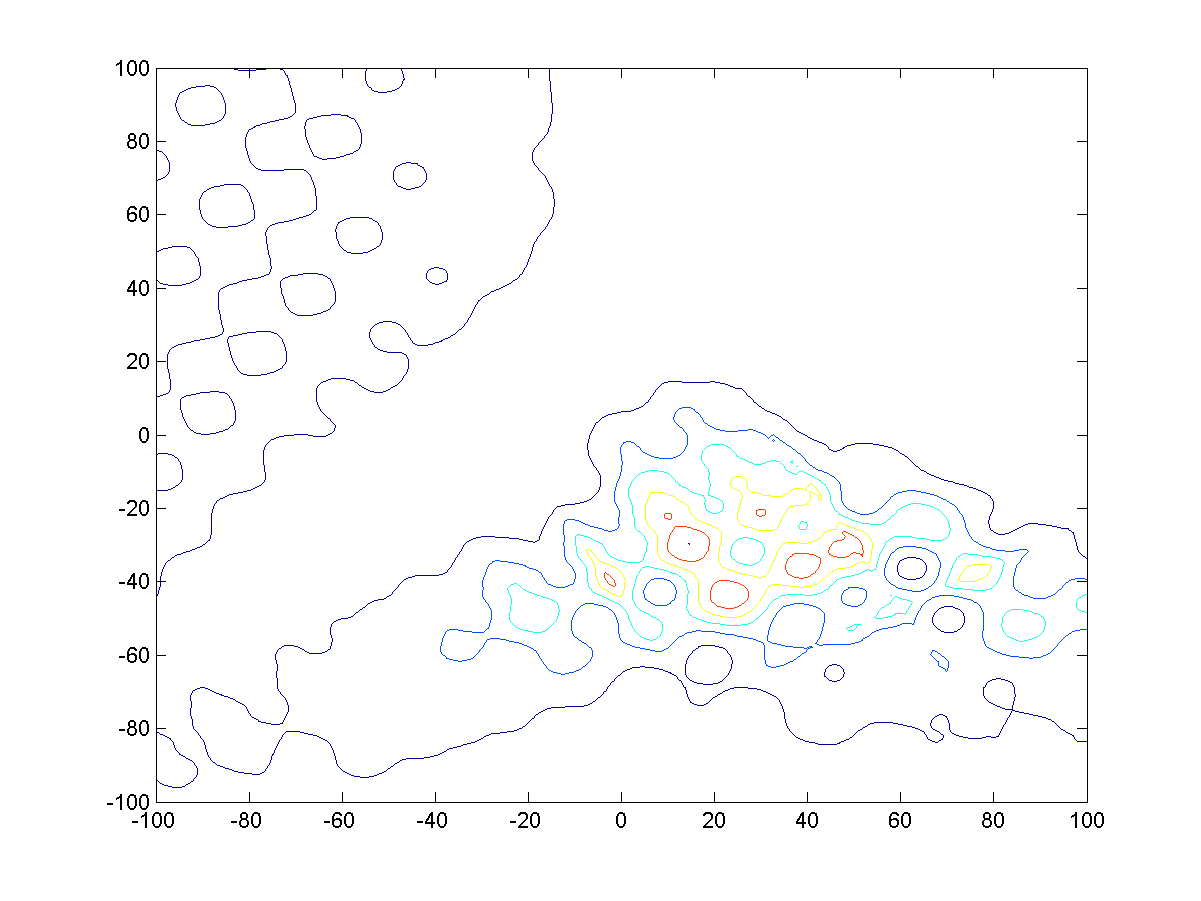}\\
  \caption{Composition Function 5}\label{fig:composition1}
\end{figure}

\begin{figure}[tbp]
  \centering
  \includegraphics[width=.5\textwidth]{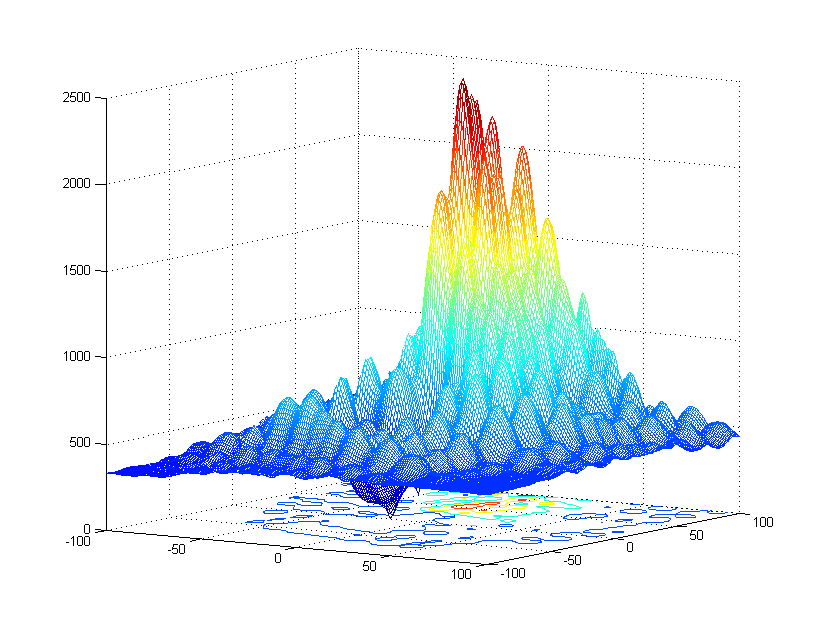}\includegraphics[width=.5\textwidth]{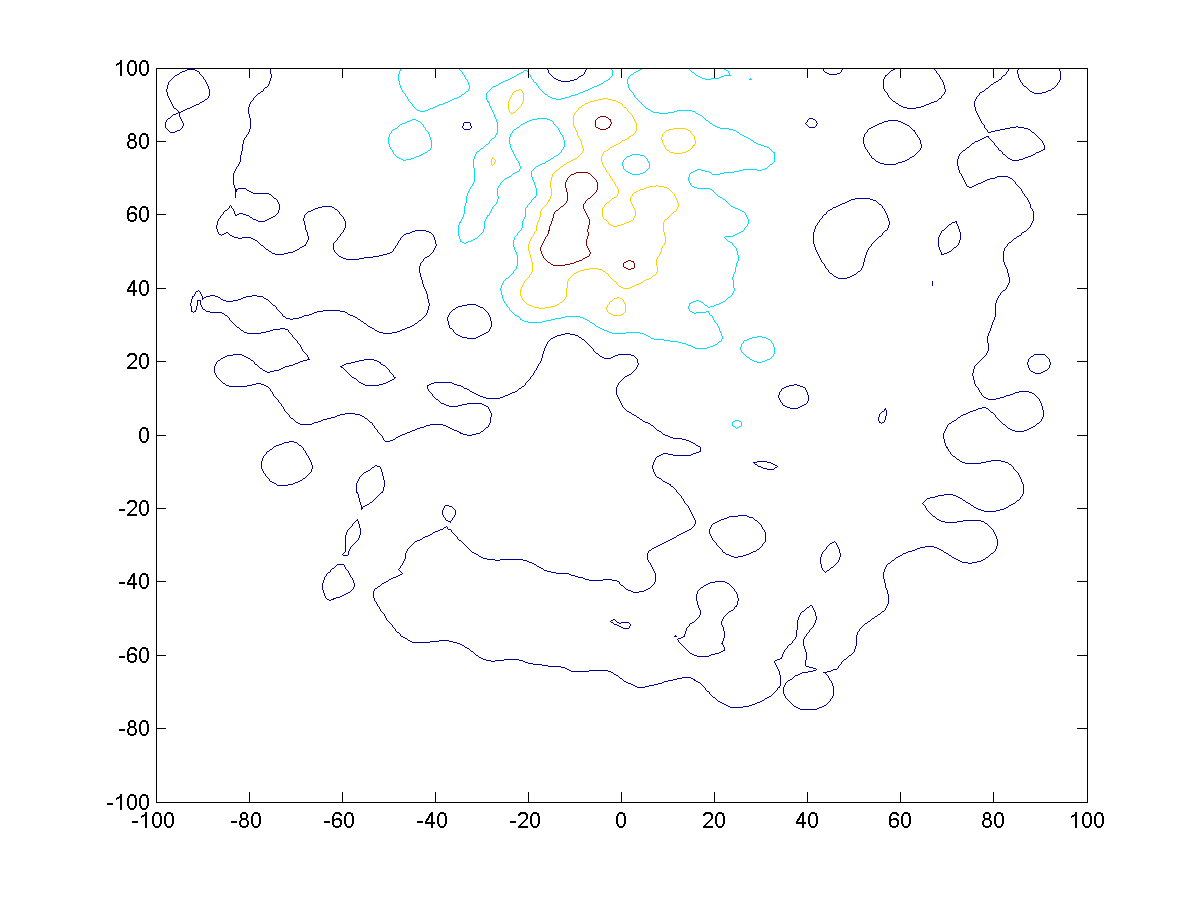}\\
  \caption{Composition Function 6}\label{fig:composition6}
\end{figure}

\end{document}